\documentclass{article}
\usepackage{iclr2027_conference,times}

\usepackage{amsmath,amsfonts,bm}

\def\eqref#1{equation~\ref{#1}}

\def\1{\bm{1}}

\DeclareMathAlphabet{\mathsfit}{\encodingdefault}{\sfdefault}{m}{sl}
\SetMathAlphabet{\mathsfit}{bold}{\encodingdefault}{\sfdefault}{bx}{n}

\usepackage{amsmath,amssymb,bm}
\usepackage{graphicx}
\usepackage{booktabs}
\usepackage{microtype}
\usepackage{placeins}
\usepackage{flafter}
\usepackage{xcolor}
\usepackage{hyperref}
\usepackage{url}
\usepackage{array,tabularx,longtable,float,fvextra,enumitem}
\usepackage{caption}
\hypersetup{hidelinks}
\colorlet{blue}{black}
\graphicspath{{./}}
\usepackage{graphicx,booktabs,array,amssymb,amsmath,caption}
\usepackage{wrapfig}
\usepackage{booktabs,threeparttable}

\newcommand{\figpanel}[3]{%
  \begin{minipage}[t]{#1}
    \centering
    \includegraphics[width=\linewidth]{#2}\\[-0.4ex]
    {\footnotesize #3}
  \end{minipage}%
}

\title{InfoField3D: Differentiable Information Field for Embodied Task-Oriented View Planning}
\author{Huan Lyu, Yawei Luo$^{*}$\\
\normalfont ReLER Lab\\
\normalfont Zhejiang University\\
\normalfont Hangzhou, China\\
\normalfont $^{*}$ Corresponding author: \texttt{yaweiluo@zju.edu.cn}}

\begin{document}
\maketitle

\begin{abstract}

Effective task execution requires embodied agents to select observation positions that provide sufficient task-relevant information. Existing approaches typically predict viewpoints with learned models or score discrete candidates, but they either struggle to incorporate new constraints such as reachability or require extensive candidate evaluation to search. We introduce InfoField3D, which lifts geometry-bound information into a differentiable analytic field, whose value at each 3D observation position represents the task-required information accessible there. It uses low-order spherical harmonics to encode directional information responses and analytic line integrals of Gaussian occupancy functions to model occlusion along lines of sight. This formulation casts task-driven camera placement as a well-studied constrained continuous optimization problem, enabling both flexible incorporation of constraints through the feasible domain and efficient viewpoint search even in the presence of occlusion. Experiments show that InfoField3D compactly preserves the pointwise information landscape with high fidelity and generally recommends higher-quality, more robust viewpoints at lower search cost than the evaluated baselines. Controlled evaluations with a frozen vision-language-action (VLA) policy further suggest that the recommended views can improve downstream task performance under information-limited initial viewpoints.

\end{abstract}
\FloatBarrier
\section{Introduction}
\label{sec:linked-1}

Effective task execution for embodied agents hinges on the acquisition of sufficient task-relevant information, whose availability is strongly conditioned by the agent’s observation viewpoint \citep{cowan1988sensor,tarabanis1995mvp}. This necessity gives rise to the task-driven view planning task \citep{breyer2022closedloopnbv,shi2025visograsp}. Existing approaches to this task commonly follow two main lines. The first line leverages learning-based models to directly predict the observation viewpoint \citep{bai2026tavp}, yet such implicit end-to-end approaches struggle to accommodate constraints in novel scenes, such as reachability. The second line adopts sampling-based models to evaluate abundant candidate viewpoints \citep{dai2025apeg,xie2025gaussmi,jiang2024fisherrf}, which can suffer from discontinuity over the continuous spatial domain. Moreover, both families of methods perform predicting or scoring operations that rely on geometry-bound surface features, for which effective occlusion handling remains nontrivial in this task and may affect the practical accuracy of the recommended viewpoints.

In this paper, we propose a new paradigm dubbed InfoField3D for embodied task-oriented view planning. Rather than directly operating on geometry-bound features, InfoField3D lifts the surface information to an analytic field, whose value at each 3D observation position represents the task-required information accessible from that location. Specifically, InfoField3D represents the inherent directional information distribution of dense surface samples with low-order spherical harmonics, enabling it to be computed in a differentiable and low-redundancy manner. Meanwhile, conventional discrete occlusion computation is reformulated as a continuous representation of object occupancy in space, with the degree of occlusion quantified by integrating occupancy along line-of-sight paths. In this way, information accessibility in the workspace is transformed into a fully differentiable analytic formulation. 

InfoField3D offers three core advantages: (i) It casts viewpoint planning as a well-established constrained continuous optimization problem defined over the feasible domain, enabling embodied 

\begin{wrapfigure}{L}{0.5\columnwidth}
  \centering
  \includegraphics[
    width=\linewidth,
    keepaspectratio
  ]{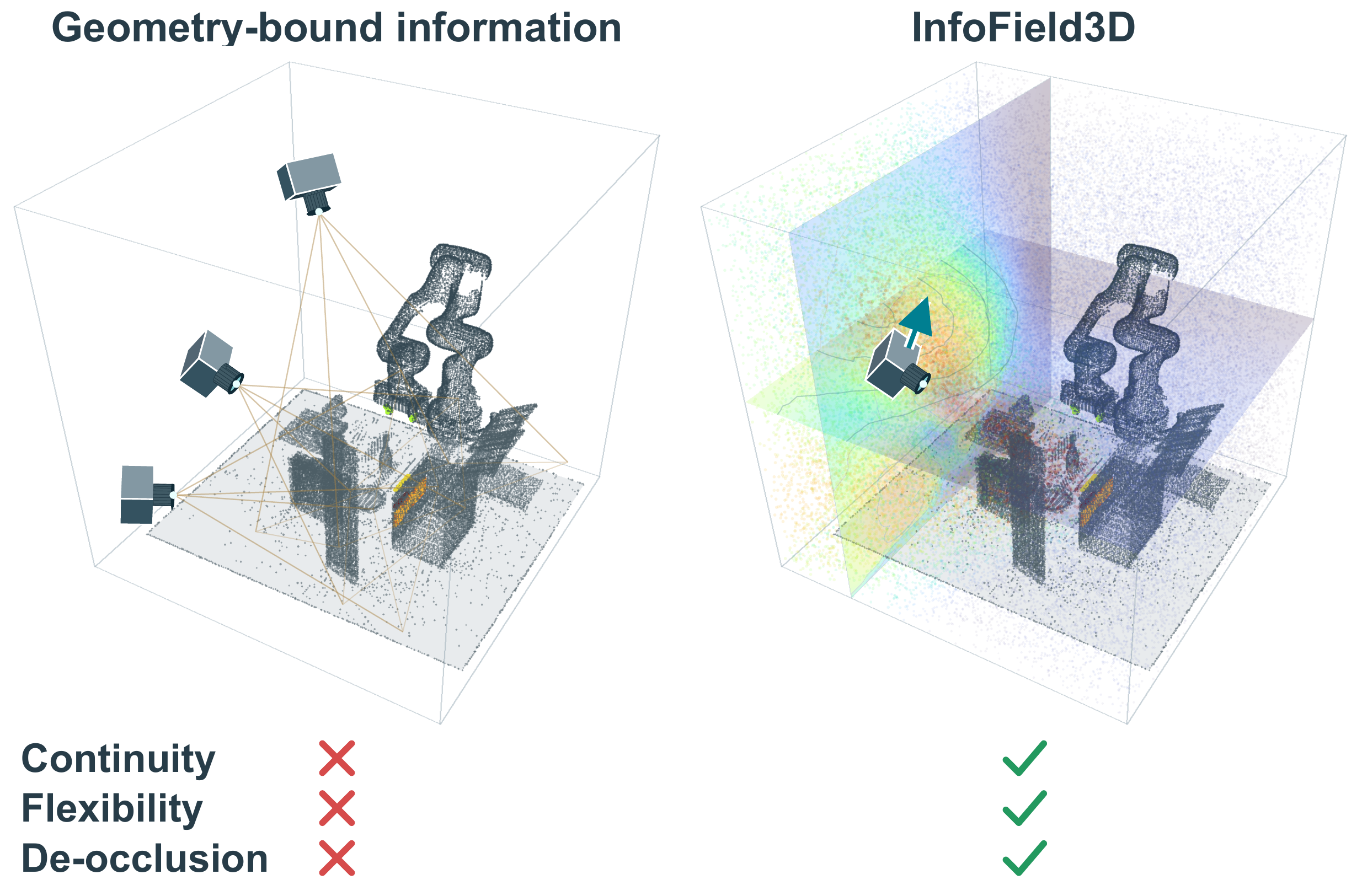}
  \caption{View planning based on (1) Geometry-bound information \emph{vs.} (2) InfoField3D.}
  \label{fig:firstfigurenew}
  \vspace{-7mm}
\end{wrapfigure}

agents to efficiently search for high-quality viewpoints using analytic gradients. (ii) New constraints (e.g., reachability) can be flexibly incorporated into the field-based formulation, facilitating better adaptation to novel scenes and tasks. (iii) Occlusion is modeled in a differentiable form, providing gradients that guide viewpoints away from occlusion and toward positions with larger visibility margins.

Experiments on publicly available benchmarks show that InfoField3D closely approximates the spatial information distribution computed directly from the original point cloud using substantially fewer representation parameters. Compared with prior art, it generally recommends higher-quality observation positions with greater robustness to positional perturbations and lower online search cost. Controlled evaluations with a frozen VLA model suggest that the recommended views can provide modest improvements in downstream task performance.

\raggedbottom
\section{Related Work}
\label{sec:linked-2}

Existing methods for task-driven observation planning can be broadly grouped into the following two categories according to their primary mechanism for selecting or updating observation positions.

\textbf{Sampling-based methods.} These methods select observation positions by scoring candidate viewpoints or paths \citep{triggs1995camera,breyer2022closedloopnbv,lee2026here,jauhri2024actpermoma}, using geometric, semantic, or information-based criteria from 3D scenes \citep{liu2026activevla,nagami2026vista,dai2025apeg,ong2026atlas,burusa2024semanticsnbv,song2025gsnbv} to rank candidates for tasks such as reconstruction, exploration, and grasping \citep{xie2025gaussmi,jiang2024fisherrf,jiang2025mllmmapping,lei2026activegrasp,strong2025nextbestsense,wilson2025popgs,song2026gopre}. These methods can accommodate diverse scenes and tasks, but candidate scores themselves do not directly provide directions for continuous position updates, so selection quality depends on candidate coverage and the evaluation budget. Precomputed discrete fields move part of the evaluation to an offline stage and support online queries through interpolation, but introduce grid-dependent approximations \citep{zhang2020fif}. Some methods differentiate utility surrogates fitted from samples \citep{wang2024cim} or discretized sensing estimators \citep{burusa2024gradientnbv,zhu2026bayesiannbv}, whose evaluation and gradients still depend on discrete sampling.  Analytic geometric control is also used in some methods to directly generate motion directions based on target--occluder relationships or line-of-sight geometry \citep{shi2025visograsp,he2022visibility}; however, such guidance serves predefined geometric objectives and need not correspond to task-adjustable information utilities and their gradients.

\textbf{Learning-based prediction methods.} These methods use neural networks to directly predict camera poses or motions, conditioned on scene observations, task text, and other inputs \citep{bai2026tavp,li2025harvestview,jayaraman2018lookaround,liu2025rasp,qin2026mvpnet,koo2025vgavs}, or jointly learn camera policies together with manipulation policies \citep{liu2026sapave,chuang2025avaloha,xiong2025visioninaction,kerr2025eyerobot,wang2025observethenact}. They support efficient inference, but their generalization to scenes, occlusion relationships, tasks, and embodiments outside the training distribution may be limited. In addition, such implicit end-to-end approaches typically lack explicit constraint interfaces, making it difficult to incorporate constraints in novel scenes, such as reachability.

\section{InfoField3D: Formulation and Methodology}
\label{sec:linked-3}

\subsection{Problem Formulation and Overview}
\label{sec:linked-3-1}

\begin{figure}[!tb]
  \centering
  \figpanel{0.210\textwidth}{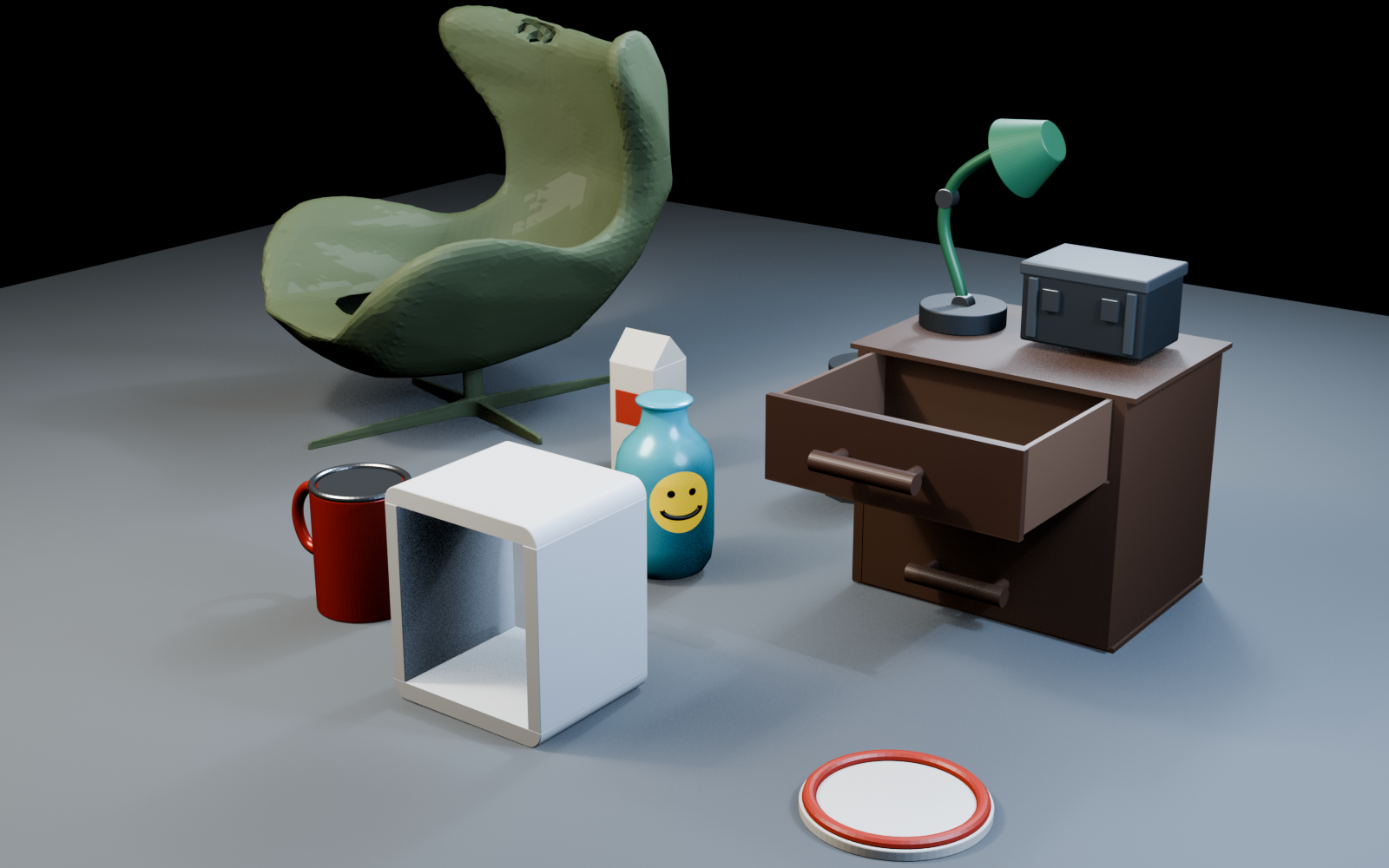}{(a)}\hfill
  \figpanel{0.151\textwidth}{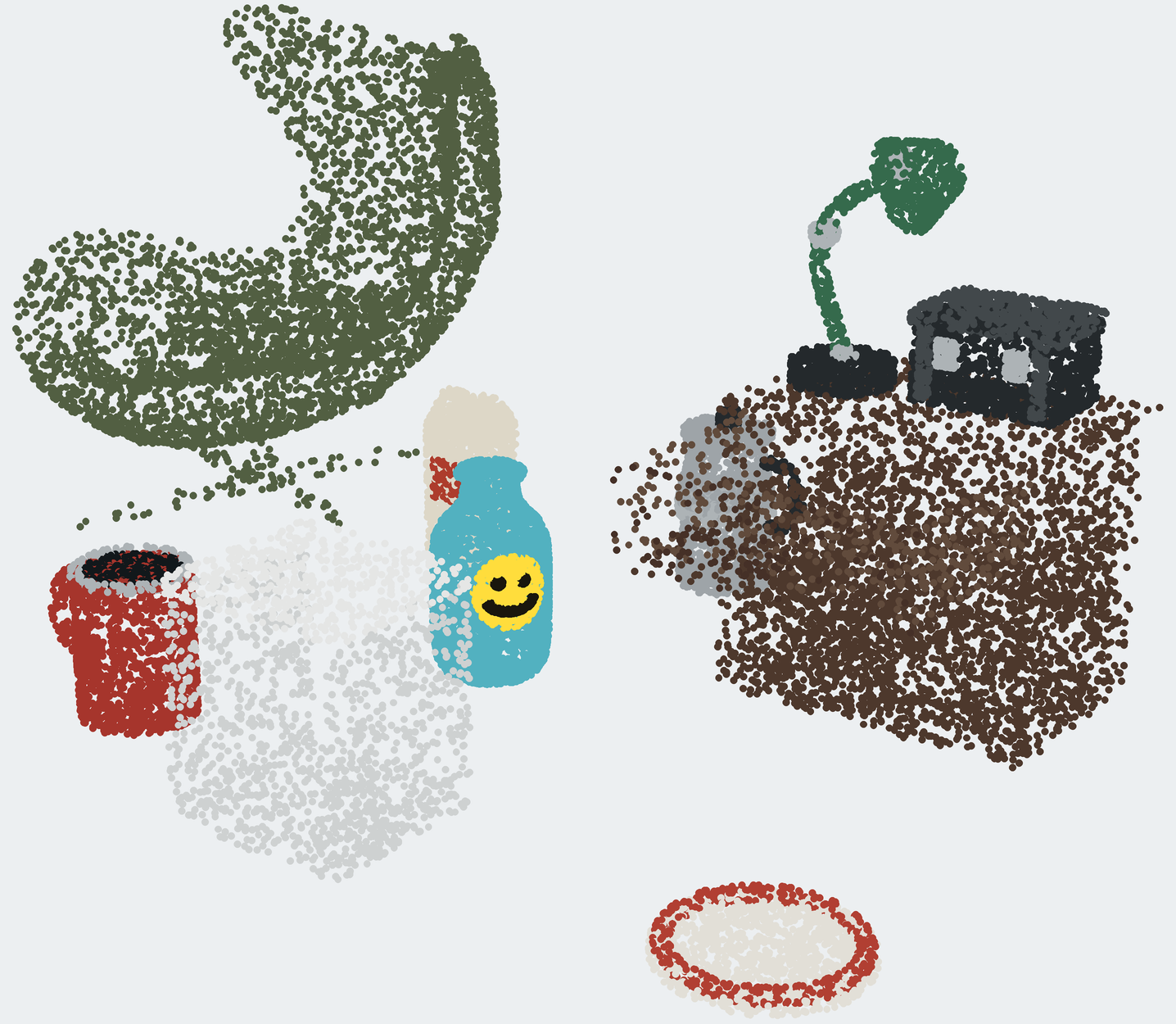}{(b)}\hfill
  \figpanel{0.151\textwidth}{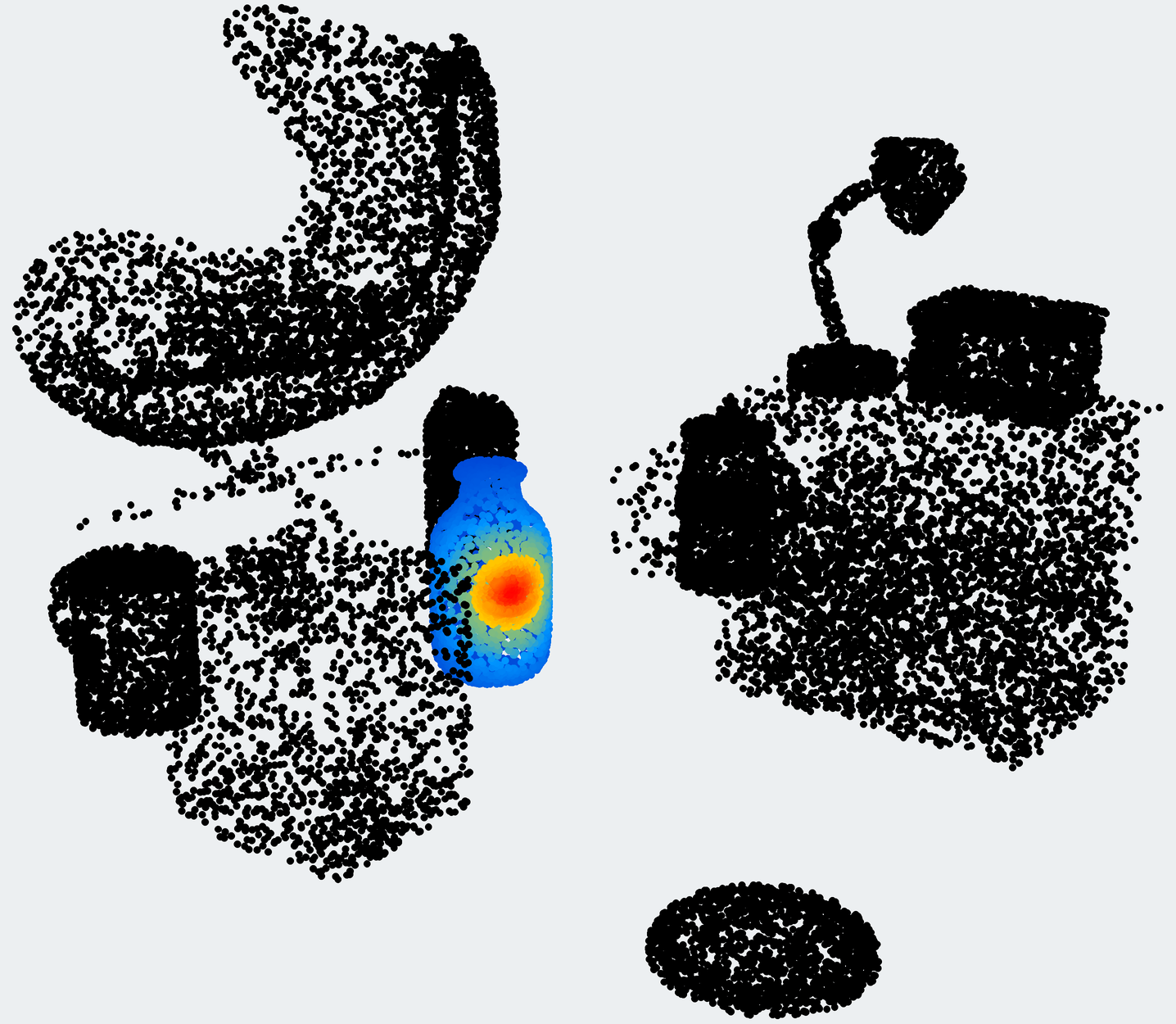}{(c)}\hfill
  \figpanel{0.133\textwidth}{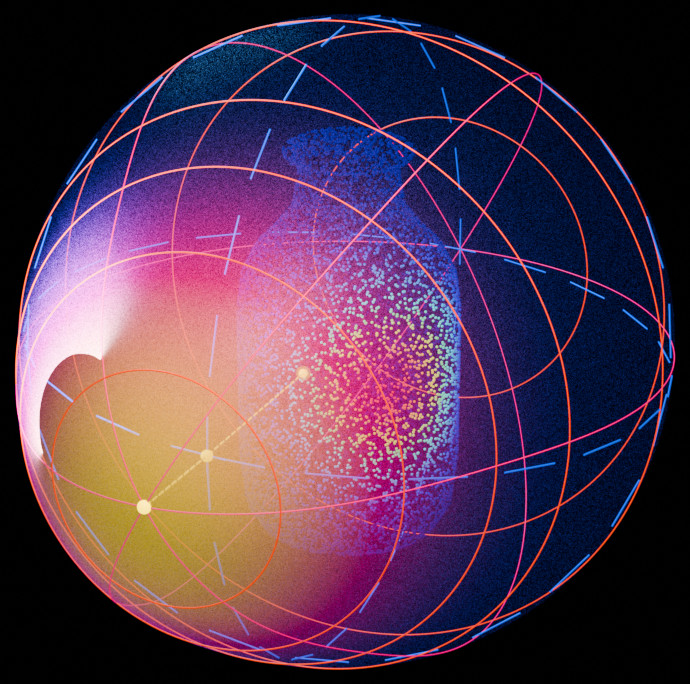}{(d)}\hfill
  \figpanel{0.153\textwidth}{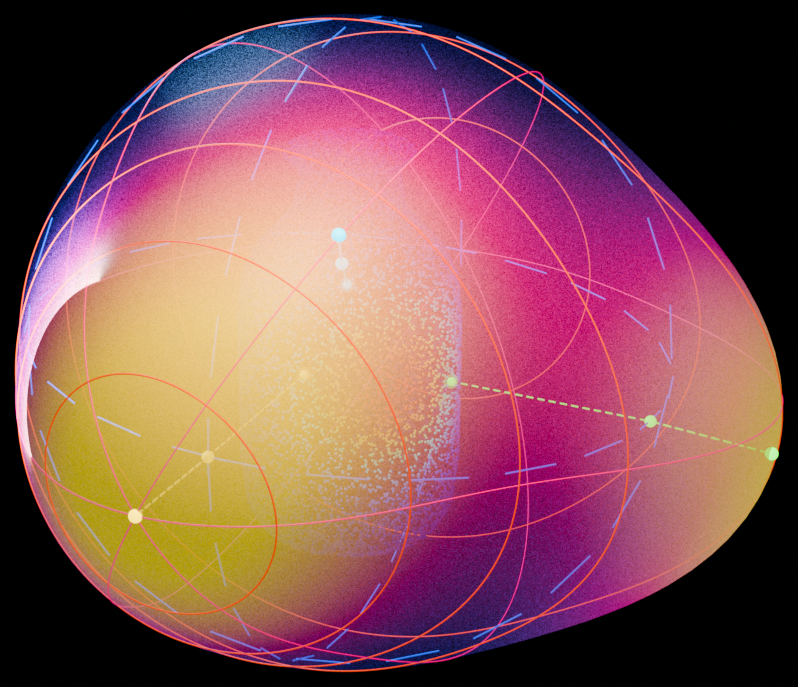}{(e)}\hfill
  \figpanel{0.137\textwidth}{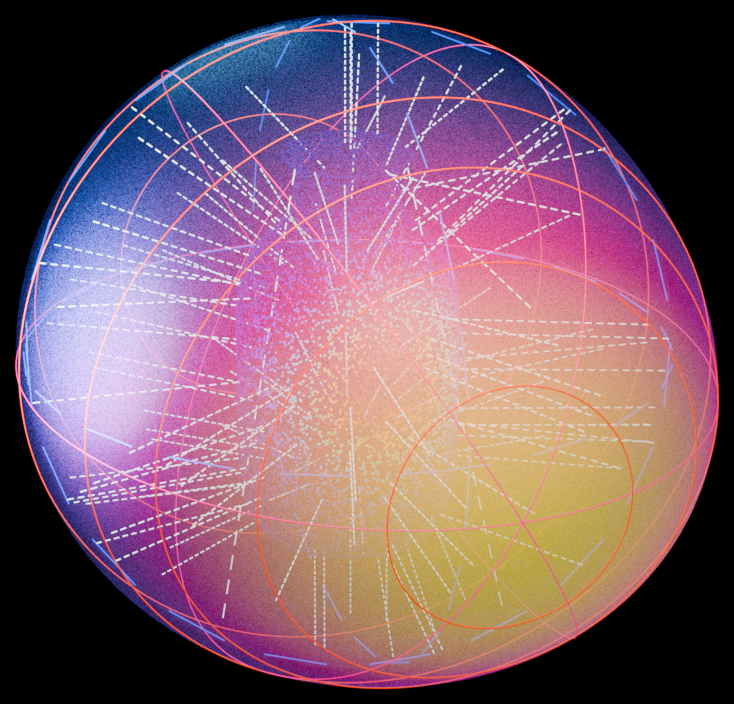}{(f)}
  \caption{Task example and the construction from pointwise responses to a compact directional representation: (a) the task scene; (b) its point cloud; (c) the task-relevance map; (d--e) progressive superposition of pointwise angular responses on the reference sphere, weighted by task relevance; (f) the aggregated object-level distribution.}
  \label{fig:point-superposition}
  \vspace{-4mm}
\end{figure}

This paper considers task-oriented view planning in a reconstructed static scene represented by a point cloud enriched with geometry-bound semantic features \citep{kerr2023lerf,zhou2024feature3dgs,qin2024langsplat}. Our goal is to model the accessibility of geometry-bound scene information as a function of the observation position $\mathbf{x}\in\Omega\subseteq\mathbb{R}^3$, where $\Omega$ denotes the feasible domain satisfying the prescribed constraints. To remain independent of specific camera configurations, we assume ideal panoramic observation and optimize only the observation position, with practical camera constraints handled through orientation rules and $\Omega$. To facilitate exposition, here we use an intuitive example to demonstrate the entire construction process of InfoField3D: \emph{Consider an embodied agent tasked with writing a signature on a smiley-face sticker attached to a bottle (Fig.~\ref{fig:point-superposition}(a)). Which observation position best supports this task?}

To determine what needs to be observed for signing, a multimodal model interprets the task condition together with the current RGB images, drawing on learned priors to identify, for example, the smiley-face sticker as the primary observation target and formulate the query “the smiley-face sticker on the bottle.” Applying this query to the semantic point cloud assigns each point a nonnegative task-relevance score $s_i$, forming a \emph{\textbf{task-relevance map}} over the scene. The total relevance score of the points visible from a given position provides an intuitive measure of observation quality.

Let \(J_{\mathcal T}(\mathbf x)\) denote a scalar measure of the task-relevant information accessible at position \(\mathbf x\) under task condition \(\mathcal T\). The desired observation position maximizes this response over $\Omega$:

\begin{equation}
\mathbf{x}^{\star}=\underset{\mathbf{x}\in\Omega}{\operatorname{arg\,max}}\;J_{\mathcal{T}}(\mathbf{x}).
\label{eq:problem-objective}
\end{equation}

\begin{figure*}[t]
  \centering
  \includegraphics[width=0.81\textwidth,keepaspectratio]{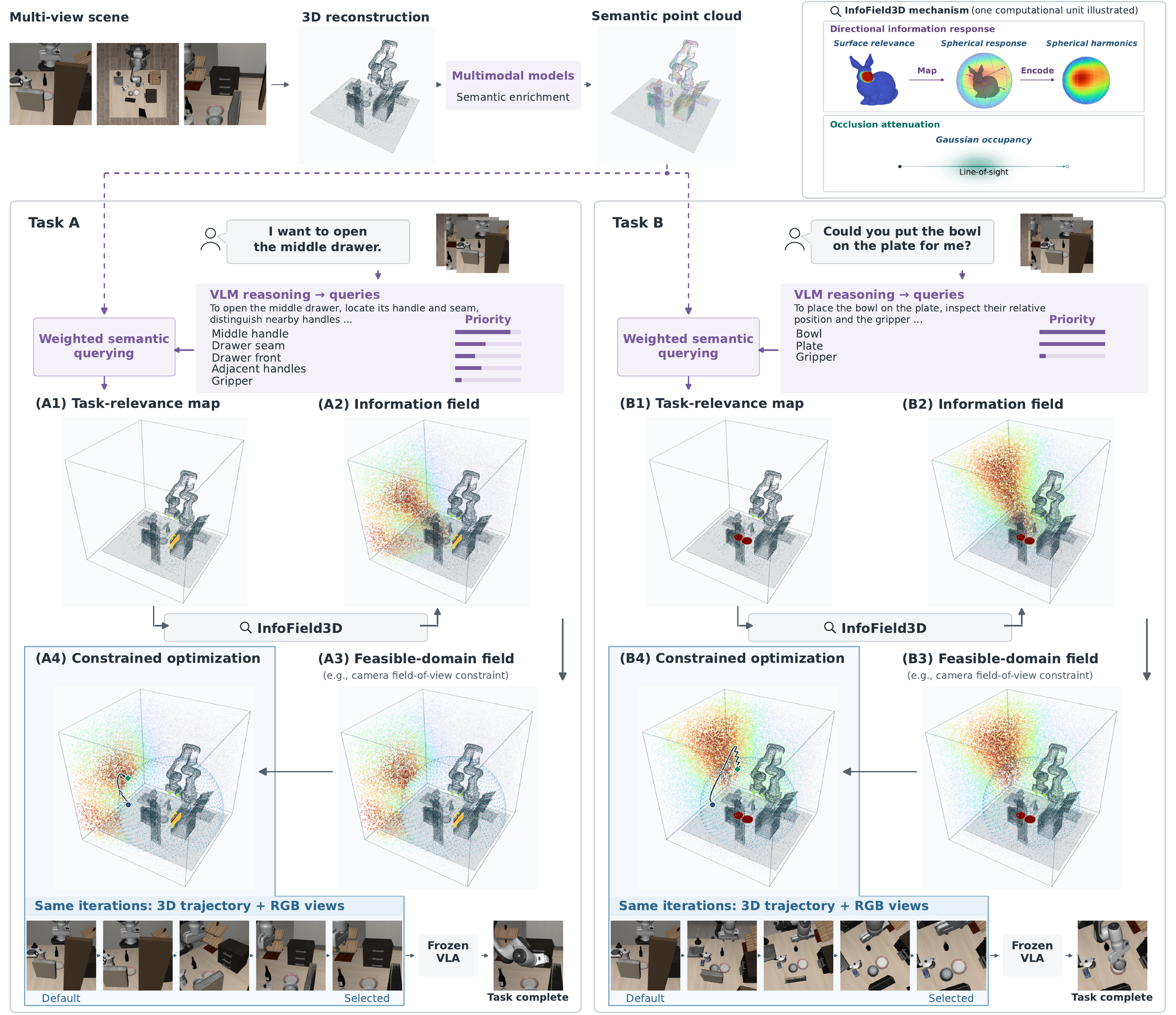}
  \caption{\textbf{Overview of InfoField3D.} Task-relevance maps (A1, B1) can be obtained from multimodal foundation models conditioned on the task and the reconstructed 3D scene; the figure shows one illustrative example. InfoField3D lifts geometry-bound information into a differentiable analytic field, whose value encodes task-relevant information accessibility at each observation position (A2, B2). The field is then restricted to the feasible domain (A3, B3), and the recommended observation position is obtained through constrained optimization over this field (A4, B4). Within the same scene, viewpoints recommended according to different task requirements are provided to a frozen VLA to execute the corresponding tasks.}
  \label{fig:omnifield3d-overview}
  \vspace{-6mm}
\end{figure*}

The problem is then reduced to determining $\mathbf{x}$ so that $J_{\mathcal{T}}(\mathbf{x})$ is maximized. InfoField3D constructs $J_{\mathcal{T}}(\mathbf{x})$ as a differentiable analytic function of the observation position. The following sections derive this construction in turn, covering directional information responses (\hyperref[sec:linked-3-2]{Section 3.2}), occlusion along the propagation path (\hyperref[sec:linked-3-3]{Section 3.3}), and the local geometric units that extend the construction to the full scene (\hyperref[sec:linked-3-4]{Section 3.4}). For a visual overview of the overall workflow, see Figure~\ref{fig:omnifield3d-overview}.  \hyperref[app:method-details]{Appendix~\ref*{app:method-details}} provides additional construction details and several practical extensions.

\flushbottom
\subsection{Differentiable Modeling of Information Propagation}
\label{sec:linked-3-2}

The derivation first considers the bottle without occlusion from other objects. Throughout Sections 3.2–3.3, we fix \(\mathcal T\) and suppress its dependence in the notation. 

\textbf{Pointwise response.} For a surface point $\mathbf{p}_i$, its information exposure varies with the alignment between its outward surface normal and the direction toward the observation position $\mathbf{x}$. The cosine of the angle between these directions gives a continuous and differentiable measure of the information exposure of $\mathbf{p}_i$ at $\mathbf{x}$, and a smooth kernel $K(t)$ of the cosine value $t$ (exponential in our implementation)  can further adjust the response shape and angular concentration. For $\mathbf{p}_i$ with unit normal $\mathbf{n}_i$ and task-relevance score $s_i$, its information contribution at $\mathbf{x}$ ($\mathbf{x}\neq\mathbf{p}_i$) is $f_i(\mathbf{x})=s_iK\left(\mathbf{n}_i\cdot\frac{\mathbf{x}-\mathbf{p}_i}{\|\mathbf{x}-\mathbf{p}_i\|}\right)$. For an approximately uniform $N$-point cloud, the individual point contributions sum to the total response at an observation position $\mathbf{x}$, $F(\mathbf{x})=\sum_{i=1}^{N}f_i(\mathbf{x})$.

\textbf{Spherical mapping.}  The response $F(\mathbf{x})$ is continuous and differentiable at valid observation positions, but its representation size and evaluation cost remain proportional to the point count. Dense geometric sampling, however, need not imply an equally complex information distribution. This motivates seeking an object-level representation whose complexity is no longer tied to the point count. To place the points’ angular responses on a common domain, InfoField3D constructs a reference sphere $S$ centered at the center $\mathbf{c}$ of the object’s minimum enclosing sphere, with a radius $R$ larger than the minimum enclosing radius. For each point, a ray along its outward normal intersects the sphere, defining the center of a continuous angular response. Weighted by the point's task-relevance score, this response forms a directional hill that decays with angular distance from the intersection. Superimposing the hills from all points yields an object-level directional response distribution whose high-response regions identify directions with greater task-relevant information (Figure~\ref{fig:point-superposition}(d-f)).

On the spherical domain, this mapping yields the approximate response $F_{\mathrm{sph}}(\mathbf{u})=\sum_{i=1}^{N}s_iK(\mathbf{r}_i\cdot\mathbf{u})$. The above holds for all unit directions $\mathbf{u}$, where $\mathbf{r}_i$ denotes the unit direction from $\mathbf{c}$ to the intersection of the outward-normal ray from $\mathbf{p}_i$ with the reference sphere. This spherical construction provides a basis for seeking a compact representation of the object's directional response. 
Spherical harmonics provide an orthogonal basis adapted to the sphere's rotational symmetry and organized by angular frequency \citep{ramamoorthi2001irradiance}, \textcolor{blue}{making them useful for encoding view-dependent color in 3D scene representation \citep[e.g.,][]{kerbl2023gaussian} and} motivating a finite-degree expansion to test whether a few low-frequency modes can preserve the distribution's main structure.  

\textbf{Harmonic encoding.} The degree-$L$ approximation of $F_{\mathrm{sph}}$ is $F_L(\mathbf{u})=\sum_{\ell=0}^{L}\sum_{m=-\ell}^{\ell}c_{\ell m}Y_{\ell m}(\mathbf{u})$, where $Y_{\ell m}$ are the spherical harmonic basis functions and $c_{\ell m}$ are the coefficients. For a fixed degree $L$, the representation requires only $(L+1)^2$ coefficients, regardless of the point count. The evaluations in \hyperref[sec:linked-4-1]{Section 4.1} show that this compact representation preserves the original pointwise spatial responses with high fidelity, supporting the hypothesis that a few low-frequency modes suffice. The following derivation therefore uses this compact representation for observation-position optimization. For $\mathbf{x}\neq\mathbf{c}$, the unit direction from the center to the observation position is $\mathbf{u}(\mathbf{x})=(\mathbf{x}-\mathbf{c})/\|\mathbf{x}-\mathbf{c}\|$. The object-level response at $\mathbf{x}$ is then $F_L(\mathbf{u}(\mathbf{x}))$, serving as the objective function for observation-position search. The search therefore reduces to maximizing it over $\Omega$:

\begin{equation}
\mathbf{x}^{\star}=\underset{\mathbf{x}\in\Omega}{\operatorname{arg\,max}}\;F_L(\mathbf{u}(\mathbf{x})).
\label{eq:search-unoccluded}
\end{equation}

Since the spherical harmonic basis functions and direction mapping are continuously differentiable, the chain rule gives the objective's analytic spatial gradient: $\nabla_{\mathbf{x}}F_L(\mathbf{u}(\mathbf{x}))=\frac{1}{\|\mathbf{x}-\mathbf{c}\|}\left(\mathbf{I}-\mathbf{u}(\mathbf{x})\mathbf{u}(\mathbf{x})^\top\right)\nabla_{\mathbf{u}}F_L(\mathbf{u}(\mathbf{x}))$. The problem of increasing the objective can now be formulated as an optimization problem.

\subsection{Differentiable Modeling of Occlusion}
\label{sec:linked-3-3}

How should this formulation account for occlusion along the propagation path? With a cube placed in front of the bottle, an intuitive score sums the task-relevance scores of the bottle’s unoccluded points. However, classifying points as occluded or unoccluded provides no useful spatial gradients for position updates.

\textbf{Path attenuation.}  To address this limitation, InfoField3D shifts the perspective on occlusion. Instead of classifying points as occluded or unoccluded, it quantifies the spatial occupancy accumulated along the propagation path. Once this occupancy is represented as a continuously differentiable function, the accumulated quantity can be given by a line integral, which remains differentiable with respect to the observation position. In this example, a Gaussian occupancy function $\rho$\textcolor{blue}{, a primitive widely used in 3D scene representation \citep{kerbl2023gaussian},} approximates the cube’s occupied region. Using the object's reference center $\mathbf{c}$ as its representative source location, let $\tau(\mathbf{x})$ denote the occupancy accumulated along the finite segment from $\mathbf{c}$ to the observation position $\mathbf{x}$, given by $\tau(\mathbf{x})=\int_{[\mathbf{c},\mathbf{x}]}\rho(\mathbf{z})\,\mathrm{d}s$, and let the corresponding transmittance be $T(\mathbf{x})=\exp\left(-\tau(\mathbf{x})\right)$.

The transmittance $T(\mathbf{x})$ represents the fraction of the directional response retained after propagation, so multiplying it by the directional response $F_L(\mathbf{u}(\mathbf{x}))$ gives the task-relevant response available at $\mathbf{x}$ under occlusion. This product defines the single-object objective $J_{\mathrm{obj}}(\mathbf{x})=T(\mathbf{x})F_L(\mathbf{u}(\mathbf{x}))$, and observation-position search becomes

\begin{equation}
\mathbf{x}^{\star}=\underset{\mathbf{x}\in\Omega}{\operatorname{arg\,max}}\;T(\mathbf{x})F_L(\mathbf{u}(\mathbf{x}))
=\underset{\mathbf{x}\in\Omega}{\operatorname{arg\,max}}\;J_{\mathrm{obj}}(\mathbf{x}).
\label{eq:search-occluded}
\end{equation}

The objective $J_{\mathrm{obj}}(\mathbf{x})$ remains continuously differentiable and can be maximized in the same way as in \hyperref[sec:linked-3-2]{Section 3.2}. Its gradient guides the search toward positions with more task-relevant information available under occlusion.

\textbf{Visibility margins.}  Notably, scores based solely on binary visibility saturate once the target is fully visible. The present objective does not stop at this point, because the Gaussian occupancy extends over the entire space; paths passing close to an occluder still accumulate attenuation, so the objective can favor positions with larger observation margins and lower occlusion risk. Consistent with this effect, the experiments in \hyperref[sec:linked-4-3]{Section~4.3} show that InfoField3D recommends observation positions with greater robustness to positional perturbations.

\subsection{Defining the Geometric Units of the Field}
\label{sec:linked-3-4}

The preceding examples use whole objects as units. Representing an entire object's directional response around a shared center can introduce geometric mapping loss before spherical harmonic encoding, and likewise for a single-Gaussian approximation of an entire occluder. For example, a chair has spatially separated legs, each with its own directional information response and spatial occupancy, making the whole object difficult to represent around a shared center without loss. InfoField3D therefore does not adopt whole objects or predefined semantic parts as default computational units, since either may have complex geometry. Instead, it constructs local geometric units suited to its representations, each with its own reference center.

\textbf{Geometric partitioning.}  Experiments in \hyperref[app:geometric-partitioning]{Appendix~\ref*{app:geometric-partitioning}} show that geometrically complex, less rounded units tend to incur greater loss under a single-center representation. Guided by this observation, we construct local units to be more geometrically compact using a simple recursive partitioning scheme without semantic part priors: the unit with the greatest spatial spread is repeatedly split into two balanced subsets along its principal axis until the desired granularity is reached. Each resulting unit serves as a basic computational unit. \hyperref[app:geometric-partitioning]{Appendix~\ref*{app:geometric-partitioning}} further shows that a small number of such splits substantially improves accuracy.

\textbf{Field aggregation.}  In the resulting computation, each unit acts as both an information source and an occluder for other units, with the source unit's own spatial occupancy excluded from its propagation path. For example, once a cup's handle and body belong to different units, their mutual occlusion is included in the calculation. With $M$ local units, the same information-response and occlusion rules define the objective over the full scene. For unit $j$, let $\mathbf{c}_j$ denote its reference center, $\rho_j$ its Gaussian occupancy function, and $F_{j,L}^{(\mathcal{T})}$ its directional response under the task condition $\mathcal{T}$. The observation direction is $\mathbf{u}_j(\mathbf{x})=(\mathbf{x}-\mathbf{c}_j)/\|\mathbf{x}-\mathbf{c}_j\|$. The accumulated path occupancy is $\tau_j(\mathbf{x})=\int_0^1\left[\sum_{k\ne j}\rho_k\bigl(\mathbf{c}_j+t(\mathbf{x}-\mathbf{c}_j)\bigr)\right]\|\mathbf{x}-\mathbf{c}_j\|\,\mathrm{d}t$, with the corresponding transmittance $T_j(\mathbf{x})=\exp\bigl(-\tau_j(\mathbf{x})\bigr)$. Aggregating the attenuated responses over all units gives the full-scene objective under the task condition, $J_{\mathcal{T}}(\mathbf{x})=\sum_{j=1}^{M}T_j(\mathbf{x})F_{j,L}^{(\mathcal{T})}\bigl(\mathbf{u}_j(\mathbf{x})\bigr)$, which realizes the objective $J_{\mathcal{T}}(\mathbf{x})$ posed in \hyperref[sec:linked-3-1]{Section 3.1}. Observation-position search is then explicitly formulated as
\begin{equation}
\mathbf{x}^{\star}
=
\underset{\mathbf{x}\in\Omega}{\operatorname{arg\,max}}\;
J_{\mathcal{T}}(\mathbf{x})
=
\underset{\mathbf{x}\in\Omega}{\operatorname{arg\,max}}\;
\sum_{j=1}^{M}
T_j(\mathbf{x})
F_{j,L}^{(\mathcal{T})}
\bigl(\mathbf{u}_j(\mathbf{x})\bigr).
\label{eq:search-full-scene}
\end{equation}

As in the single-object examples, this objective remains continuously differentiable over $\Omega$. This completes the construction of InfoField3D, enabling continuous optimization of task-conditioned observation positions with analytic gradients. Further extensions, including a formulation for frequent task queries, are presented in Appendix~\ref{app:method-details}.

\section{Experimental Evaluation}
\label{sec:linked-4}

The experiments evaluate representation fidelity, optimization quality and efficiency, viewpoint robustness, and downstream effects on a frozen VLA policy. Detailed experimental and implementation settings are provided in Appendices~\ref{app:representation-fidelity}, \ref{app:viewpoint-search}, \ref{app:viewpoint-robustness}, and~\ref{app:controlled-vla}.

\subsection{Representation Fidelity and Compression}
\label{sec:linked-4-1}

\textbf{Experimental setup.} This section evaluates how faithfully InfoField3D's analytic field reproduces pointwise computation on the original point cloud. We use 1,655 static scenes from LIBERO and LIBERO-Plus \citep{liu2023libero,fei2026liberoplus}, with approximately 800 uniformly sampled test positions per scene, under both uniform- and task-relevance maps. At each position, InfoField3D is compared with pointwise computation using exact occlusion. \hyperref[tab:master-1]{Table 1} reports the results.

\begin{table}[!tb]
\centering
\caption{Mean fidelity of InfoField3D relative to pointwise spatial information responses.}
\label{tab:master-1}
\small
\resizebox{0.95\textwidth}{!}{%
\begin{tabular}{cccc}
\toprule
Information setting & Optimal-value attainment ↑ & Shape consistency ↑ & Ranking consistency ↑ \\
\midrule
Uniform relevance map & 97.93\% & 92.28\% & 90.51\% \\
Task-relevance map & 97.97\% & 93.86\% & 91.29\% \\
\bottomrule
\end{tabular}
}
\end{table}

\textbf{Fidelity metrics.} Optimal-value attainment measures how much of the best pointwise-reference value in the candidate set is attained at the field-selected position, emphasizing the preservation of high-value solutions most relevant to observation-position selection. Shape consistency is the Pearson correlation between response values, and ranking consistency is the correlation between their position rankings. All metrics are computed per scene and then averaged across scenes.

\textbf{Results.} As \hyperref[tab:master-1]{Table 1} shows, the representation retains high spatial fidelity under both relevance maps and, notably, uses only about 1/15 of the parameter budget of the original point cloud. To stress-test this compression, we densify the point clouds to 2×, 4×, and 16× while keeping the field budget fixed. Even at 16× density, with a point-cloud-to-field parameter ratio of about 239:1, all three metrics decrease by less than 0.4 pp, further supporting the low-complexity hypothesis in \hyperref[sec:linked-3-2]{Section 3.2}.

\subsection{Quality and Efficiency of Observation-Position Optimization}
\label{sec:linked-4-2}

\begin{figure*}[!t]
\centering

\begin{minipage}[t]{0.61\textwidth}
    \vspace{0pt}
    \centering

    \includegraphics[width=\linewidth]{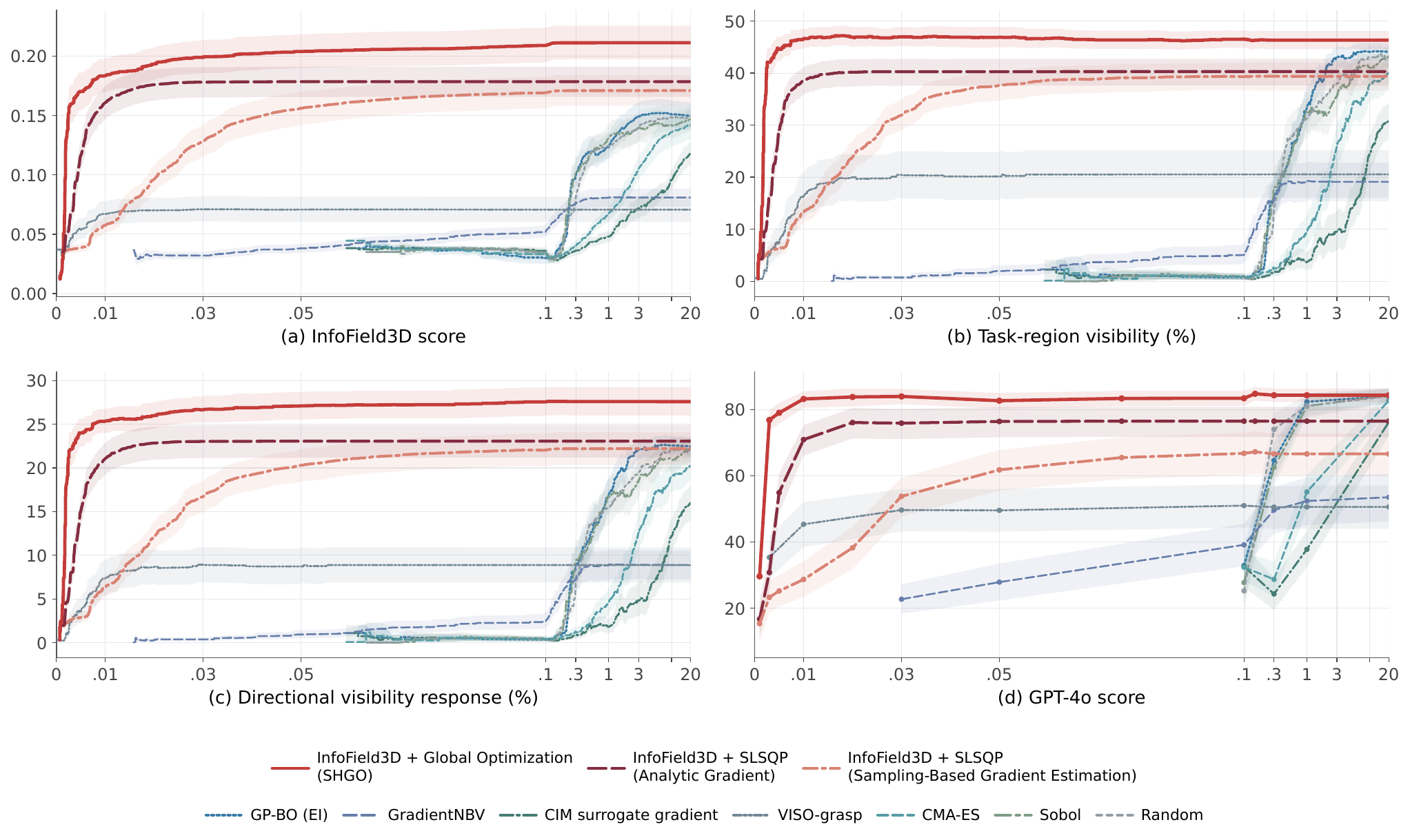}

    \vspace{-1mm}
    \caption{
Viewpoint scores of different methods versus search time. Each panel reports the metric indicated by its title on the vertical axis, with search time (s) on the horizontal axis. The first three curves optimize the same InfoField3D objective; the two SLSQP variants differ only in gradient source.
}
    \label{fig:quality-time-curves}
\end{minipage}%
\hspace{0.015\textwidth}%
\begin{minipage}[t]{0.37\textwidth}
    \vspace{0pt}
    \centering

    \begin{minipage}[c]{0.88\linewidth}
        \centering
        \includegraphics[width=0.97\linewidth]{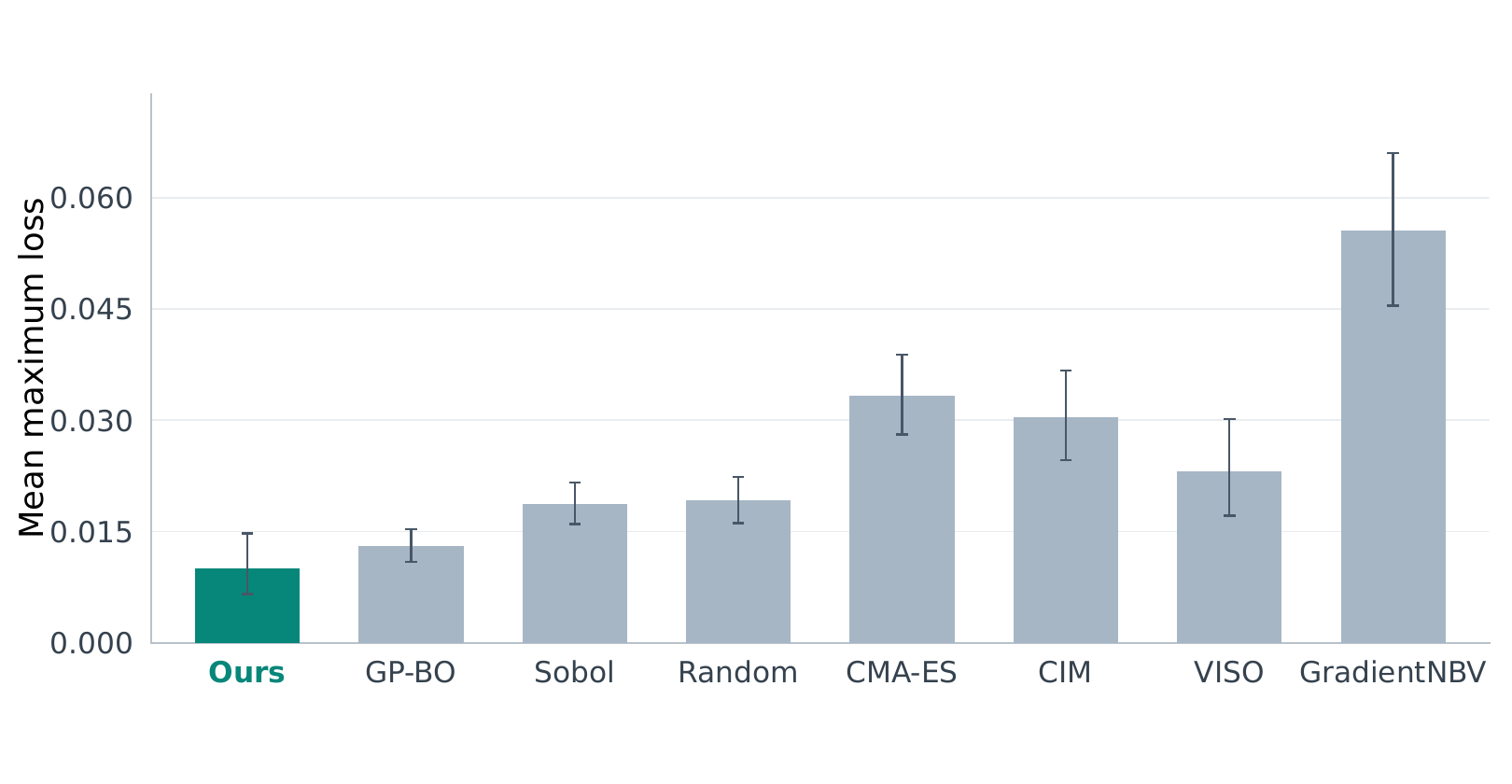}
    \end{minipage}%
    \begin{minipage}[c]{0.08\linewidth}
        \centering
        {\footnotesize (a)}
    \end{minipage}

    \vspace{0.5mm}

    \begin{minipage}[c]{0.88\linewidth}
        \centering
        \includegraphics[width=0.64\linewidth]{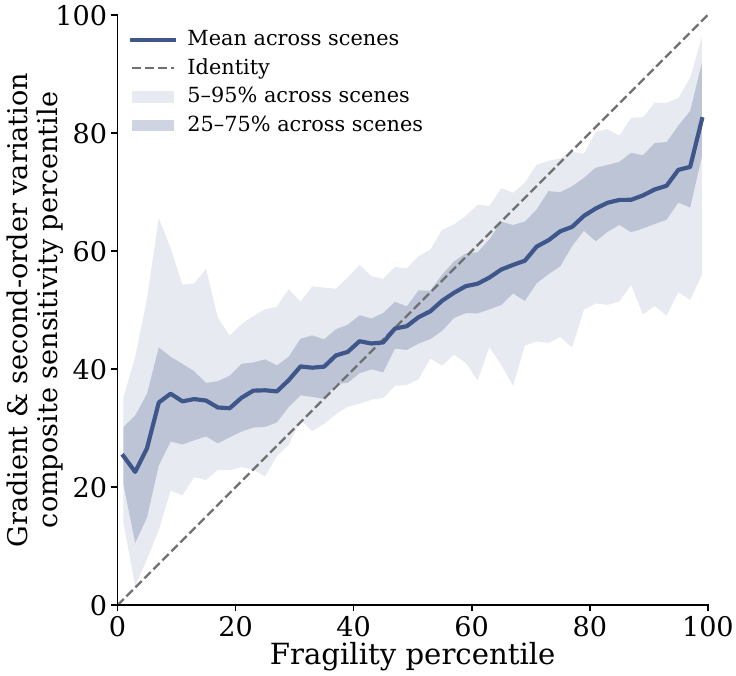}
    \end{minipage}%
    \begin{minipage}[c]{0.08\linewidth}
        \centering
        {\footnotesize (b)}
    \end{minipage}

    \vspace{-1mm}
\caption{
Fragility analysis: (a) perturbation loss of each method’s recommended viewpoints; (b) Correlation analysis between InfoField3D’s multi-order derivatives and fragility.
}
    \label{fig:fragility-analysis}
\end{minipage}

\vspace{-5mm}
\end{figure*}

\textbf{Experimental setup.} This section compares the quality and efficiency of observation-position optimization across methods. We evaluate all methods on 80 LIBERO-Plus object-layout scenes (20 per suite), with 10 runs per scene under the same evaluation protocol. The result and time cost of each iteration are recorded. Results are first averaged within each scene and then averaged equally across scenes. All methods search over the same feasible domain within a cubic 3D workspace.

In the experiments, the task-relevant target regions and their corresponding weights are generated by a fully automated multimodal annotation pipeline \citep{kirillov2023sam,radford2021clip,cherti2023openclip} and shared across all methods, thereby controlling for potential effects arising from differences in task interpretation.
Viewpoint quality is evaluated from four complementary aspects.
Task-region visibility \textbf{$V(\mathbf{x})$} measures the relevance-weighted fraction of the task-relevant surface that is visible.
Directional visibility response further weights visible task points by how directly their surfaces face the viewpoint. 
Although InfoField3D is the method proposed in this work, we believe its viewpoint score is still informative because it jointly considers viewing orientation, region exposure, and visibility margin. We therefore also report it as an auxiliary reference metric. In addition to these rule-based scoring metrics, we also use a multimodal model (GPT-4o), which scores each viewpoint based on the given task description and its rendered image, to provide an auxiliary evaluation perspective informed by the model's broad semantic and commonsense priors.

\begin{table}[!b]
\centering
\caption{Viewpoint quality under different search budgets.}
\label{tab:viewpoint-efficiency}
\small
\setlength{\tabcolsep}{2.5pt}
\renewcommand{\arraystretch}{1.10}
\resizebox{0.95\textwidth}{!}{%
\begin{tabular}{@{}l*{8}{>{\raggedleft\arraybackslash}p{3.3em}}@{}}
\toprule
& \multicolumn{2}{c}{InfoField score $\uparrow$}
& \multicolumn{2}{c}{Task vis. (\%) $\uparrow$}
& \multicolumn{2}{c}{Dir. resp. (\%) $\uparrow$}
& \multicolumn{2}{c}{GPT-4o score $\uparrow$} \\
\cmidrule(lr){2-3}\cmidrule(lr){4-5}\cmidrule(lr){6-7}\cmidrule(lr){8-9}
Method & 0.1 s & 20 s & 0.1 s & 20 s & 0.1 s & 20 s & 0.1 s & 20 s \\
\midrule
InfoField3D (SHGO) & \textbf{\textit{0.2091}} & \textbf{\textit{0.2113}} & \textbf{\textit{46.51}} & \textbf{\textit{46.32}} & \textbf{\textit{27.59}} & \textbf{\textit{27.58}} & \textbf{\textit{83.42}} & 84.33 \\
InfoField3D (Analytic Gradient) & 0.1786 & 0.1786 & 40.30 & 40.30 & 23.06 & 23.06 & 76.50 & 76.50 \\
InfoField3D (Sampling-Based Gradient) & 0.1691 & 0.1710 & 39.29 & 39.37 & 22.02 & 22.20 & 66.77 & 66.60 \\
GP-BO & 0.0299 & 0.1497 & 0.71 & 44.12 & 0.34 & 22.46 & 32.89 & 84.10 \\
CIM surrogate gradient & 0.0360 & 0.1177 & 0.93 & 30.77 & 0.45 & 15.98 & 32.61 & 76.19 \\
CMA-ES & 0.0333 & 0.1419 & 0.87 & 40.01 & 0.42 & 20.24 & 32.36 & 82.73 \\
GradientNBV & 0.0526 & 0.0809 & 5.22 & 19.11 & 2.46 & 8.89 & 39.13 & 53.50 \\
VISO-grasp & 0.0708 & 0.0707 & 20.53 & 20.54 & 8.87 & 8.87 & 50.98 & 50.58 \\
Sobol & 0.0338 & 0.1470 & 0.78 & 43.37 & 0.38 & 22.21 & 27.74 & \textbf{\textit{84.40}} \\
Random & 0.0332 & 0.1497 & 0.73 & 43.20 & 0.35 & 22.17 & 25.22 & 84.04 \\
\bottomrule
\end{tabular}%
}
\end{table}

\textbf{Optimization variants.} For our method, once InfoField3D is constructed, view planning reduces to constrained optimization over the feasible domain, allowing standard optimization methods to be directly applied. We can therefore directly use an existing off-the-shelf optimizer to solve it. In our experiments, we simply use the SHGO solver \citep{endres2018shgo} provided by SciPy \citep{virtanen2020scipy}, supplying our analytic gradients to its local optimizer. In addition, to isolate the benefit of analytic differentiation, we adopt a standard gradient-based constrained optimizer (SLSQP)  \citep{kraft1994tomp}  on the same InfoField3D field, supplying either the analytic gradient or a sampling-based gradient estimate while keeping all other settings fixed. 

\textbf{Baselines.} We further compare Sobol-sampled global search \citep{sobol1967distribution}, uniform random search, CMA-ES \citep{hansen2001cmaes}, Bayesian optimization with expected improvement (GP-BO) \citep{jones1998ego}, GradientNBV \citep{burusa2024gradientnbv}, a CIM-style surrogate-gradient method \citep{wang2024cim}, and VISO-Grasp-style analytic control \citep{shi2025visograsp}. Learned viewpoint predictors are omitted because reproducible alternatives use substantially different observation or action spaces, while methods under more comparable settings do not provide released models. Some methods, such as FIF \citep{zhang2020fif} and GO-PRE \citep{song2026gopre}, define information utility through localization observability and predictive rendering uncertainty, respectively, and can also be viewed, to some extent, as addressing camera placement. However, transferring them to the task considered in this experiment would require additional, unvalidated modeling choices. They are therefore excluded from the comparison to avoid conflating adaptation-induced errors with the capabilities of the original methods.

\textbf{Results.} As shown in Figure~\ref{fig:quality-time-curves} and
Table~\ref{tab:viewpoint-efficiency}, InfoField3D optimized with SHGO shows a pronounced efficiency advantage, reaching high viewpoint quality within the first 0.1s of search time while remaining among the best-performing methods in final viewpoint quality across the evaluated metrics. Notably, its advantage extends beyond exposing more of the task-relevant region: it also achieves a markedly stronger directional visibility response, indicating that the visible task surfaces are observed more frontally. This behavior is consistent with InfoField3D's explicit modeling of directional information response. Separately, when optimizing the same InfoField3D field with SLSQP, the analytic gradient reaches high-quality solutions substantially earlier than the gradient estimated from six local samples, while the SHGO variant remains the strongest overall, showing that InfoField3D benefits both from an optimization-friendly field and from exact derivatives. 

\subsection{Robustness of Recommended Observation Positions}
\label{sec:linked-4-3}

The preceding experiments evaluate how much task-relevant information a viewpoint can provide. However, practical deployment also requires robustness to positioning errors and environmental disturbances. If a viewpoint's observation ability collapses under slight perturbation, it is not robust---for example, when the line of sight threads a narrow gap between occluders or the target has only just emerged from behind an occluder's edge. This section evaluates whether the viewpoints recommended by each method remain reliable under such perturbations.

\textbf{Perturbation tests.} We define the \emph{fragility} of a viewpoint, the opposite of robustness, as the worst-case information loss under small spatial perturbations, where information is quantified by the score $V(\mathbf{x})$ defined in \hyperref[sec:linked-4-2]{Section 4.2}. We perturb the final viewpoints recommended by each method and record their maximal loss in task-region visibility. As shown in Figure~\ref{fig:fragility-analysis}(a), InfoField3D demonstrates advantages in the robustness of its final recommended viewpoints. 

\textbf{Derivative-based analysis.} Explicit perturbation tests can assess viewpoint robustness but incur additional computational overhead. 
Since fragile viewpoints often correspond to rapid information changes within their local neighborhoods, it is natural to seek local sensitivity signals from derivatives and their higher-order variations. As a concrete example, we sum first-order gradients with second-order information into a composite signal for analysis. As shown in Figure~\ref{fig:fragility-analysis}(b), this signal exhibits a certain degree of correlation with external fragility, suggesting that local derivative information of the field may provide useful auxiliary cues for fragility analysis.

\subsection{Controlled Downstream Evaluation with a Frozen VLA}
\label{sec:linked-4-4}

This section asks, under a frozen VLA policy ($\pi$0.5 \citep{black2025pi05} in our experiments), whether InfoField3D's viewpoint optimization can improve task execution.

\begin{figure*}[!tb]
\centering

\begin{minipage}[t]{0.34\textwidth}
    \centering
    \includegraphics[width=0.95\linewidth]  {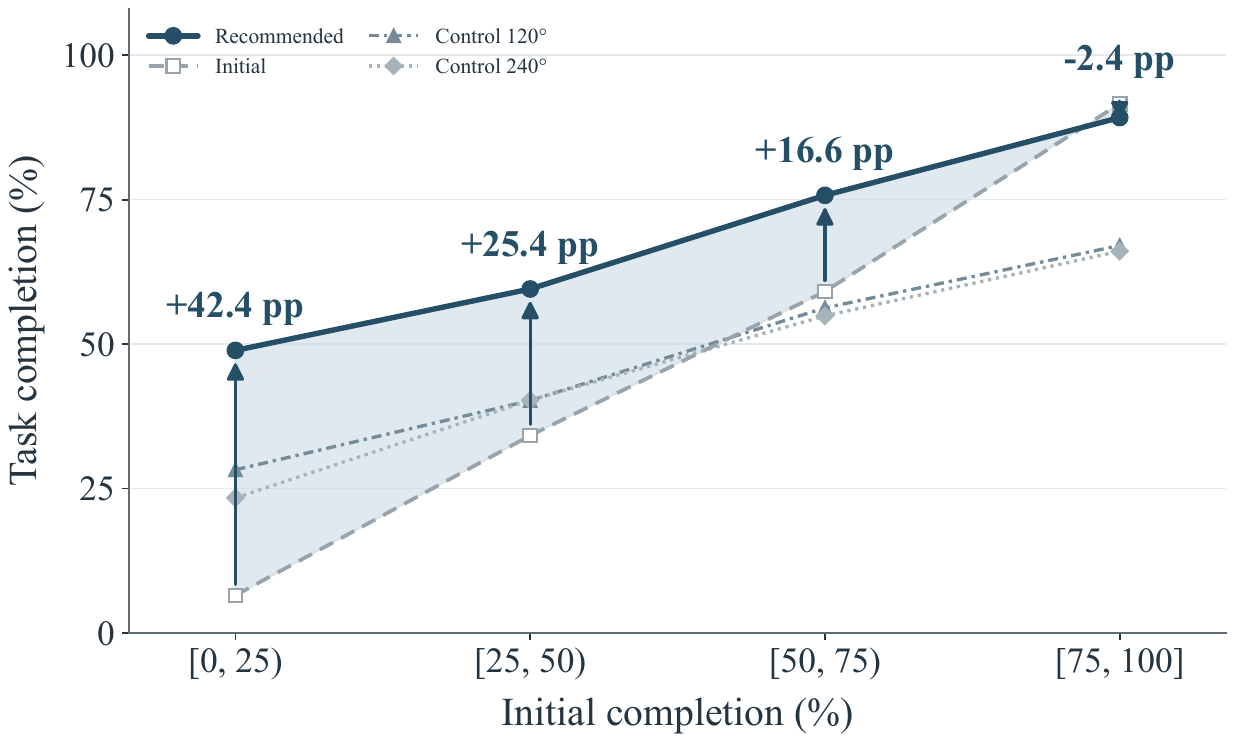}
    \par\vspace{0mm}
    {\footnotesize (A.1)}
\end{minipage}%
\hspace{0.012\textwidth}%
\begin{minipage}[t]{0.62\textwidth}
    \centering
    \includegraphics[width=\linewidth]{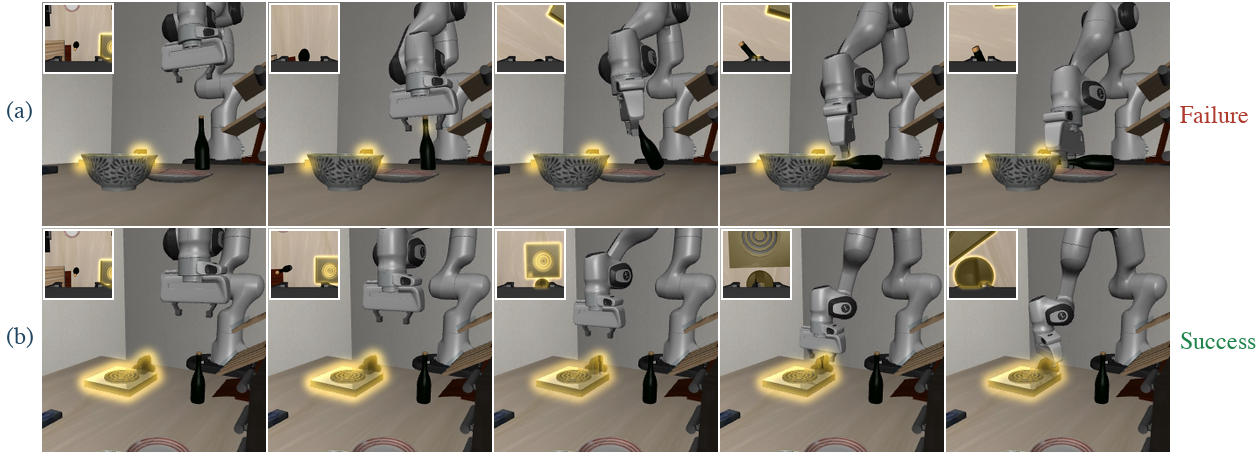}
    \par\vspace{0mm}
    {\footnotesize (B.1)}
\end{minipage}

\begin{minipage}[t]{0.34\textwidth}
    \centering
    \includegraphics[width=0.79\linewidth]
    {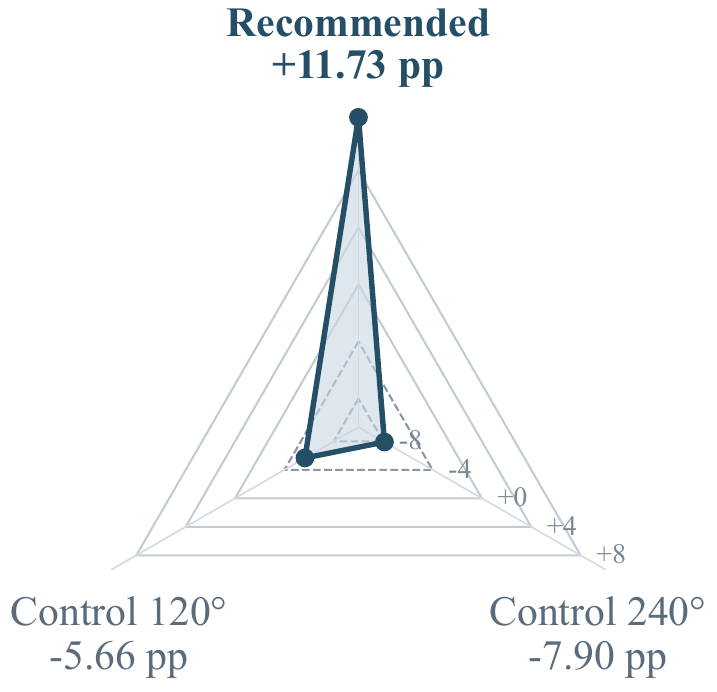}
    \par\vspace{0mm}
    {\footnotesize (A.2)}
\end{minipage}%
\hspace{0.012\textwidth}%
\begin{minipage}[t]{0.62\textwidth}
    \centering
    \includegraphics[width=\linewidth]{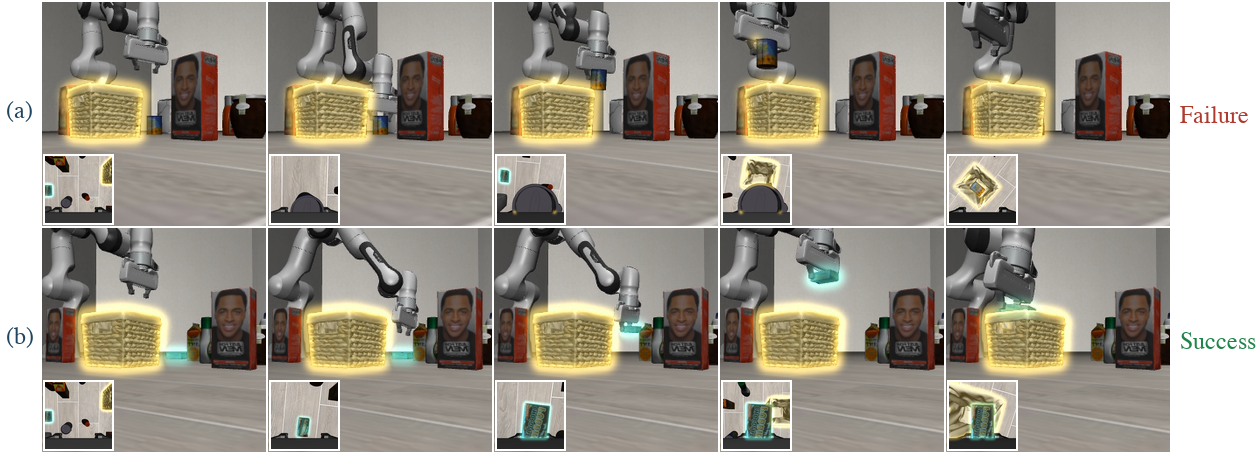}
    \par\vspace{0mm}
    {\footnotesize (B.2)}
\end{minipage}

\vspace{-2mm}

\caption{
Trident-probe outcomes and representative VLA rescue episodes.
(A.1) Mean task completion at the initial, recommended, and control viewpoints, stratified by initial-completion tier.
(A.2) Mean completion change from the trapped viewpoint for the recommendation and two equal-distance controls.
In (B.1--B.2), rows (a) and (b) show rollouts from trapped and recommended third-person viewpoints, respectively, with identical initial wrist-camera states; highlights mark task-relevant objects.
}
\label{fig:trident-and-vla-cases}

\vspace{-5mm}
\end{figure*}

\textbf{Experimental setup.} Whether a VLA policy benefits from a viewpoint depends not only on observation quality but also on camera-pose distribution shift. A global comparison over widely separated viewpoints would therefore confound these effects. We instead use a local neighborhood-rescue protocol: from a third-person viewpoint with limited access to task-relevant information, InfoField3D proposes a nearby viewpoint with improved task information, requiring only a small displacement from the starting viewpoint. Because the two viewpoints remain very close in pose space, their proximity helps reduce camera-pose distribution shift, allowing the effect of improved information accessibility to be examined more directly. The displacement to the recommendation is rotated by $\pm120^{\circ}$ to form two equal-length controls, yielding a trident probe. Each endpoint is evaluated with five frozen-VLA rollouts using different seeds. Across 80 LIBERO-Plus scenes, the above protocol yields 697 information-limited viewpoints, for which the corresponding probe groups are constructed. Quantitative outcomes and representative VLA episodes are shown in Figure~\ref{fig:trident-and-vla-cases}.

 \textbf{Results.} Figure~\ref{fig:trident-and-vla-cases}(A.1) shows that the benefit is largest when initial completion is low ($+42.4$ pp for the 0--25\% tier) and diminishes as initial completion increases. When completion is already high (75--100\%), the recommendation produces a small decrease ($-2.4$ pp), suggesting that viewpoint optimization is most beneficial when observation is a limiting factor. Figure~\ref{fig:trident-and-vla-cases}(A.2) shows that InfoField3D's recommendation improves task completion from the trapped viewpoint by $+11.73$ pp on average, substantially more than the two equal-distance controls. Figures~\ref{fig:trident-and-vla-cases}(B.1--B.2) illustrate representative cases in which moving from a trapped to the recommended viewpoint changes VLA execution from failure to success.

\section{Conclusion and Limitations}
\label{sec:linked-5}

This work treats task-conditioned observation-position selection as optimization over an explicit information landscape in free space rather than repeated evaluation of isolated viewpoints. InfoField3D reorganizes geometry-bound scene information into a differentiable analytic field whose value measures task-relevant information accessibility and whose spatial gradient directly guides viewpoint improvement. Experiments show that the field preserves the underlying information landscape, enables efficient continuous optimization, favors less fragile observation positions, and can lead to downstream improvements for a frozen VLA policy under information-limited initial viewpoints. Together, these results suggest that InfoField3D, as an explicit and differentiable organization of scene information over free space, provides a useful interface between semantic scene understanding and task-oriented view planning.

The current formulation assumes a globally reconstructed, static scene. Future work may extend it to partially observed environments, dynamically changing scenes, and information beyond semantics. More broadly, this suggests a path toward treating diverse geometry-bound sensing objectives within a unified differentiable field formulation.

\subsection*{AI use statement}

In this work, we used generative AI tools to assist with manuscript writing and language polishing, including improving clarity, readability, and grammar. We also used GPT-4o for multimodal viewpoint scoring and task-relevance annotation in the experiments described in Section 4.2 and Appendix C. We have reviewed and revised all AI-assisted text to ensure its accuracy and consistency with our research. We take responsibility for the final content of this work, including all text, claims, and artifacts produced with the aid of generative AI.

\FloatBarrier
\bibliographystyle{iclr2027_conference}
\makeatletter
\newcommand{\InfoFieldRevisionBibliography}{%
  \let\InfoFieldOriginalBibitem\@lbibitem
  \def\InfoFieldNewBibkey{kerbl2023gaussian}%
  \def\@lbibitem[##1]##2{%
    \InfoFieldOriginalBibitem[##1]{##2}%
    \def\InfoFieldCurrentBibkey{##2}%
    \ifx\InfoFieldCurrentBibkey\InfoFieldNewBibkey
      \color{blue}%
    \else
      \normalcolor
    \fi
    \ignorespaces
  }%
}
\AddToHookNext{env/thebibliography/begin}{\InfoFieldRevisionBibliography}
\makeatother
\bibliography{references}

\appendix
\clearpage
\begingroup
\setcounter{secnumdepth}{3}
\renewcommand{\thesection}{\Alph{section}}
\renewcommand{\thesubsection}{\thesection.\arabic{subsection}}
\renewcommand{\thesubsubsection}{\thesubsection.\arabic{subsubsection}}
\counterwithin{figure}{section}
\counterwithin{table}{section}
\renewcommand{\thefigure}{\thesection\arabic{figure}}
\renewcommand{\thetable}{\thesection\arabic{table}}
\renewcommand{\theHsection}{appendix.\Alph{section}}
\renewcommand{\theHsubsection}{\theHsection.\arabic{subsection}}
\renewcommand{\theHsubsubsection}{\theHsubsection.\arabic{subsubsection}}
\renewcommand{\theHfigure}{appendix.\thesection.\arabic{figure}}
\renewcommand{\theHtable}{appendix.\thesection.\arabic{table}}
\captionsetup{labelfont=bf,labelsep=period}
\makeatletter
\def\@seccntformat#1{%
  \ifcsname app@heading@#1\endcsname
    \csname app@heading@#1\endcsname
  \else
    \csname the#1\endcsname\quad
  \fi
}
\def\app@heading@section{Appendix~\thesection:\quad}
\makeatother

\raggedbottom
\setlength{\parindent}{0pt}
\setlength{\parskip}{0.55em}
\renewcommand{\arraystretch}{1.15}

\begin{center}
{\LARGE\bfseries Appendix}\par
\vspace{0.4em}
\end{center}

\section*{Contents}

\begingroup
\setlength{\parindent}{0pt}
\setlength{\parskip}{0pt}
\setlength{\tabcolsep}{0pt}
\setlength{\LTleft}{0pt}
\setlength{\LTright}{0pt}
\setlength{\LTpre}{0pt}
\setlength{\LTpost}{0pt}
\renewcommand{\arraystretch}{1}
\fontsize{10}{12}\selectfont

\makeatletter
\@for\AppTOCLabel:=app:design-analysis,app:component-ablation,app:constraint-complexity,%
app:self-occlusion,app:cost-and-vla,app:flops-memory,%
app:vla-occluder,app:task-limits-errors,app:small-critical-regions,%
app:score-execution-gap,app:approximation-limits,app:challenging-geometries,%
app:parameter-sensitivity,app:semantic-errors,app:parameter-settings\do{%
  \@namedef{AppTOCBlue:\AppTOCLabel}{}%
}
\makeatother
\newcommand{\AppTOCColor}[1]{%
  \ifcsname AppTOCBlue:#1\endcsname\leavevmode\color{blue}\fi
}

\newcommand{\AppTOCChapter}[1]{%
  \multicolumn{2}{@{}p{\textwidth}@{}}{%
    \fontsize{11}{13}\selectfont\bfseries
    \AppTOCColor{#1}\hyperref[#1]{\ref*{#1}.\quad\nameref*{#1}}%
  }\\*[4pt]
  \midrule[0.4pt]
}

\newcommand{\AppTOCEntry}[2]{%
  \AppTOCColor{#1}\textbf{\hyperref[#1]{\ref*{#1}}} &
  \AppTOCColor{#1}\textbf{\hyperref[#1]{\nameref*{#1}}}\par
  {\fontsize{9}{11}\selectfont #2\par}
  \\[5pt]
}

\newcommand{\AppTOCSubentry}[2]{%
  \hspace*{0.65em}\AppTOCColor{#1}\hyperref[#1]{\ref*{#1}} &
  \AppTOCColor{#1}\textbf{\hyperref[#1]{\nameref*{#1}}}\par
  {\fontsize{9}{11}\selectfont #2\par}
  \\[5pt]
}

\begin{longtable}{
  @{}
  >{\raggedright\arraybackslash}p{0.15\textwidth}
  >{\raggedright\arraybackslash}p{0.85\textwidth}
  @{}
}

\endfirsthead

\multicolumn{2}{@{}l@{}}{%
  \fontsize{11}{13}\selectfont\bfseries
  Contents (continued)
}\\[8pt]
\endhead

\endfoot
\endlastfoot

\AppTOCChapter{app:method-details}

\AppTOCEntry{app:field-construction}
{Construction choices and the approximations introduced at each representation step.}

\AppTOCSubentry{app:area-weighting}
{Area weights compensate for nonuniform sampling of the same surface.}

\AppTOCSubentry{app:coverage-normalization}
{An optional readout measures task response relative to geometric exposure.}

\AppTOCSubentry{app:harmonic-degree-error}
{Randomized spherical patterns quantify approximation error and guide degree selection.}

\AppTOCSubentry{app:reference-sphere-radius}
{Reference-sphere radius controls mapping accuracy at different observation distances.}

\AppTOCSubentry{app:geometric-partitioning}
{Shape-dependent mapping loss decreases as objects are partitioned into local units.}

\AppTOCSubentry{app:gaussian-occupancy}
{Gaussian occupancy, analytic path integrals and gradients, and material transparency.}

\AppTOCSubentry{app:gaussian-occlusion-illustrations}
{Field values and gradient trajectories under single and multiple occluders.}

\AppTOCEntry{app:vector-readout}
{Reusable vector coefficients support repeated queries through linear readout.}

\AppTOCEntry{app:multiple-targets}
{Separate target responses are combined to express joint observation requirements.}

\addlinespace[8pt]

\AppTOCChapter{app:representation-fidelity}

\AppTOCEntry{app:fidelity-experiment}
{Pointwise reference computation, fidelity metrics, and representation-size comparison.}

\AppTOCEntry{app:density-stress-test}
{Fidelity under increased point-cloud density with a fixed computational-unit budget.}

\addlinespace[8pt]

\AppTOCChapter{app:viewpoint-search}

\AppTOCEntry{app:search-scenes-domain}
{Scene selection and the geometric constraints defining valid observation positions.}

\AppTOCEntry{app:search-baselines}
{Compared objectives, implementation sources, and adaptations to the shared search task.}

\AppTOCEntry{app:search-metrics-results}
{Definitions of evaluation metrics and detailed numerical results for the main experiments.}

\AppTOCEntry{app:gpt4o-evaluation}
{Reference-image design, task-specific scoring and aggregation, and repeated evaluations.}

\AppTOCEntry{app:semantic-annotation}
{Task instructions and multiview images produce relevance weights on 3D surfaces.}

\AppTOCEntry{app:preprocessing-cost}
{Initial construction cost, task-update cost, and search efficiency including preprocessing.}

\addlinespace[8pt]

\AppTOCChapter{app:viewpoint-robustness}

\multicolumn{2}{@{}p{\textwidth}@{}}{%
  \fontsize{9}{11}\selectfont
  Perturbation protocol and task-region visibility losses
  at five displacement scales.
}\\[5pt]

\addlinespace[8pt]

\AppTOCChapter{app:controlled-vla}

\AppTOCEntry{app:vla-experiment-design}
{Local viewpoint rescue, equal-distance controls, frozen-policy rollouts, and completion scoring.}

\AppTOCEntry{app:vla-execution-examples}
{Paired execution examples from initial and recommended third-person viewpoints.}

\addlinespace[8pt]

\AppTOCChapter{app:design-analysis}

\AppTOCEntry{app:component-ablation}
{Contributions of continuous occlusion modeling, spherical harmonic encoding, and local geometric partitioning.}

\AppTOCEntry{app:constraint-complexity}
{Search cost under increasing numbers of keep-out zones, reduced feasible volume, and disconnected feasible regions.}

\AppTOCEntry{app:self-occlusion}
{Local-target and whole-object experiments examining self-occlusion between units of the same object.}

\addlinespace[8pt]

\AppTOCChapter{app:cost-and-vla}

\AppTOCEntry{app:flops-memory}
{Per-query FLOPs, peak memory, and viewpoint quality under cumulative FLOPs budgets.}

\AppTOCEntry{app:vla-occluder}
{Frozen-VLA evaluation from the default camera viewpoint with an added occluding panel.}

\addlinespace[8pt]

\AppTOCChapter{app:task-limits-errors}

\AppTOCEntry{app:small-critical-regions}
{Recommended viewpoints for different small target regions and their weighted combination.}

\AppTOCEntry{app:score-execution-gap}
{Representative failure cases and analysis of when improved observation does not improve execution.}

\AppTOCEntry{app:approximation-limits}
{Operating limits of the approximations and sensitivity to construction parameters.}

\AppTOCSubentry{app:challenging-geometries}
{Narrow gaps, thin shells, concave structures, and spatially concentrated targets.}

\AppTOCSubentry{app:parameter-sensitivity}
{Sensitivity to the number of units, minimum Gaussian width, and attenuation strength.}

\AppTOCEntry{app:semantic-errors}
{Viewpoint-quality loss under semantic noise, spill, and erosion at three perturbation strengths.}

\AppTOCEntry{app:parameter-settings}
{Partitioning and occlusion settings, including the shared parameter-selection rules across scenes.}

\end{longtable}
\endgroup

\clearpage
\section{Method Details and Extensions}
\label{app:method-details}

\subsection{Field Construction Details and Analysis}
\label{app:field-construction}

\subsubsection{Area-Weighting Treatment for Nonuniform Point Clouds}
\label{app:area-weighting}

The input point cloud consists of surface points $\mathbf p_i$, their associated unit outward normals $\mathbf n_i$, and semantic features $\mathbf f_i$. A task query produces point-level relevance $s_i$ from the semantic features, representing the relevance of the surface at that location to the current observation requirement; the normal describes the directional response of that surface toward different observation positions.

For an approximately uniformly sampled point cloud, the surface area represented by each point is approximately the same, so the main text directly sums the contribution of every point. Under nonuniform sampling, surface regions of the same size may contain different numbers of points. If equal geometric weight is still assigned to each point, densely sampled regions receive a larger accumulated contribution even when their task relevance has not increased. Without resampling the point cloud, this sampling-density bias can be reduced by introducing, for each point, the surface area $a_i$ that it represents and writing the weight used in response accumulation as

\begin{equation}
w_i=a_i s_i.
\end{equation}

Here, $a_i$ describes how much surface the sample point represents, and $s_i$ describes the degree to which that surface is relevant to the task. When more sample points are added to the same surface, the area represented by each point should decrease correspondingly so that the total area weight of the region remains constant, preventing the response from being amplified solely by denser sampling. Under uniform sampling with $a_i=a$, area weighting and direct summation differ only by a common positive scale factor $a$, and therefore do not change the ranking of observation positions by the unnormalized response.

\subsubsection{Coverage Normalization for Large, Low-Relevance Surfaces}
\label{app:coverage-normalization}

The point-cloud sampling-density issue has been reasonably addressed in \hyperref[app:area-weighting]{Section \ref*{app:area-weighting}}, but even for uniform point clouds, the responses of large, low-relevance surfaces may still account for a substantial share of the sum, biasing the total response toward positions that expose more geometry. To reduce the influence of overall geometric coverage scale, coverage normalization may be used as an optional readout. It retains area weights and the other response-computation rules, but sets the task relevance of all surface points to 1 to obtain a geometric coverage response without task-relevance weights, and then uses this response as the normalization basis.

In this subsection, we fix the task condition $\mathcal T$ and write $J\equiv J_{\mathcal T}$ for brevity.
For the multi-unit field constructed below, the numerator and denominator use the same occupancy attenuation:

\begin{equation}
C_L(\mathbf x)=\sum_jT_j(\mathbf x)F_{j,L}^{\mathrm{coverage}}(\mathbf u_j(\mathbf x)),
\qquad
\widetilde J(\mathbf x)=\frac{J(\mathbf x)}{C_L(\mathbf x)+\varepsilon}.
\end{equation}

Here, $F_{j,L}^{\mathrm{coverage}}$ is constructed from area weights and does not include task-relevance weights, and $\varepsilon>0$ is a numerical-stability term. The normalized readout measures the task response relative to overall geometric exposure. Whether to use this method depends on whether the task should emphasize the total response or the relative degree of relevance. 

\subsubsection{Spherical Harmonic Degree and Approximation Error}
\label{app:harmonic-degree-error}

To determine how many spherical-harmonic coefficients are sufficient, we isolate the approximation of $F_{\mathrm{sph}}$ by $F_L$ defined in \hyperref[sec:linked-3-2]{Section 3.2}. We generate six families of relevance patterns directly on unit-sphere point clouds (\hyperref[fig:app-sphere-patterns]{Figure \ref*{fig:app-sphere-patterns}}), avoiding the non-spherical mapping effects analyzed in \hyperref[app:geometric-partitioning]{Appendix \ref*{app:geometric-partitioning}}. Each family is evaluated over 100 random seeds, with 1,024 uniformly sampled surface points and newly randomized patterns per seed. The six families share each seed's point cloud. We use an exponential directional kernel, evaluate a range of spherical-harmonic degrees, and compare against the direct pointwise kernel sum over a dense set of fixed spherical test directions. The implementation applies the same normalization factor to both responses, which leaves all reported metrics unchanged.

\begin{figure}[H]

\centering
\includegraphics[width=0.9\textwidth]{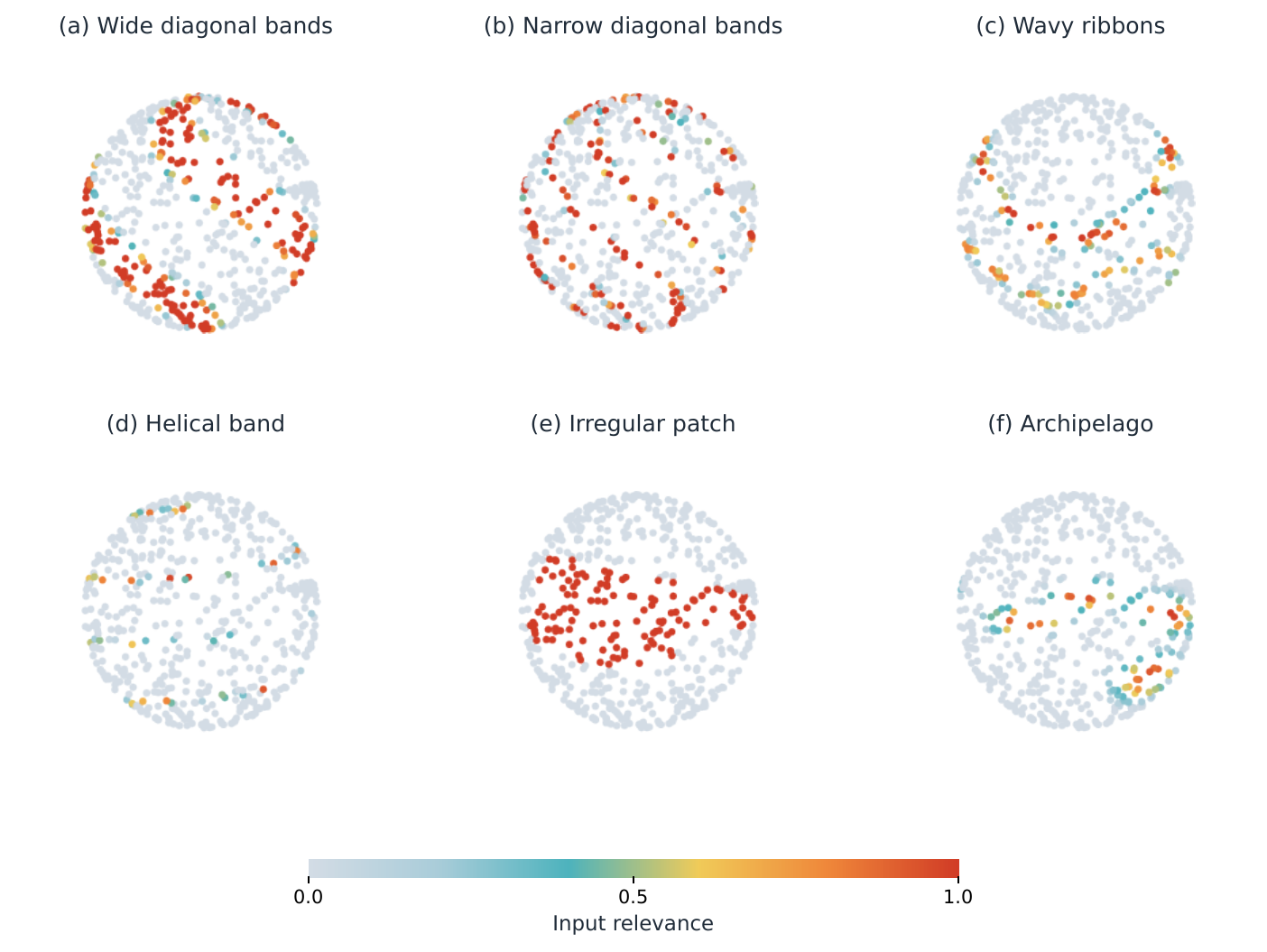}
\caption{Six randomized pattern families: wide and narrow diagonal bands, wavy ribbons, a helical band, an irregular patch, and an archipelago. }\label{fig:app-sphere-patterns}
\end{figure}

\hyperref[fig:app-harmonic-degree-errors]{Figure \ref*{fig:app-harmonic-degree-errors}} reports relative $L_2$ error before and after separately removing each response's mean, peak-direction error, and top-10\% direction-set overlap. The mean-removed error measures directional variation. Lines average 100 trials per family, and the two shaded bands are pointwise 90\% and 95\% bootstrap confidence intervals for the means.

\begin{figure}[H]

\centering
\includegraphics[width=0.95\textwidth]{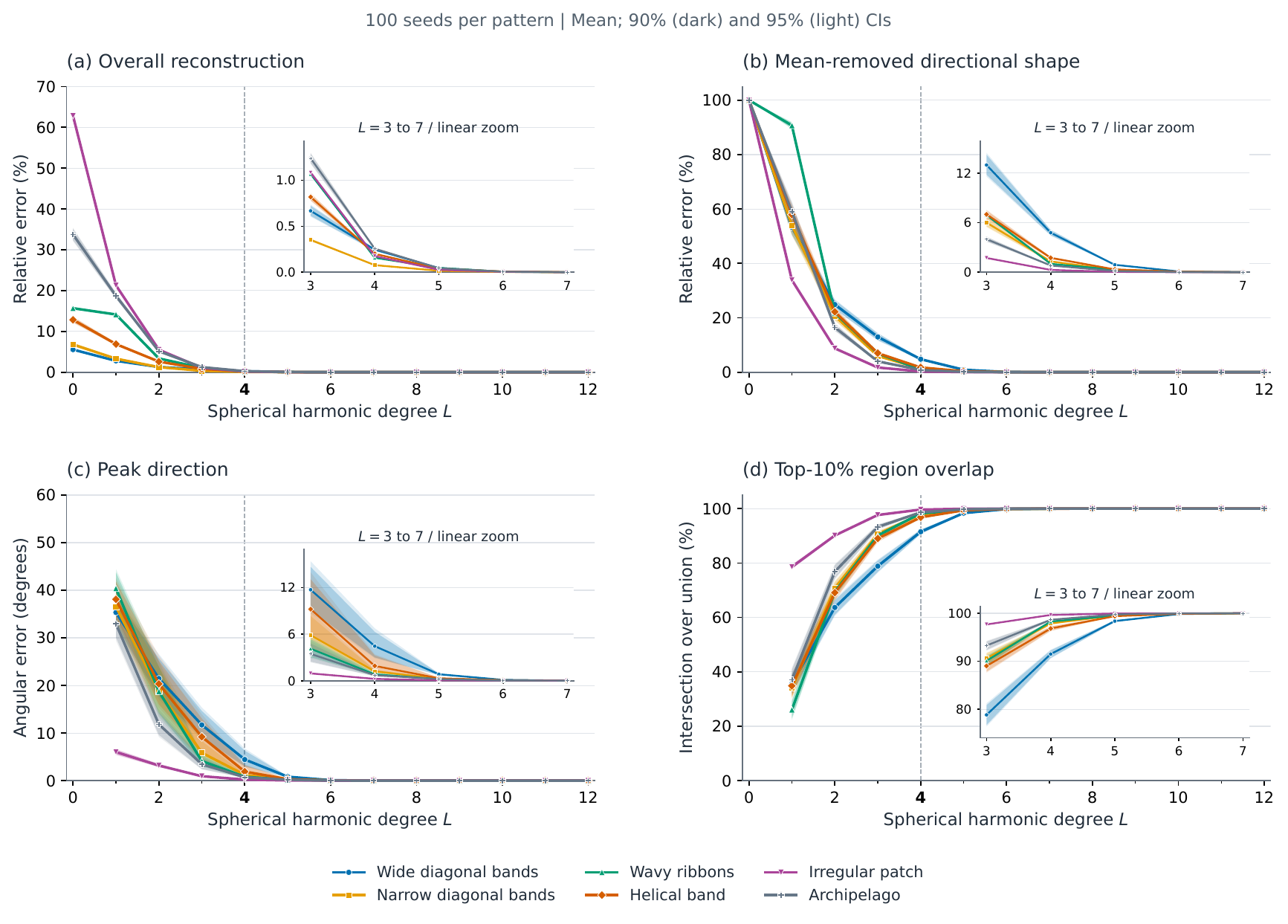}
\caption{Approximation quality versus spherical-harmonic degree. Dark inner and light outer bands indicate 90\% and 95\% confidence intervals for the means. Insets use linear axes. The dashed line marks $L=4$; peak and overlap metrics are undefined at $L=0$ and omitted.}\label{fig:app-harmonic-degree-errors}
\end{figure}

Most of the improvement occurs at low degrees. The largest family-mean overall error decreases from 5.51\% at $L=2$ to 1.24\% at $L=3$, 0.256\% at $L=4$, and 0.048\% at $L=5$. Peak-direction error can be less regular because distant peaks of similar height may exchange order. \hyperref[fig:app-peak-diagnostic]{Figure \ref*{fig:app-peak-diagnostic}} therefore also measures the relative loss of the direct response at the direction selected by $F_L$. At $L=4$, the largest family-mean loss by this measure is only 0.0424\%, indicating that a shift in peak location need not substantially reduce the attained response.

\begin{figure}[H]

\centering
\includegraphics[width=0.95\textwidth]{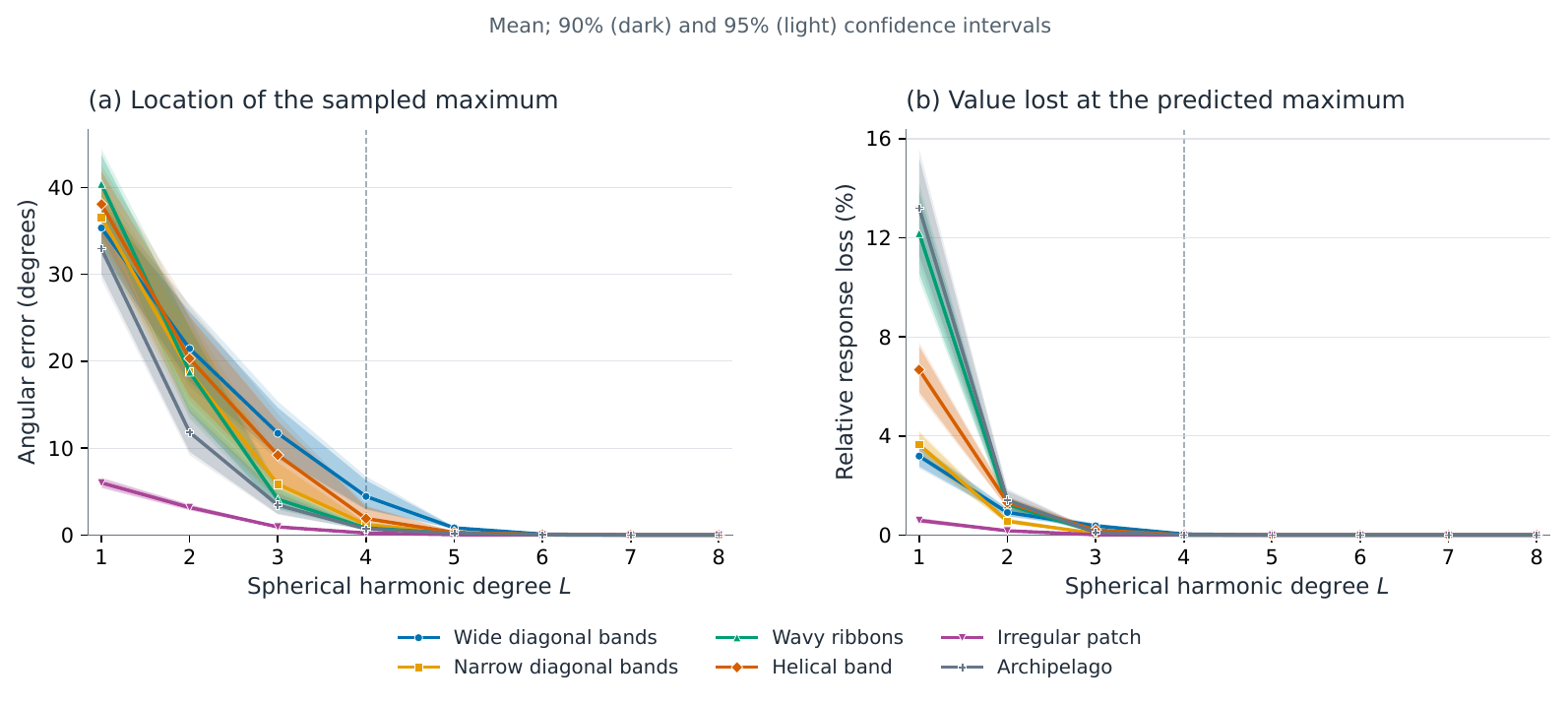}
\caption{Peak-direction error and the relative direct-response loss at the predicted maximum, evaluated on the same test directions. Lines show means with 90\% and 95\% confidence bands. Similar response values can occur at widely separated directions.}\label{fig:app-peak-diagnostic}
\end{figure}

These results suggest $L=4$ as a compact default: 25 coefficients per scalar response give at most 0.256\% overall error and 4.777\% mean-removed error when comparing the six family means. For stricter accuracy requirements, $L=5$ uses 36 coefficients and reduces these values to 0.048\% and 0.898\%; $L=6$ uses 49 coefficients and reduces them further to 0.008\% and 0.089\%. Thus, $L=5$ is a useful choice when sub-0.1\% mean overall error and sub-1\% mean directional-variation error are desired, while $L=6$ provides a higher-precision option. Balancing representation size and the small response errors observed here, we use $L=4$ in the paper's scene experiments.

In our implementation, we use orthonormal real spherical harmonics with the Condon--Shortley phase convention, with negative and positive orders corresponding to sine and cosine components, respectively. The basis is evaluated directly from unit Cartesian directions using polynomial recurrences, avoiding explicit angular coordinates and their coordinate singularities at the poles. Real-valued coefficients are obtained by area- and task-weighted projection of the mapped samples onto this basis, followed by degree-wise scaling with the angular kernel's Funk--Hecke multipliers.

\subsubsection{Mapping Reference-Sphere Radius}
\label{app:reference-sphere-radius}

\begin{figure}[H]

\centering
\includegraphics[width=0.9\textwidth]{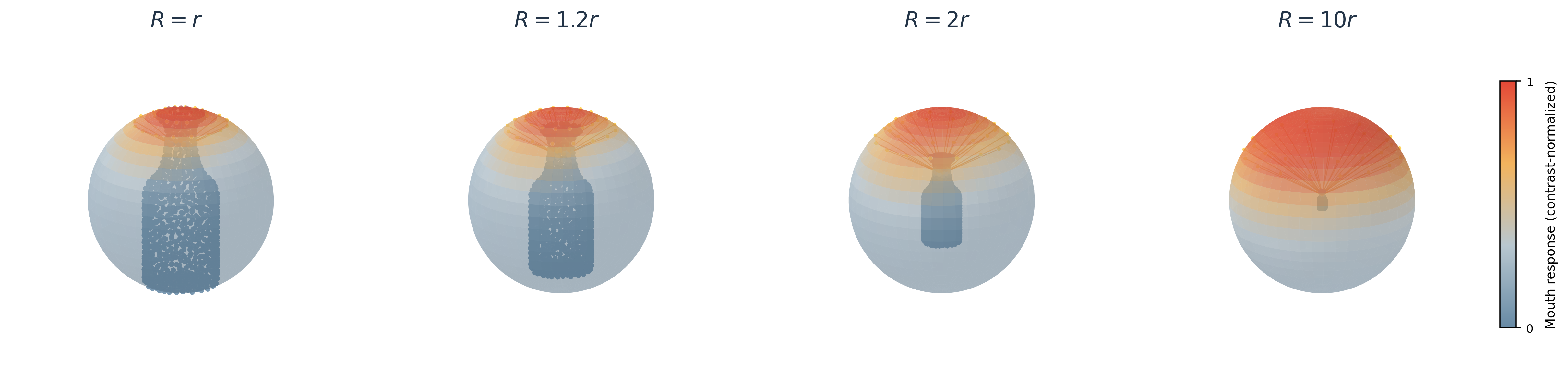}
\caption{Differences in the directional responses formed under different reference-sphere radii $R$.}\label{fig:app-reference-sphere-radius}
\end{figure}

In \hyperref[sec:linked-3-2]{Section 3.2}, we unified the directional responses of the point cloud onto a reference sphere to form an object-level directional response. As shown in \hyperref[fig:app-reference-sphere-radius]{Figure \ref*{fig:app-reference-sphere-radius}}, changing $R$ changes the mapped directional response. A smaller reference sphere and a shorter observation distance can preserve more information about the spatial distribution of the point cloud; a larger reference sphere and a longer observation distance can more fully express the directional distribution of surface normals. In experimental observations, an extremely small mapping radius, such as the radius $r$ of the minimum bounding sphere, may excessively truncate the directional distribution of surface normals and produce a large representation loss at ordinary observation distances. By contrast, when the mapping radius is between $4r$ and $10r$, a relatively small loss can usually be obtained, and the differences among values in this range are limited. This range can therefore serve as a recommended range for general scenes. As $R$ tends to infinity, the mapping result can be regarded as retaining only the distribution of surface normals.

\subsubsection{Spherical-Mapping Loss and Obtaining Computational Units through Geometric Partitioning}
\label{app:geometric-partitioning}

\hyperref[sec:linked-3-4]{Section 3.4} of the main text briefly discusses why a complete object is not necessarily suitable as one computational unit and the corresponding solution strategy. A complete analysis and argument are provided here. For a curved faucet, for example, its front and rear ends are difficult to summarize accurately with the same center. When the directional responses of these surfaces are jointly mapped onto a reference sphere, the original relationship between position and normal changes. The choice of computational units must therefore consider geometric form rather than only object name or membership.

For a point set with reference center $\mathbf c$ and reference-sphere radius $R$, let $\mathbf d_i=\mathbf p_i-\mathbf c$. The positive intersection of a ray emitted from surface point $\mathbf p_i$ along the unit outward normal $\mathbf n_i$ with the reference sphere is

\begin{equation}
t_i=-\mathbf n_i^\top\mathbf d_i+
\sqrt{(\mathbf n_i^\top\mathbf d_i)^2+R^2-\|\mathbf d_i\|^2},
\qquad
\mathbf r_i=\frac{\mathbf p_i+t_i\mathbf n_i-\mathbf c}{R}.
\end{equation}

Let $\mathbf d_i^\perp=(\mathbf I-\mathbf n_i\mathbf n_i^\top)\mathbf d_i$. The expression above can also be written as

\begin{equation}
\mathbf r_i=\frac{\mathbf d_i^\perp}{R}
+\sqrt{1-\frac{\|\mathbf d_i^\perp\|^2}{R^2}}\,\mathbf n_i.
\end{equation}

When a sphere uses its center as the reference center and its normals are radial, $\mathbf d_i^\perp=0$, and the mapped direction is identical to the original normal. For extended, bent, or separated structures, a shared center may not align the normals everywhere simultaneously. This geometric mapping occurs before spherical-harmonic encoding. Increasing the spherical-harmonic order can only improve the encoding of the mapped function and cannot eliminate the change introduced by the mapping itself.

To examine these geometric differences, \hyperref[fig:app-shape-mapping-loss]{Figure \ref*{fig:app-shape-mapping-loss}} compares spherical-response mapping loss across geometries ranging from a standard spherical point cloud to shapes that progressively depart from a sphere.

Directions are sampled uniformly and densely on the sphere. For each direction, the mean angular response over all points is computed using the original normals and the mapped directions, respectively. Each of the two response sets is divided by its own mean over these directions. Using the normalized original response as the reference, the relative $L_2$ error is computed and reported as a percentage. The results are shown in \hyperref[fig:app-shape-mapping-loss]{Figure \ref*{fig:app-shape-mapping-loss}}.

As shown, spheres and near-spherical geometries have low loss, whereas elongated polyhedra and chairs have higher loss. Based on this observation, we conjecture that the applicability of a single-center representation is related to geometric shape and that using more rounded computational units may reduce the loss.

\begin{figure}[H]

\centering
\includegraphics[width=0.95\textwidth]{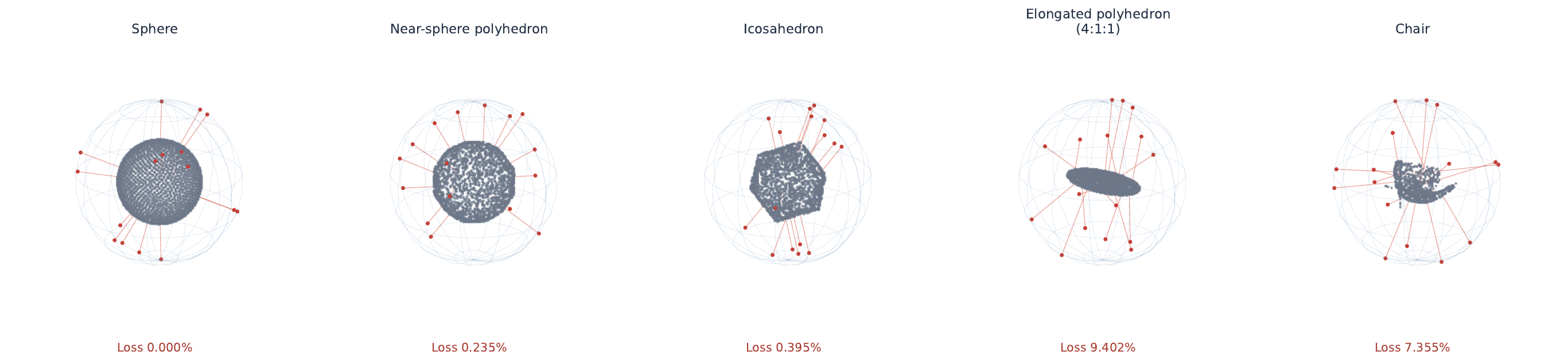}
\caption{Spherical-response mapping loss for point clouds of different geometric shapes.}\label{fig:app-shape-mapping-loss}
\end{figure}

We therefore attempt to geometrically partition the point cloud so that the geometric structures of the computational units become more rounded. This paper uses simple recursive principal-axis bisection: at each step, the unit with the largest spatial extent is selected and split along its principal axis into two balanced subsets until the desired unit budget is reached. For the chair at the end of \hyperref[fig:app-shape-mapping-loss]{Figure \ref*{fig:app-shape-mapping-loss}}, the effect of its geometric partitioning on representation loss is shown in Figures \hyperref[fig:app-uniform-partition-loss]{\ref*{fig:app-uniform-partition-loss}} and \hyperref[fig:app-query-partition-loss]{\ref*{fig:app-query-partition-loss}}. This experiment focuses on the difference in directional response before and after spherical mapping and does not account for occlusion between units.

\begin{figure}[H]

\centering
\includegraphics[width=0.95\textwidth]{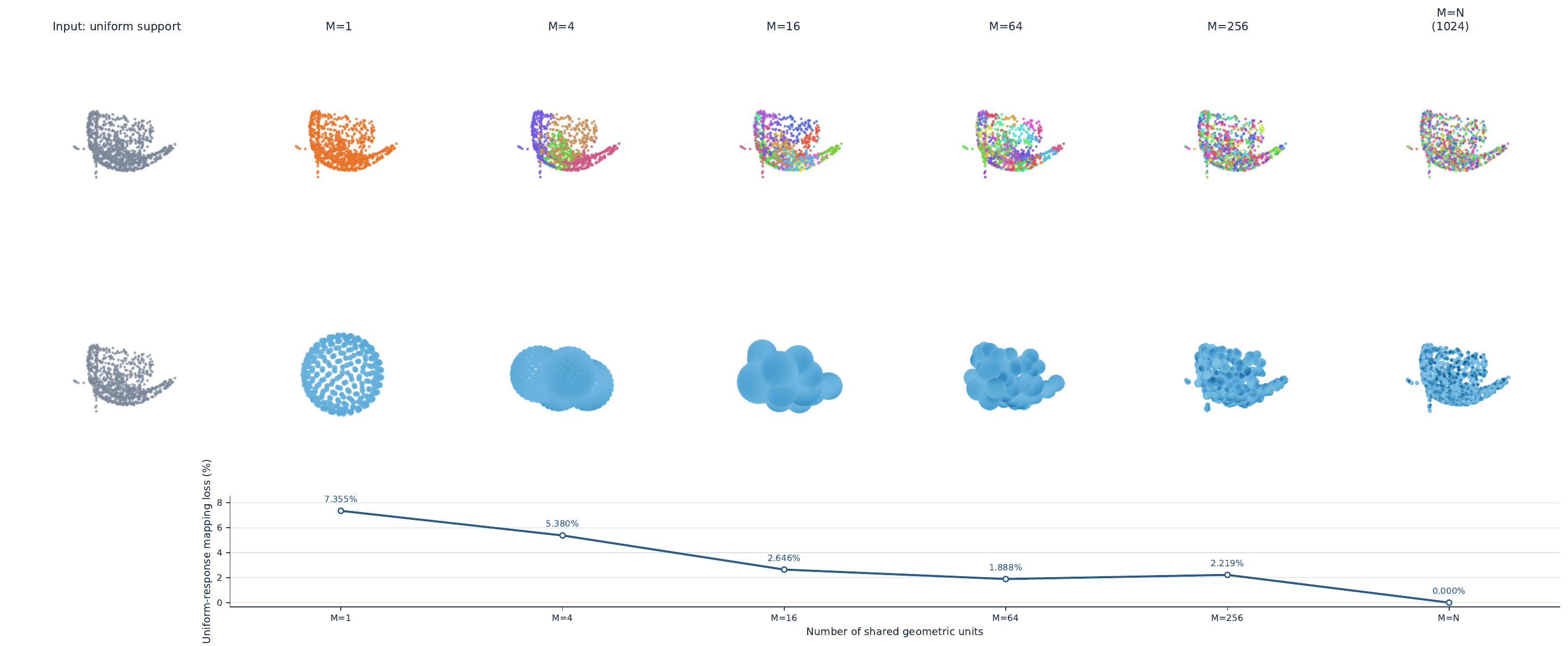}
\caption{Geometric-partitioning loss for a uniformly weighted point cloud.}\label{fig:app-uniform-partition-loss}
\end{figure}

\begin{figure}[H]

\centering
\includegraphics[width=0.95\textwidth]{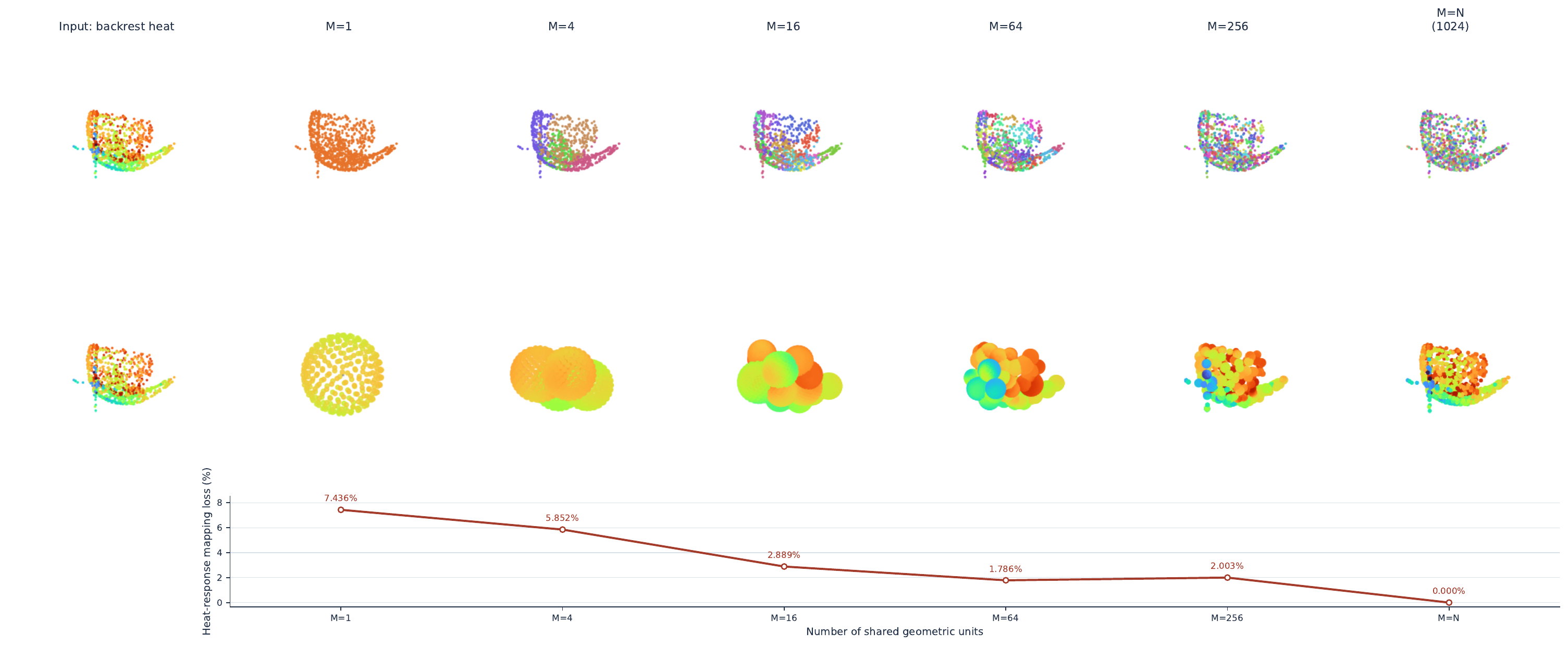}
\caption{Geometric-partitioning loss for a point cloud whose query value is the backrest.}\label{fig:app-query-partition-loss}
\end{figure}

\hyperref[fig:app-uniform-partition-loss]{Figure \ref*{fig:app-uniform-partition-loss}} places greater emphasis on observing the loss caused by point-cloud geometry, whereas \hyperref[fig:app-query-partition-loss]{Figure \ref*{fig:app-query-partition-loss}} examines whether the weight distribution of a query-weighted point cloud can be retained. As shown, the accuracy loss decreases overall as the number of partitions increases. When the number of partitions reaches the limit, namely one point per computational unit, the loss falls to 0.

We further scale up this experiment and conduct the same experiment on 100 randomly selected objects from the PartNet dataset \citep{mo2019partnet}. The results are as follows.

\begin{figure}[H]

\centering
\includegraphics[width=0.95\textwidth]{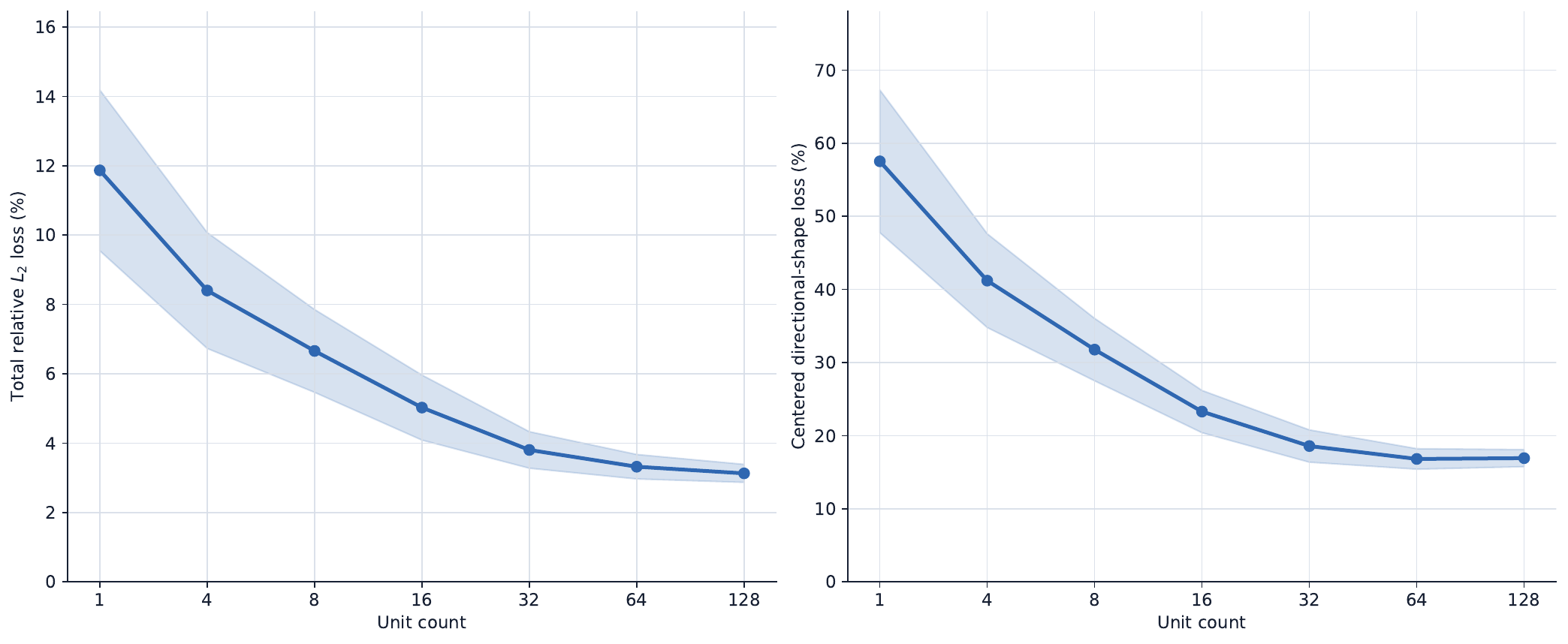}
\caption{Trend analysis of the number of geometric partitions and representation loss.}\label{fig:app-partition-budget}
\end{figure}

Overall, the representation loss continues to decrease as the number of partitions increases. At the partitioning limit, the computation can be equivalent to pointwise computation, apart from the truncation loss of the spherical harmonics themselves. In practical use, however, such extreme partitioning is unnecessary. As shown, a small number of early partitions can recover most of the representation accuracy, while the benefit of continued partitioning diminishes later.

\subsubsection{Gaussian Occupancy}
\label{app:gaussian-occupancy}
\begingroup
\setlength{\jot}{7pt}

The core of spatial occupancy modeling is to represent discrete geometry as a continuous, differentiable spatial occupancy function and to describe occlusion attenuation by integrating along observation paths, thereby making the occlusion computation differentiable as well. This construction is not restricted to any particular method. In principle, any functional form or construction method may be used, as long as the resulting spatial occupancy function remains differentiable and provides a reasonable description of occlusion. For example, more expressive continuous scene representations, such as 3D Gaussian Splatting \textcolor{blue}{\citep{kerbl2023gaussian}}, could in principle be used to construct a higher-fidelity occupancy model. 

In this work, we use Gaussian functions to represent local occupancy, and this section presents only one example construction used in our experimental implementation.

\paragraph{Occupancy construction.}

For a geometric unit $k$ with surface samples $\mathbf p_i$ and area weights
$a_i>0$, define its represented area and Gaussian mean as
\begin{equation}
\begin{aligned}
A_k
&= \sum_{i\in k} a_i, \\
\boldsymbol\mu_k
&= \frac{1}{A_k}\sum_{i\in k}a_i\mathbf p_i.
\end{aligned}
\end{equation}
We recommend an area-weighted covariance with a positive minimum width:
\begin{equation}
\mathbf\Sigma_k
=
\frac{1}{A_k}\sum_{i\in k}
a_i(\mathbf p_i-\boldsymbol\mu_k)
(\mathbf p_i-\boldsymbol\mu_k)^\top
+\sigma_0^2\mathbf I.
\end{equation}
Here $a_i$ is the geometric area weight defined in \hyperref[app:area-weighting]{Section \ref*{app:area-weighting}}, without the task-relevance factor $s_i$; $\sigma_0>0$ ensures positive definiteness. The Gaussian mean
$\boldsymbol\mu_k$ is distinguished from the directional-response reference
center $\mathbf c_k^{\mathrm{ref}}\equiv\mathbf c_k$ defined in
\hyperref[sec:linked-3-4]{Section~3.4}: the two may coincide, but the framework
does not require this.

The resulting occupancy density is
\begin{equation}
\rho_k(\mathbf z)
=
\alpha_k\exp\!\left[
-\frac12(\mathbf z-\boldsymbol\mu_k)^\top
\mathbf\Sigma_k^{-1}
(\mathbf z-\boldsymbol\mu_k)
\right],
\end{equation}
with amplitude
\begin{equation}
\alpha_k
=
\frac{\kappa_k A_k}
{(2\pi)^{3/2}|\mathbf\Sigma_k|^{1/2}},
\qquad \kappa_k\ge0.
\end{equation}
The covariance describes spatial shape, while $\kappa_k$ controls attenuation
strength. This normalization yields
\begin{equation}
\int_{\mathbb R^3}\rho_k(\mathbf z)\,\mathrm d\mathbf z
=\kappa_k A_k,
\end{equation}
so increasing sample density while preserving area weights does not itself
increase occupancy mass. Here $\rho_k$ is an attenuation density.

\paragraph{Finite-segment attenuation.}

For a source unit with reference center $\mathbf c_j^{\mathrm{ref}}$ and an
observation position $\mathbf x$, define
\begin{equation}
\begin{aligned}
\ell
&= \bigl\|\mathbf x-\mathbf c_j^{\mathrm{ref}}\bigr\|, \\
\mathbf u
&= \frac{\mathbf x-\mathbf c_j^{\mathrm{ref}}}{\ell}.
\end{aligned}
\end{equation}
The contribution of an occluding unit $k$ is integrated along this finite
segment. Accumulating the contributions gives the optical depth and
transmittance:
\begin{equation}
\begin{aligned}
I_{jk}(\mathbf x)
&= \int_0^\ell
\rho_k\!\left(\mathbf c_j^{\mathrm{ref}}+s\mathbf u\right)
\,\mathrm ds, \\
\tau_j(\mathbf x)
&= \sum_{k\ne j}I_{jk}(\mathbf x), \\
T_j(\mathbf x)
&= \exp\!\left[-\tau_j(\mathbf x)\right].
\end{aligned}
\end{equation}
This construction excludes the source unit itself; other units, including
those belonging to the same object, remain potential occluders.

For one source--occluder pair, introduce
\begin{equation}
\begin{aligned}
\mathbf Q
&= \mathbf\Sigma_k^{-1}, \\
\mathbf r
&= \mathbf c_j^{\mathrm{ref}}-\boldsymbol\mu_k,
\end{aligned}
\end{equation}
and the scalar coefficients
\begin{equation}
\begin{aligned}
a &= \mathbf u^\top\mathbf Q\mathbf u, \\
b &= \mathbf u^\top\mathbf Q\mathbf r, \\
h &= \mathbf r^\top\mathbf Q\mathbf r.
\end{aligned}
\end{equation}
Completing the square yields
\begin{equation}
\begin{aligned}
I_{jk}
&= \alpha_k
\exp\!\left[-\frac12\left(h-\frac{b^2}{a}\right)\right]
\sqrt{\frac{\pi}{2a}} \\
&\quad\times
\left[
\operatorname{erf}\!\left(\frac{a\ell+b}{\sqrt{2a}}\right)
-
\operatorname{erf}\!\left(\frac{b}{\sqrt{2a}}\right)
\right],
\end{aligned}
\end{equation}
where $\operatorname{erf}$ is the error function.

\paragraph{Spatial derivatives.}

With scene parameters fixed, moving the observation position changes both
path length and direction. Write the density along the segment as
\begin{equation}
g(s)
=
\alpha_k\exp\!\left[
-\frac12\left(as^2+2bs+h\right)
\right].
\end{equation}
Its first two weighted moments have the closed forms
\begin{equation}
m_1
= \int_0^\ell s\,g(s)\,\mathrm ds
= \frac{g(0)-g(\ell)-bI_{jk}}{a},
\end{equation}
\begin{equation}
m_2
= \int_0^\ell s^2g(s)\,\mathrm ds
= \frac{I_{jk}-\ell g(\ell)-bm_1}{a}.
\end{equation}
The spatial derivative of the Gaussian contribution is therefore
\begin{equation}
\begin{aligned}
\nabla_{\mathbf x}I_{jk}
&= g(\ell)\mathbf u \\
&\quad-
\frac{\mathbf I-\mathbf u\mathbf u^\top}{\ell}
\left(m_2\mathbf Q\mathbf u+m_1\mathbf Q\mathbf r\right).
\end{aligned}
\end{equation}
The first term accounts for path-length variation and the second for
path-direction variation. The transmittance gradient follows directly:
\begin{equation}
\nabla_{\mathbf x}T_j
=
-T_j\sum_{k\ne j}\nabla_{\mathbf x}I_{jk}.
\end{equation}
These expressions are valid for
$\mathbf x\ne\mathbf c_j^{\mathrm{ref}}$.
The integral and its derivative are exact for the chosen Gaussian density;
agreement with physical scene occlusion remains a property of the occupancy
approximation.

One possible direction for extension is that, when the information field is used solely to describe visual accessibility, the occupancy function could further account for the light-transmission properties of different materials. For example, transparent glass and an opaque brick with the same shape and size may occupy similar geometric extents but produce different occlusion effects. With the same Gaussian geometric parameters, this difference can be modeled by adjusting the occupancy strength, such that transparent units induce weaker attenuation along propagation paths.

\endgroup

\subsubsection{Illustrations of the Effect of Gaussian Occlusion in a Differentiable Information Field}
\label{app:gaussian-occlusion-illustrations}

These two figures visualize the effect of Gaussian occlusion in a differentiable information field and are provided only as illustrations to aid understanding. They respectively show the field-value and gradient distributions formed by a single information source under one occluder and under multiple occluders, and illustrate how a given starting point moves along the gradient direction from an occluded region to a region with a higher field value.

\begin{figure}[H]

\centering
\includegraphics[width=0.95\textwidth]{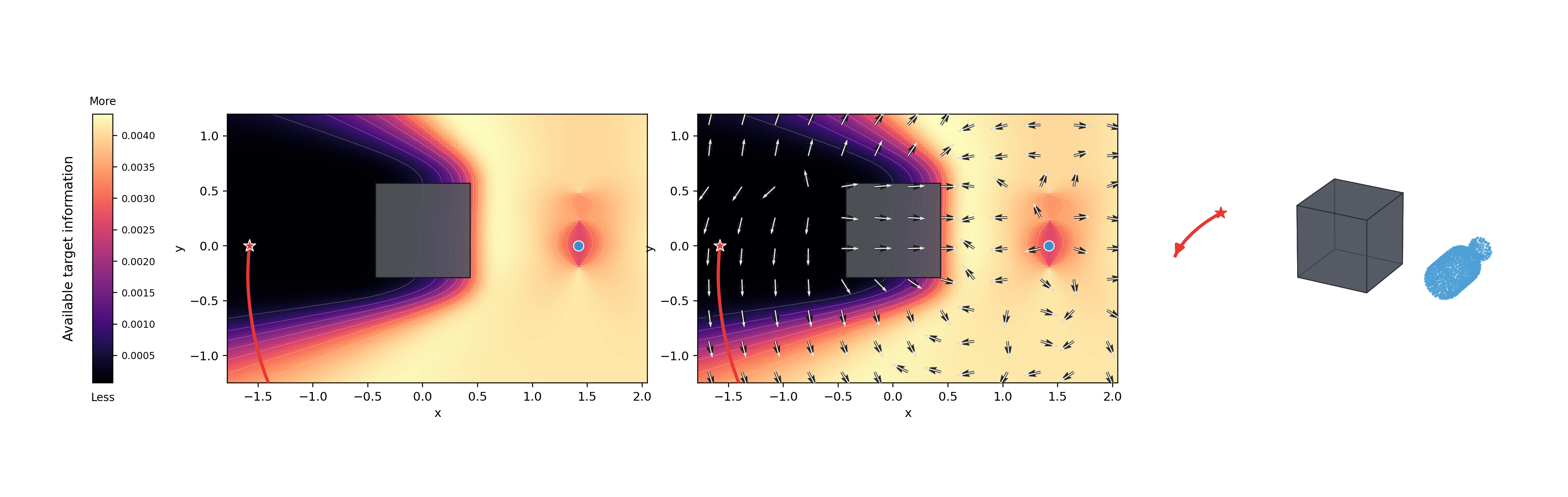}
\caption{Illustration of InfoField3D under a single occluder.}\label{fig:app-single-occluder}
\end{figure}

\begin{figure}[H]

\centering
\includegraphics[width=0.95\textwidth]{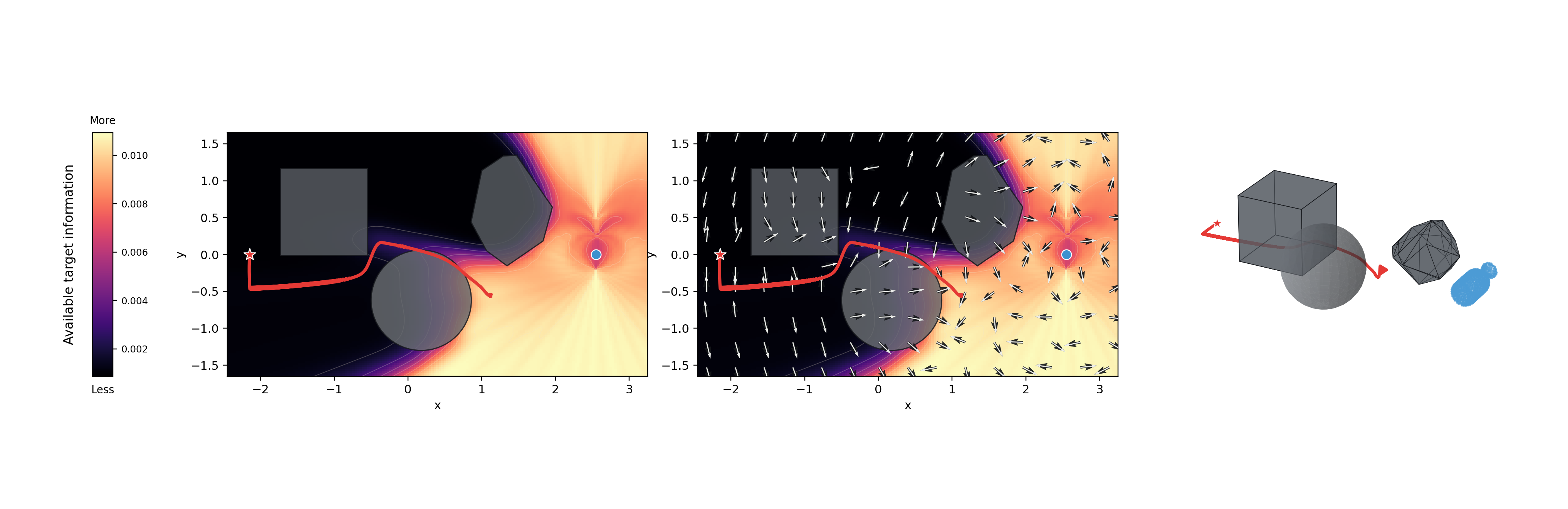}
\caption{Illustration of InfoField3D under multiple occluders.}\label{fig:app-multiple-occluders}
\end{figure}

\subsection{Vector-Valued Spherical Harmonics for Repeated Queries}
\label{app:vector-readout}

The main text discusses first querying the semantic point cloud according to the task condition and then using spherical harmonics to encode the corresponding task response. In this process, the semantic point cloud is already reusable, but each new query still requires pointwise computation of semantic relevance, so the computational scale remains tied to the point cloud. To address this, the earlier idea of using spherical harmonics to compress and encode point-cloud responses can be extended by using each point's semantic vector as the vector weight of its directional response. This first yields an uncompressed vector directional field

\begin{equation}
\mathbf F_{j,\mathrm{sph}}(\mathbf u)=\sum_{i\in j}a_i\mathbf f_iK(\mathbf r_i^\top\mathbf u),
\end{equation}

which is then encoded with spherical harmonics to obtain a finite-order vector-valued spherical-harmonic function

\begin{equation}
\mathbf F_{j,L}(\mathbf u)=\sum_{\ell=0}^{L}\sum_{m=-\ell}^{\ell}
\mathbf c_{j,\ell m}Y_{\ell m}(\mathbf u),
\qquad\mathbf c_{j,\ell m}\in\mathbb R^D.
\end{equation}

Here, $D$ is the dimension of the semantic features. The original spherical-harmonic function produces a scalar response for a particular query; here, the point-cloud semantic field, a vector map, is directly encoded to obtain a vector-valued spherical-harmonic function. Its function value in each direction represents the semantic features aggregated along that direction and is independent of a particular query. When a query arrives, given a query vector $\mathbf q$, the vector coefficients or directional function of the vector-valued spherical-harmonic function can be inner-producted with it at the readout stage, thereby reading out the entire vector-valued spherical-harmonic function as a scalar spherical-harmonic function for that query semantics. Only after an observation direction is further specified is the scalar response in that direction obtained. In this way, once the scene geometry and encoding settings are fixed, the same vector coefficients can serve multiple linear queries without traversing the original point cloud again for every query. When geometric partitioning, directional mapping, area weights, kernel functions, and fitting settings are fixed, and both encoding and querying are linear operations, querying before encoding and encoding before readout yield equivalent scalar spherical-harmonic functions. Formally,

\begin{equation}
F_{j,L}^{(\mathbf q)}(\mathbf u)
=\mathbf q^\top\mathbf F_{j,L}(\mathbf u)
=\sum_{\ell,m}(\mathbf q^\top\mathbf c_{j,\ell m})Y_{\ell m}(\mathbf u).
\end{equation}

\begin{figure}[H]

\centering
\includegraphics[width=0.95\textwidth]{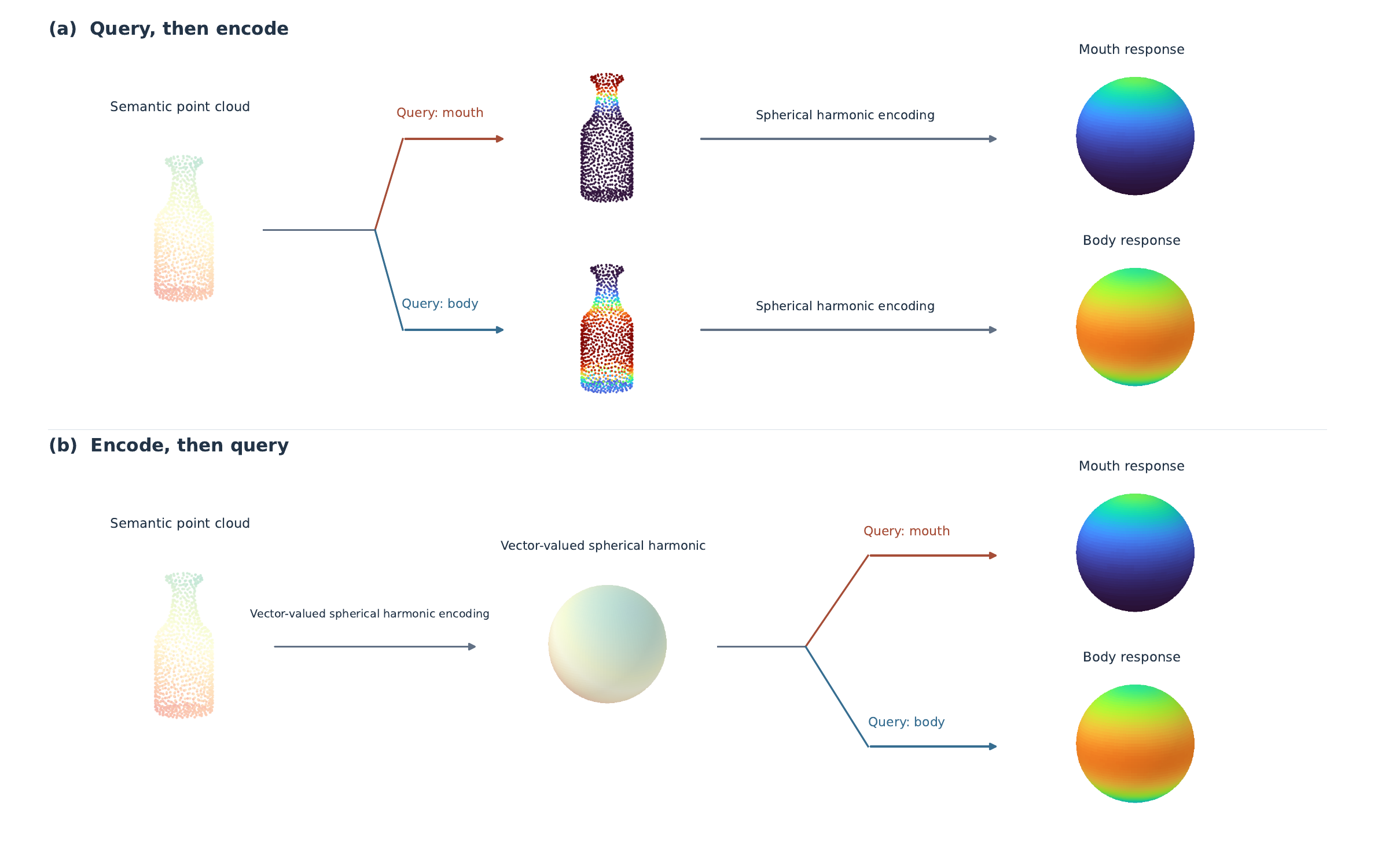}
\caption{Two construction forms: querying before encoding and encoding before readout.}\label{fig:app-vector-readout}
\end{figure}

After fixing geometric partitioning, mapping, fitting weights, and occupancy functions, spherical-harmonic encoding can be written as a linear operator $\mathcal E$, so that

\begin{equation}
\mathcal E(\{\mathbf q^\top\mathbf f_i\}_i)
=\mathbf q^\top\mathcal E(\{\mathbf f_i\}_i).
\end{equation}

Likewise, the vector response after occlusion attenuation can be aggregated before readout:

\begin{equation}
\mathbf G(\mathbf x)=\sum_jT_j(\mathbf x)\mathbf F_{j,L}(\mathbf u_j(\mathbf x)),
\qquad J_{\mathbf q}(\mathbf x)=\mathbf q^\top\mathbf G(\mathbf x).
\end{equation}

Therefore, under the linear conditions above, querying before encoding and encoding before readout produce the same response. With the latter, a new query no longer requires computation over the point cloud and only requires a readout over a small number of spherical-harmonic parameters, enabling faster task switching in scenarios that require frequent changes of task.

\subsection{Construction Details When the Task Condition Contains Multiple Targets}
\label{app:multiple-targets}

When textual task descriptions are used to obtain the task-relevance map, one point worth noting is that, for tasks involving multiple targets, directly constructing a single objective from the entire task text may not always be the best choice.

For a task condition containing multiple targets, one direct approach is to query the complete task description as a whole. For example, ``turn on the desk lamp and place the bottle in the drawer'' can be encoded as one query vector, producing a unified relevance map. This approach requires constructing and optimizing only one objective, and its computational form is no different from that of a single-target query. However, responses for different targets are merged into one function, and subsequent optimization considers only whether the total response increases, without explicitly controlling the contribution of each target. When the relevant region of a target is small or its response scale is relatively weak, it may be masked by the high response of other targets, so a position with a high overall score may still fail to observe that target sufficiently.

Another potentially better approach is to query multiple targets separately, for example, ``desk lamp,'' ``bottle,'' and ``drawer,'' define an independent objective function for each target, and then aggregate these objective functions into a joint objective by a weighted product. Let $\mathcal Q=\{\mathbf q_1,\ldots,\mathbf q_P\}$ be the set of queries corresponding to multiple targets, and let $\overline{J}_{\mathbf q_k}(\mathbf x)$ denote the nonnegative response of target $k$ at position $\mathbf x$ after accounting for occlusion and applying a common scale. Separate queries can be constructed through the pointwise scalar procedure described above or read directly from the vector-valued spherical-harmonic function in \hyperref[app:vector-readout]{Appendix \ref*{app:vector-readout}}. This requires multiple readouts but enables the joint objective to explicitly express the relative priority of each target and the requirement that they be jointly satisfied.

One feasible joint form is the weighted product

\begin{equation}
J_{\mathcal Q}^{\mathrm{prod}}(\mathbf x)
=\prod_{k=1}^{P}\left[\overline{J}_{\mathbf q_k}(\mathbf x)\right]^{\lambda_k},
\qquad\lambda_k>0,\quad\sum_{k=1}^{P}\lambda_k=1.
\end{equation}

Here, $\lambda_k$ denotes the relative priority of target $k$. In this form, a low response from any target directly affects the joint response: when the response of one target approaches zero, the product also approaches zero; when one target has a low response, the gain from further improving other targets while ignoring it is suppressed. The optimization therefore tends, within the feasible range, to find positions that satisfy all targets simultaneously without requiring an additional visibility threshold. Comparable channel scales aid interpretation of the joint score; when the responses are strictly positive, the product remains smooth. Depending on task preferences, a weighted sum, harmonic mean, or another aggregation form can also be used.

\clearpage
\begingroup
\setlength{\parskip}{0.2em}
\section{Representation Fidelity and Compression}
\label{app:representation-fidelity}

\subsection{Representation-Fidelity Experiment Details}
\label{app:fidelity-experiment}

The experiment uses 130 LIBERO and 1,525 LIBERO-Plus static scenes, for a total of 1,655 scenes. Each simulated scene is converted into a point cloud containing 47,500 points. Approximately 800 points are uniformly sampled in the feasible domain as the test set. At each test point, the information response is computed using the original surface point cloud, including point weights, and occlusion is determined using the simulation mesh; the analytic field is evaluated at the same position using the same weights and aggregation rules. Two settings are used for the weights on the scene point-cloud surface: uniform relevance, which places greater emphasis on geometry itself, and relevance weights obtained from the objects designated by the task in the scene, which introduce more variation. In the experiment, our method still obtains computational units through recursive principal-axis partitioning, using approximately 600 units per scene on average and a spherical-harmonic function with $L=4$ for each unit, requiring 25 coefficients.

Among the metrics, the optimal-value attainment rate measures whether the position at which InfoField3D takes its maximum in the test set remains a high-value position under computation with the original point cloud. It is measured by comparing the value of that position under the original point-cloud computation with the maximum selected by the original point-cloud computation itself. Shape consistency uses the Pearson correlation of response values; ranking consistency uses the correlation of position-ranking indices, with equal response values ordered according to the original candidate order.

The compression ratio described in the main text compares geometric and scalar-response representations. Each pointwise sample stores 7 scalars (position, normal, and area weight), whereas each InfoField3D unit stores 34 scalars (center, covariance, and 25 spherical-harmonic coefficients). Aggregated over the dataset, this corresponds to approximately a $15\times$ reduction in representation size.

\subsection{Stress Test under Different Point-Cloud Densities}
\label{app:density-stress-test}

With the number-of-units budget of InfoField3D held constant, only the sampling density of the scene point cloud is increased to stress-test the representation capacity of InfoField3D. The results are shown in \hyperref[tab:app-density-stress-test]{Table \ref*{tab:app-density-stress-test}}.

\begin{table}[H]

\centering
\caption{Representation fidelity under the stress test.}\label{tab:app-density-stress-test}
\resizebox{\textwidth}{!}{%
\begin{tabular}{lrrrr}
\toprule
Relevance & Reference density & Optimal-value attainment $\uparrow$ & Shape consistency $\uparrow$ & Ranking consistency $\uparrow$ \\
\midrule
Uniform & $1\times$ & 97.93\% & 92.28\% & 90.51\% \\
Uniform & $2\times$ & 97.90\% & 92.17\% & 90.37\% \\
Uniform & $4\times$ & 97.87\% & 92.16\% & 90.32\% \\
Uniform & $16\times$ & 97.86\% & 92.11\% & 90.25\% \\
Task-related & $1\times$ & 97.97\% & 93.86\% & 91.29\% \\
Task-related & $2\times$ & 97.89\% & 93.74\% & 91.08\% \\
Task-related & $4\times$ & 97.84\% & 93.73\% & 91.02\% \\
Task-related & $16\times$ & 97.82\% & 93.68\% & 90.93\% \\
\bottomrule
\end{tabular}}
\end{table}

Under both relevance settings, even when the reference density is increased by a factor of 16, the count ratio is approximately 239:1, and the change in every metric relative to its initial value remains below 0.4 percentage points.

\endgroup
\clearpage
\section{Viewpoint Search}
\label{app:viewpoint-search}

\subsection{Experimental Scene Selection and Feasible-Domain Setting}
\label{app:search-scenes-domain}

The experiment uses 20 scenes randomly selected from each of the four task suites of the LIBERO-Plus object-layout-variation scenes, for a total of 80 scenes. The scene list is as follows.

\begin{table}[H]

\centering
\caption{Experimental scene list.}\label{tab:app-scene-list}
\scriptsize
\begin{tabularx}{\textwidth}{@{}lX@{}}
\toprule
Task suite & Scene IDs \\
\midrule
LIBERO-Spatial & 1904, 2062, 1991, 1791, 2098, 1733, 1772, 2018, 1976, 1830, 2033, 1989, 1732, 1859, 1901, 2022, 1799, 1889, 2051, 2010 \\
LIBERO-Object & 1888, 2071, 1962, 1878, 2013, 1858, 1940, 2129, 2169, 2182, 1958, 1989, 1864, 2015, 2051, 1924, 2087, 2156, 1982, 2009 \\
LIBERO-Goal & 2258, 2305, 2055, 2290, 2298, 1930, 2234, 2306, 2119, 1925, 2227, 2269, 2144, 2100, 2033, 2074, 2009, 2252, 2312, 2160 \\
LIBERO-10 & 2157, 1981, 1972, 2162, 1982, 2141, 1970, 2160, 1980, 2130, 1971, 2156, 1976, 2080, 2051, 2112, 1946, 1967, 2190, 2115 \\
\bottomrule
\end{tabularx}
\end{table}

For each scene, a cube above the tabletop/floor where the task is located is defined as the feasible domain. At the same time, to ensure that the camera frustum can at least cover the task region, a minimum bounding sphere is constructed for the set of task-related objects. A sphere is then formed using the radius at which the camera frustum, here the default LIBERO camera frustum, can cover the minimum bounding sphere and using the center of the minimum bounding sphere as its center. The region inside this sphere is defined as infeasible. The final feasible domain is the space inside the cube with the inner-sphere region removed; points on the boundaries of the cube and inner sphere are allowed. The side length of the cube is set to 2.2 times the radius of the inner sphere.

\subsection{Baselines and Task Adaptation}
\label{app:search-baselines}

Details of the compared methods are provided in Table \hyperref[tab:app-search-baselines]{\ref*{tab:app-search-baselines}}.

\begingroup
\small
\setlength{\tabcolsep}{4pt}
\renewcommand{\arraystretch}{1.08}
\setlength{\LTpre}{0.5em}
\setlength{\LTpost}{0.5em}
\setlength{\LTcapwidth}{\textwidth}
\captionsetup[longtable]{
  font=small,
  labelfont=bf,
  labelsep=period,
  justification=raggedright,
  singlelinecheck=false
}

\noindent
\textit{Adaptation note.}
For baselines requiring adaptation to the shared task and camera-position setting, we retain the core algorithmic mechanisms of the original methods and make only the minimal changes needed to provide the common inputs, objective interface, and feasible-domain constraints. \textcolor{blue}{The task pixel score in this table, also written as the task-weighted hard-ray pixel score, is the Task-region visibility defined in Table C3.}

\begin{longtable}{@{}>{\raggedright\arraybackslash}p{0.18\textwidth}>{\raggedright\arraybackslash}p{\dimexpr0.82\textwidth-2\tabcolsep\relax}@{}}
\caption{Compared methods, objectives, implementation sources, and experimental adaptations.}\label{tab:app-search-baselines}\\
\toprule
Method & Objective and implementation \\
\midrule
\endfirsthead

\caption*{\textbf{Table \thetable\ (continued).} Compared methods, objectives, implementation sources, and experimental adaptations.}\\
\toprule
Method & Objective and implementation \\
\midrule
\endhead

Gaussian-process Bayesian optimization (GP-BO) &
Task pixel score and expected improvement. Uses an in-house implementation of Gaussian-process regression and expected improvement. Camera positions are normalized to the feasible domain, evaluated using the shared task pixel scorer, and restricted to feasible candidates. \\
\midrule

CIM surrogate gradient &
Surrogate gradients fitted to sampled task pixel scores. Builds on the per-line sampling, polynomial fitting, and gradient-update mechanisms of the official IGLOV code. The task pixel score is used as the task-specific information measure, with the corresponding threshold adjustment; sampling and movement are adapted to three-dimensional camera positions. \\
\midrule

Covariance matrix adaptation evolution strategy (CMA-ES) &
Task-weighted hard-ray pixel score. Uses the public pycma library while retaining its optimization core. The optimization variables are three-dimensional camera positions, the initial search scale is adjusted to scene size, and infeasible candidates are resampled. \\
\midrule

GradientNBV &
Task-adapted differentiable ray gain. Builds on the official voxel-grid and ray-sampling components while retaining their differentiable sampling and occlusion mechanisms. Frozen geometry provides the input, task relevance is used in place of semantic uncertainty, and multiple task roles are aggregated. The method is interfaced with the shared camera model and three-dimensional position-update formulation. \\
\midrule

VISO-grasp &
Analytic geometric velocity field defined by targets and occluders. Builds on the geometric velocity-field mechanism from the official vMF-Contact repository, retaining its anchor construction, occluder filtering, and velocity aggregation. Frozen geometry is used in place of online perception, task objects jointly define the target, and position updates are expressed within the shared feasible-domain constraints. \\
\midrule

Sobol &
Task-weighted hard-ray pixel score. Uses SciPy's Sobol sampler, maps three-dimensional samples to the feasible domain, and evaluates them with the shared task pixel scorer. \\
\midrule

Random &
Task-weighted hard-ray pixel score. Uses NumPy to sample camera positions uniformly within the feasible domain and evaluates them with the shared task pixel scorer. \\
\bottomrule
\end{longtable}
\endgroup

For InfoField3D, we use SciPy’s SHGO, with InfoField3D’s analytic gradient supplied to its local optimizer, and SLSQP implementations with default solver options, keeping all SLSQP settings fixed across the two gradient variants.

Specifically, in this experiment, we sum the occlusion-attenuated responses of the local units separately for each task target and normalize each target’s response by its total area-weighted relevance, which is independent of viewpoint. This preserves the objective’s emphasis on the total amount of available task-relevant information. We do not apply viewpoint-dependent coverage normalization, so the score cannot increase merely because less irrelevant geometry is observed. For multi-role tasks, we adopt the product aggregation described in Appendix A.3, specifically the equally weighted geometric mean of the normalized target scores, to balance the observation requirements of all targets. Low-scoring targets thus constrain the overall score, reducing the risk of a well-observed target masking another that is barely visible. SHGO, SLSQP with analytic gradients, and SLSQP with sampling-based gradient estimates all optimize this same objective. The comparison metrics are evaluated consistently across methods under this experimental setting.

Several learning-based viewpoint-prediction methods were also considered, including TVVE (repository name: TAVP), MVPNet, and VG-AVS. However, their currently available implementations either do not provide all components required for direct reproduction under our setting or adopt viewpoint-prediction formulations that differ from the free-space observation-position search considered in this paper. We therefore do not include them in the experimental comparison.

\subsection{Evaluation Metrics and Numerical Results}
\label{app:search-metrics-results}

The evaluation metrics are calculated as shown in \hyperref[tab:app-evaluation-metrics]{Table \ref*{tab:app-evaluation-metrics}}, and the numerical results in the figures of the main text are shown in Tables \hyperref[tab:app-infofield-scores]{\ref*{tab:app-infofield-scores}} through \hyperref[tab:app-gpt4o-scores]{\ref*{tab:app-gpt4o-scores}}.

Additional details on the figures in the main text: In Figure~\ref{fig:quality-time-curves}, curves show mean performance across scenes, and shaded bands indicate 95\% confidence intervals for the mean, obtained by bootstrap resampling of scenes with replacement. Error bars in Figure~\ref{fig:fragility-analysis}(a) indicate 95\% confidence intervals for the mean. In Figure~\ref{fig:fragility-analysis}(b), the solid curve shows the cross-scene mean, while the dark and light bands indicate the 25th--75th and 5th--95th percentile ranges across scenes, respectively.

\begin{table}[H]

\centering
\caption{Evaluation metrics and scoring mechanisms.}\label{tab:app-evaluation-metrics}

\small
\renewcommand{\arraystretch}{1.08}

\begin{tabularx}{\textwidth}{|p{4.3cm}|X|}
\hline
Metric & Definition or configuration \\
\hline

Task-region visibility (geometric mean across task roles)
& Measures the visible fraction of task-relevant surfaces, weighted by surface area and task relevance. A surface sample counts as visible only when it lies within the camera's field of view, \textcolor{blue}{does not face away from the camera},  and is not occluded\textcolor{blue}{; frontal and oblique views count equally}. Visibility is computed separately for each task role and combined using the geometric mean, so severe occlusion of one role lowers the overall score. \\
\hline

Directional visibility response
& Extends task-region visibility by weighting visible surfaces according to their viewing angle. The weight is highest for a frontal view and decreases exponentially as the view becomes more oblique. Occluded samples contribute nothing. Scores are combined using the geometric mean across task roles, reflecting both visibility and viewing direction. \\
\hline

InfoField3D score
& Although this is the objective value of our own method, we believe it is still informative for viewpoint evaluation because it jointly considers favorable viewing orientation, maximal exposure of the observation region, and reduced occlusion risk. We therefore include it as an additional reference metric.\\
\hline

GPT-4o semantic score
& Presents a high-quality reference image and the evaluated image from the same scene to GPT-4o. The reference calibrates the upper end of the scoring scale. Based on visible evidence, the model uses two items for single-role tasks and three for multi-role tasks, as defined in \hyperref[app:gpt4o-evaluation]{Section \ref*{app:gpt4o-evaluation}}. Each item receives an integer score from 0 to 10 and a brief justification. Item scores are averaged equally and converted to a 0--100 scale. \\
\hline

\end{tabularx}
\end{table}

\textbf{Note.} Higher scores indicate better performance for all metrics. 

``---'' indicates that the method had not produced an output within the corresponding time budget.

\begin{table}[H]

\centering
\caption{InfoField3D Score.}\label{tab:app-infofield-scores}
\begin{tabular}{lrrrrr}
\toprule
Method / search time (s) & 0.03 & 0.1 & 0.3 & 1 & 20 \\
\midrule
InfoField3D (SHGO) & \textbf{\textit{0.1995}} & \textbf{\textit{0.2091}} & \textbf{\textit{0.2113}} & \textbf{\textit{0.2113}} & \textbf{\textit{0.2113}} \\
InfoField3D (Analytic Gradient) & 0.1781 & 0.1786 & 0.1786 & 0.1786 & 0.1786 \\
InfoField3D (Sampling-Based Gradient) & 0.1282 & 0.1691 & 0.1710 & 0.1710 & 0.1710 \\
GP-BO & --- & 0.0299 & 0.0993 & 0.1243 & 0.1497 \\
CIM surrogate gradient & --- & 0.0360 & 0.0363 & 0.0482 & 0.1177 \\
CMA-ES & --- & 0.0333 & 0.0405 & 0.0676 & 0.1419 \\
GradientNBV & 0.0321 & 0.0526 & 0.0760 & 0.0810 & 0.0809 \\
VISO-grasp & 0.0710 & 0.0708 & 0.0707 & 0.0707 & 0.0707 \\
Sobol & --- & 0.0338 & 0.1020 & 0.1308 & 0.1470 \\
Random & --- & 0.0332 & 0.0814 & 0.1244 & 0.1497 \\
\bottomrule
\end{tabular}
\end{table}

\begin{table}[H]

\centering
\caption{Task-Region Visibility (\%).}\label{tab:app-task-visibility-scores}
\begin{tabular}{lrrrrr}
\toprule
Method / search time (s) & 0.03 & 0.1 & 0.3 & 1 & 20 \\
\midrule
InfoField3D (SHGO) & \textbf{\textit{46.96}} & \textbf{\textit{46.51}} & \textbf{\textit{46.32}} & \textbf{\textit{46.32}} & \textbf{\textit{46.32}} \\
InfoField3D (Analytic Gradient) & 40.27 & 40.30 & 40.30 & 40.30 & 40.30 \\
InfoField3D (Sampling-Based Gradient) & 31.92 & 39.29 & 39.37 & 39.37 & 39.37 \\
GP-BO & --- & 0.71 & 16.45 & 33.67 & 44.12 \\
CIM surrogate gradient & --- & 0.93 & 1.82 & 3.73 & 30.77 \\
CMA-ES & --- & 0.87 & 2.39 & 9.53 & 40.01 \\
GradientNBV & 0.72 & 5.22 & 15.80 & 19.28 & 19.11 \\
VISO-grasp & 20.37 & 20.53 & 20.54 & 20.54 & 20.54 \\
Sobol & --- & 0.78 & 17.66 & 32.50 & 43.37 \\
Random & --- & 0.73 & 13.97 & 31.25 & 43.20 \\
\bottomrule
\end{tabular}
\end{table}

\begin{table}[H]

\centering
\caption{Directional visibility response (\%).}\label{tab:app-directional-visibility-scores}
\begin{tabular}{lrrrrr}
\toprule
Method / search time (s) & 0.03 & 0.1 & 0.3 & 1 & 20 \\
\midrule
InfoField3D (SHGO) & \textbf{\textit{26.70}} & \textbf{\textit{27.59}} & \textbf{\textit{27.58}} & \textbf{\textit{27.58}} & \textbf{\textit{27.58}} \\
InfoField3D (Analytic Gradient) & 23.03 & 23.06 & 23.06 & 23.06 & 23.06 \\
InfoField3D (Sampling-Based Gradient) & 16.71 & 22.02 & 22.20 & 22.20 & 22.20 \\
GP-BO & --- & 0.34 & 8.80 & 17.02 & 22.46 \\
CIM surrogate gradient & --- & 0.45 & 0.87 & 1.82 & 15.98 \\
CMA-ES & --- & 0.42 & 1.15 & 4.61 & 20.24 \\
GradientNBV & 0.36 & 2.46 & 7.38 & 8.96 & 8.89 \\
VISO-grasp & 8.88 & 8.87 & 8.87 & 8.87 & 8.87 \\
Sobol & --- & 0.38 & 9.30 & 16.66 & 22.21 \\
Random & --- & 0.35 & 6.76 & 15.47 & 22.17 \\
\bottomrule
\end{tabular}
\end{table}

\begin{table}[H]

\centering
\caption{GPT-4o Score.}\label{tab:app-gpt4o-scores}
\begin{tabular}{lrrrrr}
\toprule
Method / search time (s) & 0.03 & 0.1 & 0.3 & 1 & 20 \\
\midrule
InfoField3D (SHGO) & \textbf{\textit{83.98}} & \textbf{\textit{83.42}} & \textbf{\textit{84.33}} & \textbf{\textit{84.33}} & 84.33 \\
InfoField3D (Analytic Gradient) & 75.90 & 76.50 & 76.50 & 76.50 & 76.50 \\
InfoField3D (Sampling-Based Gradient) & 53.77 & 66.77 & 66.60 & 66.60 & 66.60 \\
GP-BO & --- & 32.89 & 64.67 & 82.35 & 84.10 \\
CIM surrogate gradient & --- & 32.61 & 24.27 & 37.77 & 76.19 \\
CMA-ES & --- & 32.36 & 28.69 & 55.13 & 82.73 \\
GradientNBV & 22.69 & 39.13 & 49.63 & 52.38 & 53.50 \\
VISO-grasp & 49.65 & 50.98 & 50.52 & 50.58 & 50.58 \\
Sobol & --- & 27.74 & 62.54 & 81.00 & \textbf{\textit{84.40}} \\
Random & --- & 25.22 & 74.00 & 81.17 & 84.04 \\
\bottomrule
\end{tabular}
\end{table}

\clearpage
\begingroup
\setlength{\parskip}{0.2em}
\subsection{GPT-4o Viewpoint Evaluation}
\label{app:gpt4o-evaluation}

In our viewpoint-quality evaluation, to provide a more diverse set of evaluation metrics, we use a multimodal model in addition to scores based purely on geometric rules. This model-based score serves mainly as a relatively subjective external reference. We make every effort to ensure fairness in the scoring procedure and conduct additional tests to verify that the scores are largely stable for the same viewpoint. The detailed evaluation procedure is described below.

GPT-4o receives the task description, a reference image, and the image to be scored. The reference is selected by the highest task-region visibility among candidate viewpoints in the same scene and calibrates the upper end of the scale. All methods share this reference. Images are rendered at $224\times224$ pixels with consistent identity annotations to reduce object ambiguity; method names and search scores are withheld. The model evaluates task-related visibility from visible evidence, with temperature set to 0.

Scoring items follow the scene's fixed task roles and relevant-part lists (Table \hyperref[tab:app-gpt4o-scoring-items]{\ref*{tab:app-gpt4o-scoring-items}}). These are shared across methods. Source and placement roles are predefined, except for three-role LIBERO-Spatial tasks, where the placement target is resolved from the supplied task statement.

\begin{table}[H]

\centering
\caption{Task-specific GPT-4o scoring items.}\label{tab:app-gpt4o-scoring-items}
\small
\begin{tabularx}{\textwidth}{@{}p{0.22\textwidth}X@{}}
\toprule
Task roles & Scoring items \\
\midrule
Single role & Target identification; visibility of task-relevant parts and their spatial relationships. \\
Multiple roles & Source-object identification; grasp-region clarity; visibility of the placement target's receiving region. \\
\bottomrule
\end{tabularx}
\end{table}

For each item, GPT-4o gives a brief observation and an integer score from 0 to 10. The fixed rubric anchors 0 at an invisible or unrecognizable target or region, 5 at approximately half of the key region being visible with ambiguity or occlusion, and 10 at fully clear and complete task-relevant content. For placement, the criterion concerns the receiving region itself. The two scores for single-role tasks or three for multi-role tasks are averaged equally and multiplied by 10, giving a 0--100 score.

Fixed-seed subsets comprising approximately 20\% of unique evaluated views are scored again under the same task, reference, and model settings. The mean absolute difference between the original and repeated total scores ranges from 3.22\% to 6.42\% of the full scoring scale across methods. 

\endgroup
\clearpage
\subsection{Automated Multimodal Annotation Pipeline}
\label{app:semantic-annotation}

The task-relevance map indicates which regions of the 3D scene are important to observe for task execution. Its construction is not restricted to any particular method: it may be provided by multimodal models based on task understanding, derived from predefined rules, or even specified manually by humans. In our experiments, we use GPT-4o for task understanding and open-source models to generate the corresponding annotations.

The figure below shows an example from the LIBERO task ``Pick the ketchup and place it in the basket.'' GPT-4o identifies task-relevant basket regions and produces weighted positive and negative text queries from the task and nine initial RGB views (one shown). Modified-SAM \citep{kirillov2023sam} proposes image regions, and OpenCLIP \citep{radford2021clip,cherti2023openclip} scores them against the queries to produce image-space relevance. Depth and simulator instance correspondence bind these scores to the basket's 3D surface, where the rim and interior contributions are fused into a task-relevance map.

\begin{figure}[H]
\centering
\includegraphics[width=0.9\textwidth]{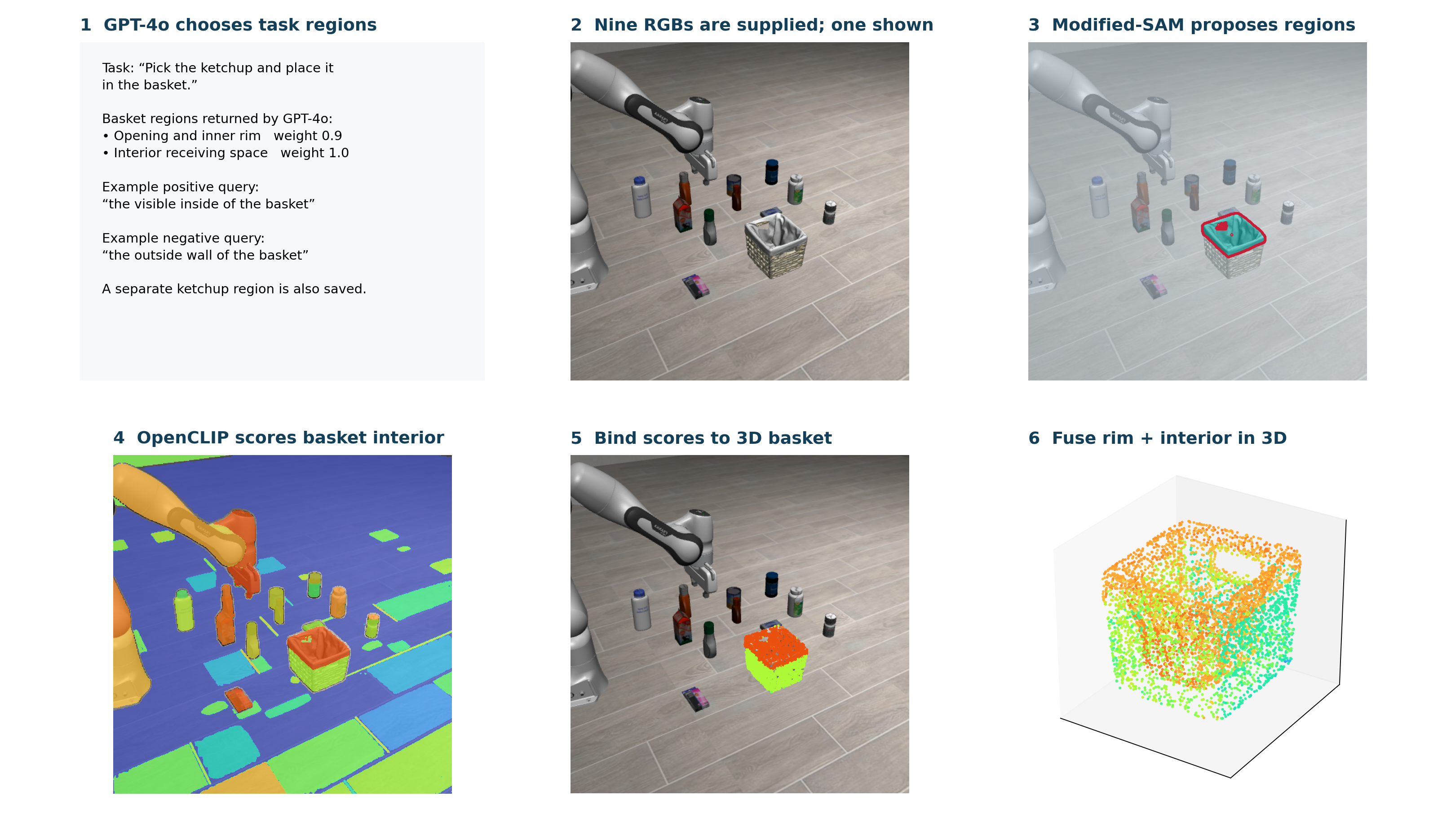}
\caption{\textbf{From task instructions to a task-conditioned 3D surface relevance map.}
}\label{fig:app-semantic-pipeline}

\end{figure}

\subsection{Preprocessing Cost}
\label{app:preprocessing-cost}

We measure construction from preloaded geometry and task-relevance
weights, excluding input generation, file I/O, query-engine
compilation and warm-up, and search.
All point-based methods use the same 43,955-point inputs;
our implementation uses 600 units.
VISO retains its native collision-mesh input.
Measurements use the same RTX 4090-equipped server, with one
PyTorch CPU thread and GPU synchronization at completion.
The five discrete baselines share the same semantic spatial-index
builder and therefore receive the same measured cost.

We additionally evaluate the vector-valued spherical harmonics described in \hyperref[app:vector-readout]{Appendix~\ref*{app:vector-readout}} as a comparison variant. It compresses both the original point cloud and the semantic features attached to the points into the representation, enabling fast responses to subsequent queries. Here, we evaluate 128- and 512-dimensional semantic features.

The results are shown in Table~\ref{tab:app-construction-cost}. When the initial construction cost is included, the search efficiency of the compared methods is shown in Figure~\ref{fig:app-preprocessing-efficiency}.

\begin{table}[H]
\centering
\caption{Representation construction and task-update costs with preloaded
geometry and task relevance.}
\label{tab:app-construction-cost}

\footnotesize
\setlength{\tabcolsep}{4pt}
\begin{tabular}{llrrr}
\toprule
Method & Representation & Construction (ms) & Update (ms) & Time saved (\%) \\
\midrule
Ours        & Continuous field & 66.81 & 2.35  & 96.48 \\
Ours        & Continuous field (128D) & 117.38 & 0.222 & 99.81 \\
Ours        & Continuous field (512D) & 153.56 & 0.278 & 99.82 \\
GradientNBV & Task-weighted voxel grid & 23.00 & 11.78 & 48.78 \\
VISO        & Bounding-box-based scoring function & 23.39 & 0.095 & 99.59 \\
GP-BO       & Semantic spatial index & 3.59 & 0.198 & 94.48 \\
CIM         & Semantic spatial index & 3.59 & 0.198 & 94.48 \\
CMA-ES      & Semantic spatial index & 3.59 & 0.198 & 94.48 \\
Sobol       & Semantic spatial index & 3.59 & 0.198 & 94.48 \\
Random      & Semantic spatial index & 3.59 & 0.198 & 94.48 \\
\bottomrule
\end{tabular}
\end{table}

\noindent
Construction times are averaged over 80 scenes
with five runs per scene in a warmed runtime. Updates use three task changes
per scene with five runs per change. Furthermore, we fix the scene and vary only the task-relevance weights by randomly selecting a new focus center and generating a new Gaussian relevance map, in order to measure the cost of task switching within the same scene. For each scene, we test three such task changes, each repeated five times, and measure the update time when reusing the cached geometry information. For the vector-valued SH variants, scene geometry and point features remain fixed; an update changes the query vector
and reads out the corresponding field. 

It can be seen that, although our method incurs a relatively higher upfront construction cost, it achieves a high degree of reuse across different tasks. With a larger upfront investment, the representation can be further extended to vector-valued spherical harmonics, enabling even faster switching between task queries. Unlike conventional discrete precomputation, however, the resulting representation is a general semantic 3D vector field that can directly accommodate different task queries.

\begin{figure}[H]
\centering
\includegraphics[width=0.9\textwidth]{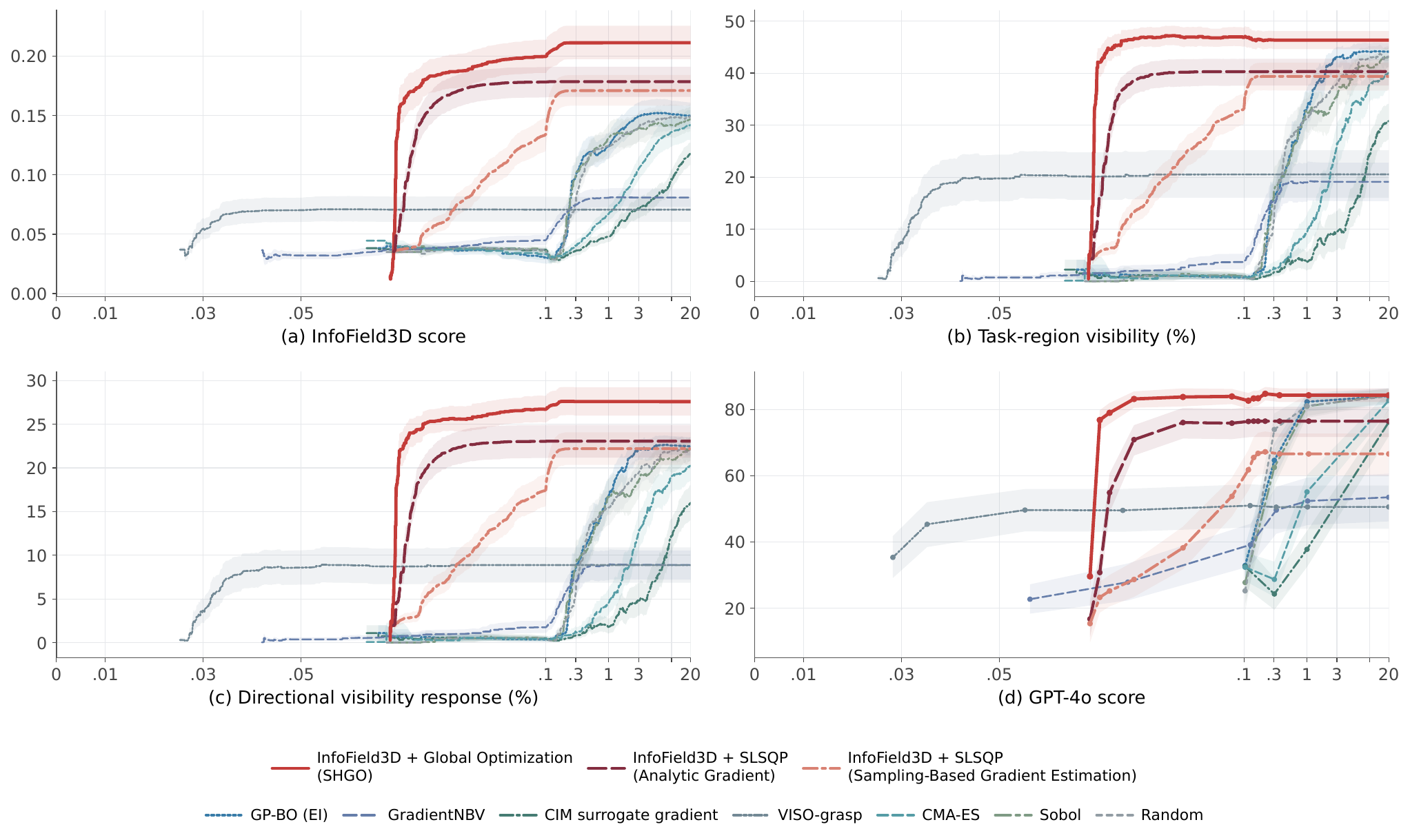}
\caption{\textbf{Efficiency curve accounting for preprocessing cost.}
}\label{fig:app-preprocessing-efficiency}

\end{figure}

\clearpage
\section{Viewpoint Robustness}
\label{app:viewpoint-robustness}

We evaluate the robustness of the viewpoints recommended by eight methods to positional perturbations in the 80 scenes of \hyperref[tab:app-scene-list]{Table \ref*{tab:app-scene-list}}. For each recommended viewpoint, six positional perturbations are applied along the positive and negative directions of a random orthonormal basis shared within the scene. The perturbation magnitudes are $\varepsilon\in\{0.005,0.01,0.02,0.04,0.08\}l$, where $l$ is the side length of the feasible-domain cube. After perturbation, the camera continues to face the same scene center. Evaluation uses the unified external visibility metric $V$. Perturbation loss is defined as the maximum nonnegative visibility decrease over valid test directions. The losses of the points recommended by each method under different perturbation strengths are shown in \hyperref[tab:app-perturbation-loss]{Table \ref*{tab:app-perturbation-loss}}.

\begin{table}[H]

\centering
\caption{Mean sampled perturbation loss at five displacement scales (percentage points; lower is better).}\label{tab:app-perturbation-loss}
\small
\resizebox{\textwidth}{!}{%
\begin{tabular}{lrrrrr}
\toprule
Method & Loss ($0.005l$) $\downarrow$ & Loss ($0.01l$) $\downarrow$ & Loss ($0.02l$) $\downarrow$ & Loss ($0.04l$) $\downarrow$ & Loss ($0.08l$) $\downarrow$ \\
\midrule
\textbf{InfoField3D (Ours)} & \textbf{0.122} & \textbf{0.224} & \textbf{0.513} & \textbf{0.998} & \textbf{2.522} \\
GP-BO & 0.174 & 0.308 & 0.595 & 1.301 & 2.841 \\
Sobol & 0.266 & 0.509 & 0.939 & 1.870 & 3.689 \\
Random & 0.281 & 0.507 & 0.935 & 1.920 & 3.969 \\
CIM & 0.450 & 0.796 & 1.627 & 3.039 & 5.931 \\
CMA-ES & 0.308 & 0.629 & 1.534 & 3.324 & 7.079 \\
GradientNBV & 0.501 & 1.001 & 2.537 & 5.556 & 9.838 \\
VISO-grasp & 0.464 & 0.825 & 1.431 & 2.316 & 2.833 \\
\bottomrule
\end{tabular}}
\end{table}

\clearpage
\section{Controlled VLA Evaluation}
\label{app:controlled-vla}

\subsection{Controlled VLA Experimental-Design Details}
\label{app:vla-experiment-design}

This experiment uses the same scene set as \hyperref[tab:app-scene-list]{Table \ref*{tab:app-scene-list}}.

In this experiment, we seek to minimize interference from the VLA model's own training distribution in the result analysis, thereby focusing more directly on the relationship between the information sufficiency of an observation viewpoint and downstream VLA performance, and on whether the movement direction recommended by InfoField3D is better than other equidistant directions. To this end, we use a controlled-variable experimental design and constrain additional changes beyond viewpoint as much as possible.

First, we constrain the feasible viewpoint region. A camera that is too far from or too close to the task region, or that departs substantially from the third-person observation distribution commonly seen by a VLA, may cause large changes in policy performance, and these changes may not arise from whether task information is sufficient. To reduce this factor, the experiment fixes the camera-to-task-center distance. The task center is defined as the center of the minimum bounding sphere of the task-related object set, and the camera radius is the distance from the default third-person camera to this center. Candidate positions are therefore first constrained to a sphere of fixed radius. Further considering that the default third-person camera is usually on the side opposite the robot arm, we exclude a spherical region of approximately $50^\circ$ on the robot-arm side and also remove physically invalid positions below the floor or tabletop. The result is an upper-hemisphere candidate region on the side opposite the robot arm with an opening angle of approximately $130^\circ$.

On this basis, we further construct a local viewpoint-rescue experiment. Initial viewpoints with a clear information bottleneck are first selected according to their hard-visibility scores. This is because, when an initial viewpoint can already observe the task region sufficiently, changes in performance caused by a further small viewpoint shift may reflect random variation in the policy itself more than improved information accessibility. For these information-limited initial viewpoints, InfoField3D recommends an optimized viewpoint in the local neighborhood and limits its angular displacement from the starting point to within $20^\circ$ to prevent a large viewpoint change from substantially altering the camera distribution.

Under these rules, we obtain 697 information-limited initial viewpoints, referred to below as trapped viewpoints. For each trapped viewpoint, we evaluate both the starting viewpoint and the viewpoint recommended by InfoField3D. At the same time, to test whether the recommended movement direction is better than other local directions, we construct a trident probe: while keeping the distance from the starting point unchanged, we rotate the recommended movement direction by $\pm120^\circ$ to obtain two equidistant control viewpoints, as shown in \hyperref[fig:app-trident-probe]{Figure \ref*{fig:app-trident-probe}}. Because the three endpoints have the same movement magnitude and all lie in a local region near the starting point, this design can control changes in camera distribution to a substantial extent, allowing comparisons among directions to reflect differences in task-relevant information accessibility more directly.

The experiment uses a frozen $\pi0.5$ as the VLA policy and replaces only its third-person camera viewpoint. For the starting viewpoint, recommended viewpoint, and two control viewpoints in each trident probe, VLA rollouts are conducted with 5 different random seeds. In total, 2,788 viewpoint locations are evaluated, producing 13,940 VLA execution results.

\begin{figure}[H]

\centering
\begin{minipage}[c]{0.38\textwidth}
\includegraphics[width=\linewidth]{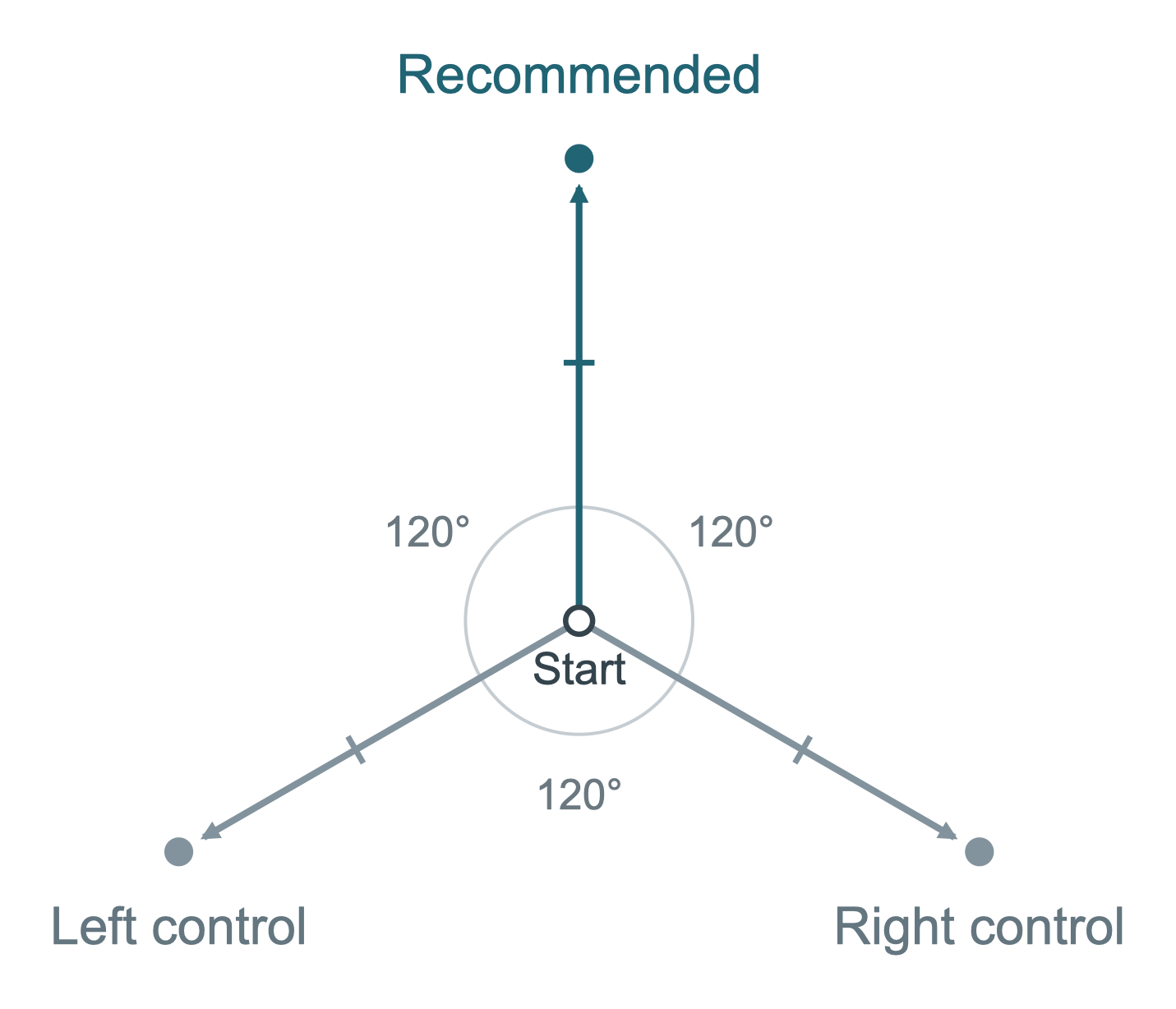}
\end{minipage}\hfill
\begin{minipage}[c]{0.58\textwidth}
\caption{Local trident-probe schematic. Three equal-length directions extend from the starting viewpoint: the InfoField3D recommendation and two controls rotated by $+120^\circ$ and $-120^\circ$. The drawing illustrates their local directional relationship; the observation-sphere constraint is described in the text.}\label{fig:app-trident-probe}
\end{minipage}
\end{figure}

To analyze changes in the downstream performance of the VLA more finely, this experiment specifically examines the completion degree of each VLA execution. Completion-degree scores are determined by the specific task type, using the criteria shown in \hyperref[tab:app-completion-criteria]{Table \ref*{tab:app-completion-criteria}}.

\begingroup
\footnotesize
\setlength{\tabcolsep}{4pt}
\setlength{\LTpre}{0.5em}
\setlength{\LTpost}{0.5em}
\begin{longtable}{@{}>{\raggedright\arraybackslash}p{0.15\textwidth}*{3}{>{\raggedright\arraybackslash}p{\dimexpr0.283333\textwidth-2\tabcolsep\relax}}@{}}
\caption{Quantitative scoring criteria for task completion degree.}\label{tab:app-completion-criteria}\\
\toprule
Task type & 25 points & 50 points & 75 points \\
\midrule
\endfirsthead
\caption*{\textbf{Table \thetable\ (continued).} Quantitative scoring criteria for task completion degree.}\\
\toprule
Task type & 25 points & 50 points & 75 points \\
\midrule
\endhead
Single-object grasp and placement & The left or right finger group makes real contact with the correct source object, with contact distance no greater than 0. & Both finger groups are in contact, and the height of the source-object root is at least 2 cm above its initial height for 3 consecutive control steps. & The 50-point stage has been reached, and the source object is currently within the valid placement range of the target region, but the full official success criterion has not yet been met. \\
Placement into a target region & The correct fingers contact the source object. & Two-finger contact is established and the object is lifted by at least 2 cm for 3 consecutive control steps. & The object has previously been lifted, and it is currently within the horizontal extent of the official region and above the effective target surface. \\
Multi-object placement & Any not-yet-completed object is contacted by the correct fingers. & If only one object remains incomplete, that object is supported by both fingers and lifted for 3 consecutive control steps; if multiple objects remain, at least one native placement subgoal becomes true. & Targets satisfied initially remain satisfied; at least one active target is currently true; and the other objects must previously have been lifted correctly and are currently in deliverable positions. \\
Articulation operation & The correct movable part is contacted; contact with the cabinet body, base, or another drawer does not count. & Progress from the initial joint value to the native success boundary reaches $1/3$ for 3 consecutive control steps. & Progress reaches $2/3$ for 3 consecutive control steps. \\
Pushing into a target region & The correct palm or fingers contact the target object. & Planar distance to the official target region decreases by at least $1/3$, while planar displacement of the source object is at least $5\,\mathrm{mm}$ and its height change is no more than $1\,\mathrm{cm}$, for 3 consecutive control steps. & The same conditions hold, but the planar distance decreases by at least $2/3$, for 3 consecutive control steps. \\
Switch-and-placement combination & The correct source object or movable part is contacted. & The native switch subgoal is completed, or the prescribed sustained lifting or articulation progress is achieved. & The placement, lifting, delivery geometry, and maintenance of initially satisfied targets are all satisfied simultaneously. \\
\midrule
\multicolumn{4}{p{0.97\textwidth}}{The stages in this table represent task-specific discrete execution progress rather than the physical proportion of the task completed; 100 points means that the official task-success criterion is met.} \\
\bottomrule
\end{longtable}
\endgroup

\clearpage
\subsection{Execution Examples}
\label{app:vla-execution-examples}

Figure \hyperref[fig:app-vla-execution-examples]{\ref*{fig:app-vla-execution-examples}}, including its continuations, collects eight execution examples. In each image strip, row (a) uses the initial third-person viewpoint, and row (b) uses the recommended viewpoint from the same scene; the inset shows the first-person camera observation, and the yellow overlay marks the task-relevant region.

\begin{figure}[H]

\centering
\includegraphics[width=\textwidth]{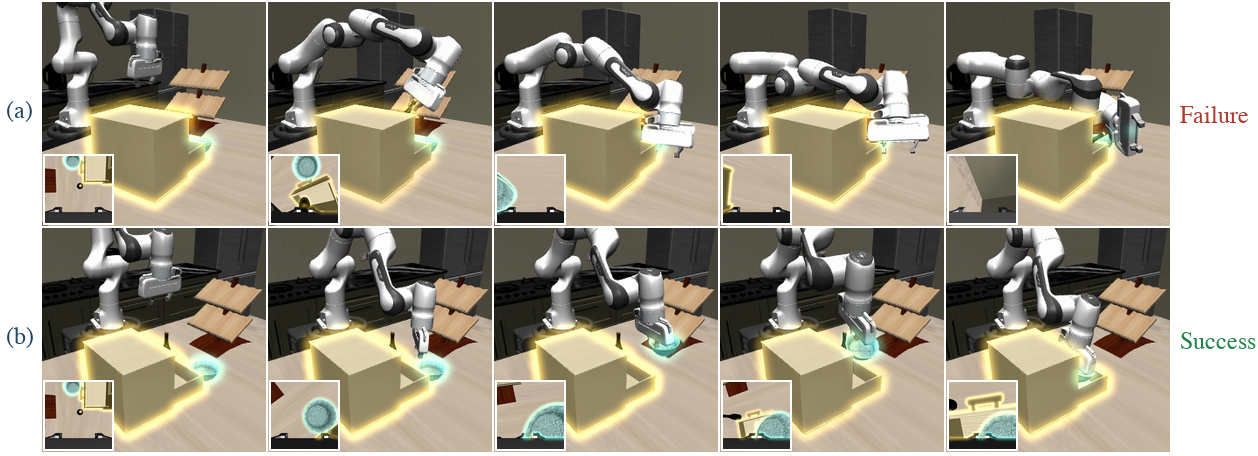}
\caption*{(a) LIBERO-10, scene 1976.}
\includegraphics[width=\textwidth]{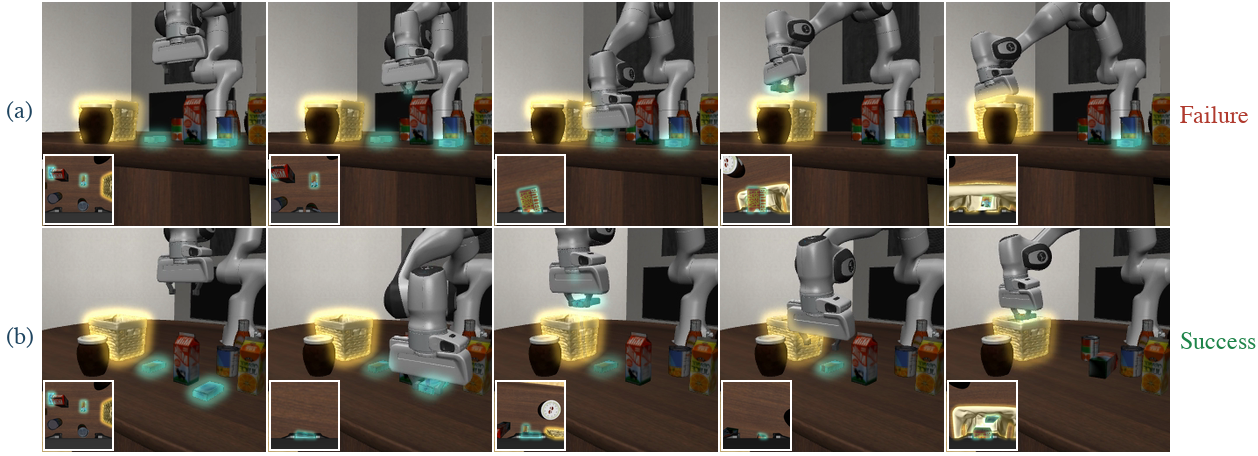}
\caption*{(b) LIBERO-10, scene 2112, starting point g05.}
\caption{Paired execution examples. Within each case, image row (a) uses the initial third-person viewpoint and row (b) uses the recommended viewpoint. Insets show wrist-camera observations; yellow overlays mark task-relevant regions. Case labels (a)--(h) run across this figure and its continuations.}\label{fig:app-vla-execution-examples}
\end{figure}

\clearpage
\begin{figure}[H]
\ContinuedFloat
\captionsetup{labelsep=space}

\centering
\includegraphics[width=\textwidth]{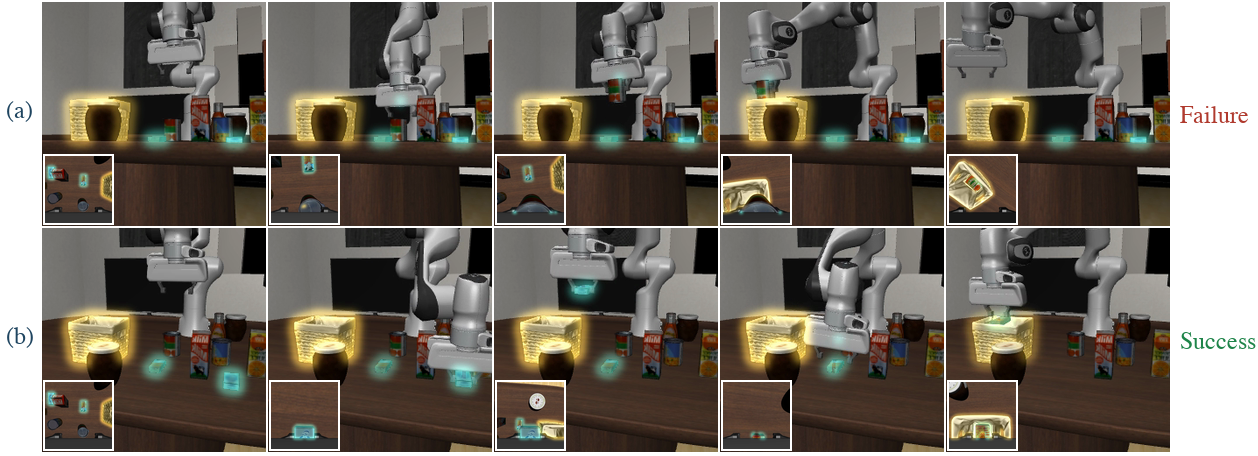}
\caption*{(c) LIBERO-10, scene 2112, starting point g07.}
\includegraphics[width=\textwidth]{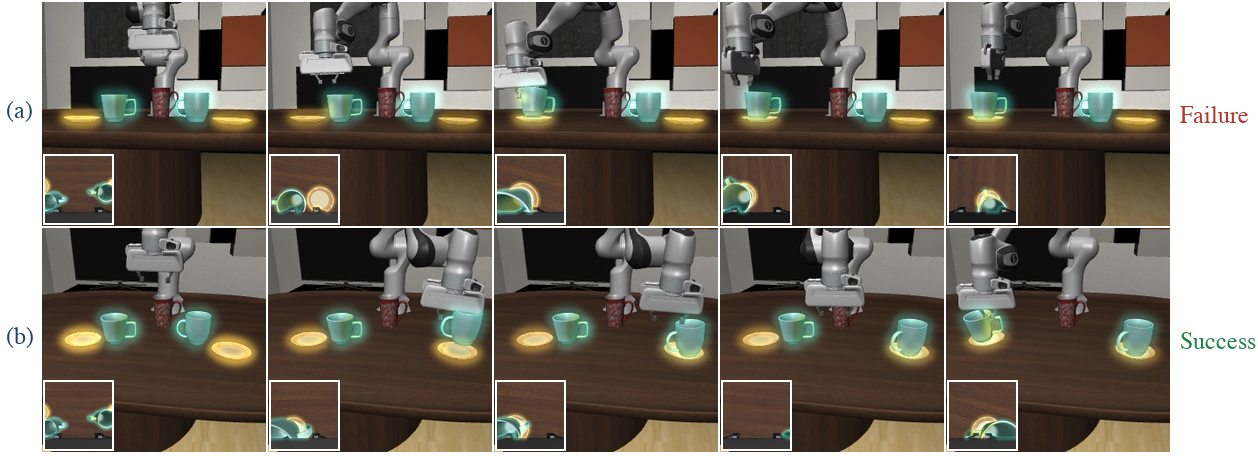}
\caption*{(d) LIBERO-10, scene 2156.}
\includegraphics[width=\textwidth]{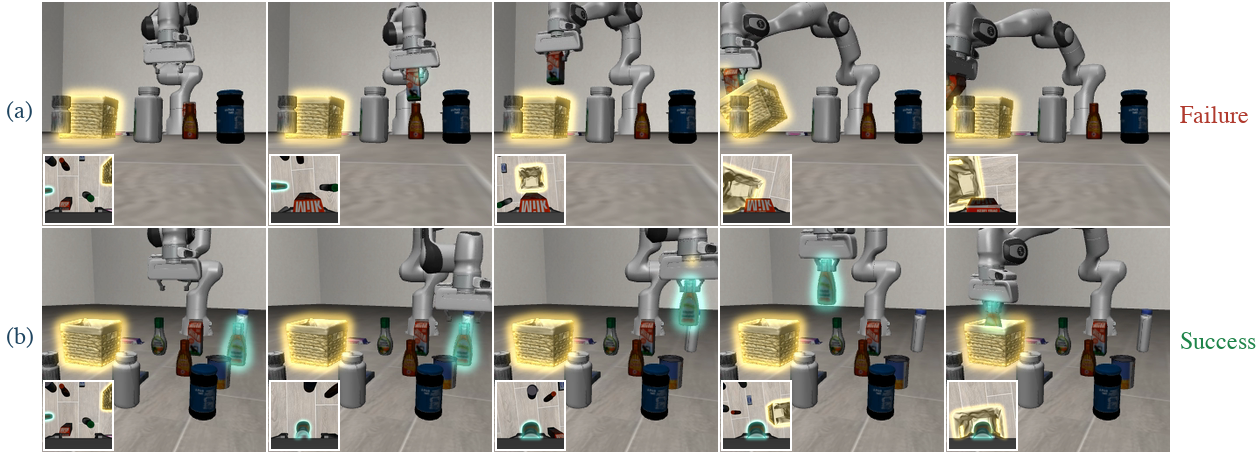}
\caption*{(e) LIBERO-Object, scene 2015.}
\caption{\textbf{(continued).} Paired execution examples, cases (c)--(e).}\label{fig:app-vla-execution-examples-middle}
\end{figure}

\clearpage
\begin{figure}[H]
\ContinuedFloat
\captionsetup{labelsep=space}

\centering
\includegraphics[width=\textwidth]{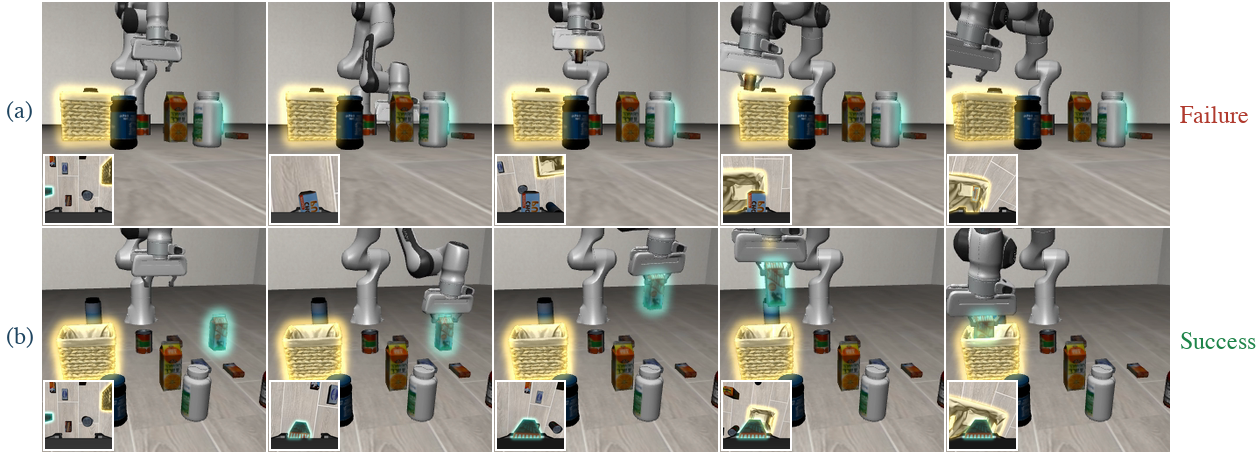}
\caption*{(f) LIBERO-Object, scene 2051, starting point g02.}
\includegraphics[width=\textwidth]{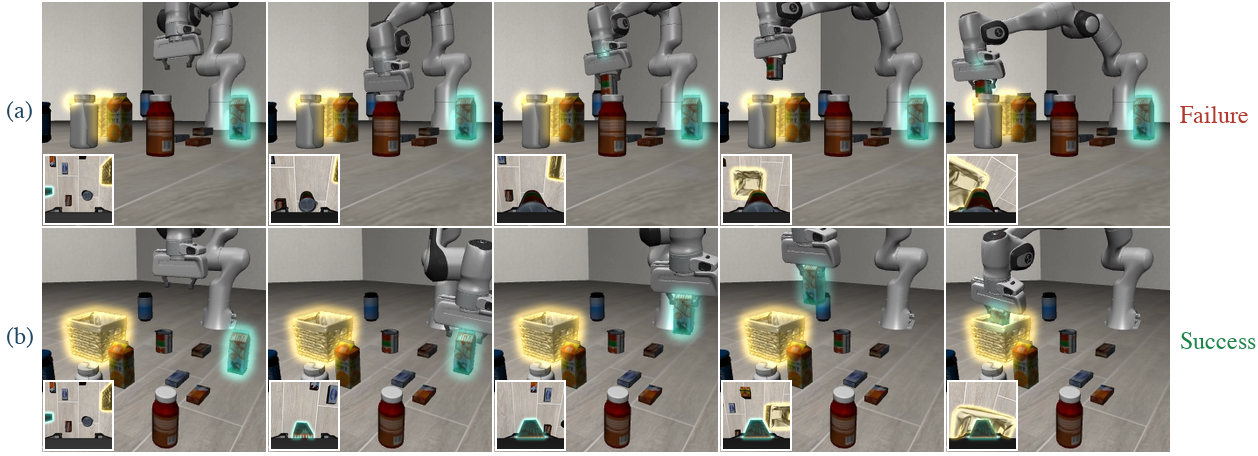}
\caption*{(g) LIBERO-Object, scene 2051, starting point g07.}
\includegraphics[width=\textwidth]{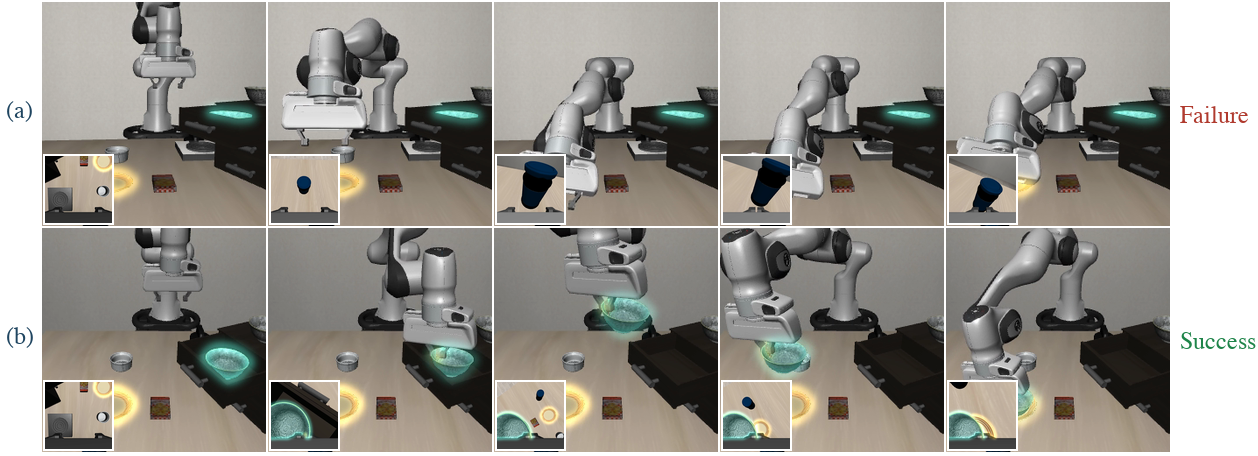}
\caption*{(h) LIBERO-Spatial, scene 1830.}
\caption{\textbf{(continued).} Paired execution examples, cases (f)--(h).}\label{fig:app-vla-execution-examples-final}
\end{figure}

\endgroup

\clearpage
\begingroup
\setcounter{secnumdepth}{3}
\renewcommand{\thesection}{\Alph{section}}
\renewcommand{\thesubsection}{\thesection.\arabic{subsection}}
\renewcommand{\thesubsubsection}{\thesubsection.\arabic{subsubsection}}
\counterwithin{figure}{section}
\counterwithin{table}{section}
\renewcommand{\thefigure}{\thesection\arabic{figure}}
\renewcommand{\thetable}{\thesection\arabic{table}}
\renewcommand{\theHsection}{appendix.\Alph{section}}
\renewcommand{\theHsubsection}{\theHsection.\arabic{subsection}}
\renewcommand{\theHsubsubsection}{\theHsubsection.\arabic{subsubsection}}
\renewcommand{\theHfigure}{appendix.\thesection.\arabic{figure}}
\renewcommand{\theHtable}{appendix.\thesection.\arabic{table}}
\captionsetup{labelfont=bf,labelsep=period}
\makeatletter
\def\@seccntformat#1{%
  \ifcsname app@heading@#1\endcsname
    \csname app@heading@#1\endcsname
  \else
    \csname the#1\endcsname\quad
  \fi
}
\def\app@heading@section{Appendix~\thesection:\quad}
\makeatother

\raggedbottom
\setlength{\parindent}{0pt}
\setlength{\parskip}{0.55em}
\renewcommand{\arraystretch}{1.15}

\providecommand{\appFGHfigdir}{.}

\newcommand{\FGHref}[1]{\hyperref[#1]{\ref*{#1}}}
\newcommand{\FGHapp}[1]{\hyperref[#1]{Appendix~\ref*{#1}}}
\newcommand{\FGHtab}[1]{\hyperref[#1]{Table~\ref*{#1}}}
\newcommand{\FGHfig}[1]{\hyperref[#1]{Figure~\ref*{#1}}}

\color{blue}
\captionsetup{font={color=blue}}
\section{Design Analysis of the Field}
\label{app:design-analysis}

\hyperref[sec:linked-3]{Section~3} of the main text presents the construction of InfoField3D, and \hyperref[sec:linked-4]{Section~4} evaluates it as a whole. This appendix further analyzes this design from three perspectives. \FGHref{app:component-ablation} examines the contribution of each of the three construction components to the recommended viewpoints. \FGHref{app:constraint-complexity} examines whether search over this field remains efficient when the constraints of the feasible domain become complex. \FGHref{app:self-occlusion} examines whether it is reasonable to count other units of the same object as occluders. Unless otherwise specified, the experiments use the 80 scenes in \FGHtab{tab:app-scene-list}, and the evaluation metrics are defined in \FGHtab{tab:app-evaluation-metrics}.

\subsection{Contribution of Each Construction Component}
\label{app:component-ablation}

The field of InfoField3D is jointly constituted by three designs: representing occlusion with continuous occupancy integrals (\hyperref[sec:linked-3-3]{Section~3.3}), encoding directional responses with low-order spherical harmonics (\hyperref[sec:linked-3-2]{Section~3.2}), and partitioning objects into local geometric units (\hyperref[sec:linked-3-4]{Section~3.4}). The experiments in the main text evaluate them as a whole, and Appendices \FGHref{app:harmonic-degree-error} and \FGHref{app:geometric-partitioning} analyze only two of them, at the level of representation error. Direct evidence is still lacking for how much each of them contributes to the final recommended viewpoints. We therefore conducted a component ablation: each time, only one component is removed or replaced while the rest remain unchanged, and we observe how the quality of the recommended viewpoints and the query cost change.

The three variants correspond to the three components. The first variant removes continuous occlusion modeling by setting the transmittance $T_j(\mathbf x)$ of all units to 1, so that the field retains only the directional response. The second variant does not perform spherical harmonic compression and directly uses the kernel sum over the mapped points, $F_{\mathrm{sph}}$, in place of $F_L$, to test whether the compactness of spherical harmonic encoding comes at the cost of quality. The third variant removes local geometric partitioning and represents each object instance with only one unit. The recommended viewpoints are evaluated with Task-region visibility and Directional visibility response. These two metrics are computed exactly with hard rays and do not go through any approximation of the field, so the variants can be compared fairly.

\begin{table}[H]
\color{blue}
\centering
\caption{Ablation results of the field construction components. Values are ratios relative to the full method (taken as 100\%). Viewpoint quality is taken at 20\,s of search. Per-query time is first taken as the median within each scene and then as the median across scenes.}\label{tab:app-component-ablation}
\small
\begin{tabularx}{\textwidth}{@{}>{\raggedright\arraybackslash}X*{2}{>{\centering\arraybackslash}p{2.9cm}}>{\centering\arraybackslash}p{1.7cm}@{}}
\toprule
Variant & Task-region visibility & Directional visibility response & Per-query time \\
\midrule
Full method & 100\% & 100\% & 100\% \\
Without continuous occlusion modeling & 81.65\% & 85.96\% & 96.66\% \\
Pointwise computation in place of spherical harmonic encoding & 99.98\% & 100.00\% & 182.19\% \\
Without local geometric partitioning & 97.00\% & 94.14\% & 75.94\% \\
\bottomrule
\end{tabularx}
\end{table}

\FGHtab{tab:app-component-ablation} shows that the three components play different roles. Occlusion modeling has the largest impact on viewpoint quality. After it is removed, Task-region visibility drops to 81.65\% of the full method, and Directional visibility response drops to 85.96\%. Without the occlusion term, the field also treats directions blocked by other objects as high-response directions, so the recommended viewpoints are more likely to fail to see the task region.

The role of spherical harmonic encoding is reflected in efficiency. After switching to pointwise computation, both metrics remain essentially unchanged, indicating that the degree-4 truncation loses almost no information useful for viewpoint selection, which is consistent with the analysis in \FGHapp{app:harmonic-degree-error}. However, the per-query time of direct pointwise summation is $1.82\times$ that of spherical harmonic encoding, which shows that spherical harmonic encoding substantially improves efficiency while preserving information well. 

Local geometric partitioning mainly affects directional information. After partitioning is removed, Directional visibility response drops to 94.14\% and Task-region visibility drops to 97.00\%, indicating that local partitioning contributes to both metrics, with a larger effect on directional information. This agrees with the observation in \FGHapp{app:geometric-partitioning}: when a whole object is represented with a single center, the correspondence between normal and position is distorted in the spherical mapping, and directional details are the first to be lost. Because the number of units decreases, the per-query time of this variant drops to 75.94\%, showing that local partitioning trades a certain amount of computation for directional fidelity.

Overall, occlusion modeling contributes most to whether the recommended viewpoints can see the task region, local geometric partitioning makes them see it more frontally, and spherical harmonic encoding makes these computations sufficiently compact and efficient.

\subsection{Search Cost under Increasing Constraint Complexity}
\label{app:constraint-complexity}

InfoField3D uniformly represents practical constraints such as camera reachability as the shape of the feasible domain $\Omega$ (\hyperref[sec:linked-3-1]{Section~3.1}). The experiments in the main text use a feasible domain of simple shape, namely a cubic workspace with a spherical region around the task objects removed (\FGHapp{app:search-scenes-domain}). Constraints in practical applications may be much more complex, for example, a large number of obstacles that must be avoided, or reachable space that is fragmented into several narrow regions. A natural question is whether search on the field remains efficient in such cases.

No matter how complex the constraints are, the optimization variable of InfoField3D is always the 3D observation position; the constraints change only the shape of the feasible domain and do not increase the dimensionality of the problem. We therefore examine three kinds of changes in the shape of the feasible domain, each corresponding to a mechanism that may slow down the search.

\begin{itemize}[leftmargin=1.4em,itemsep=0.25em,topsep=0.2em]
\item \textbf{Number of keep-out zones.} As the number of constraints increases, the number of conditions to be checked for each query increases accordingly. We add spherical keep-out zones within the feasible domain, at four levels: 0, 50,  500,  and 5,000.
\item \textbf{Feasible-domain volume.} As feasible positions become scarce, sampling-based methods need more attempts to obtain a feasible candidate. We progressively shrink the feasible domain, retaining 100\%, 50\%,  20\%,  and 10\% of it, for four levels in total.
\item \textbf{Feasible-domain connectivity.} Once the feasible domain is split apart, local search cannot move from one region to another. This is particularly worth testing for gradient-based methods. We split the feasible domain into 2/5/10 mutually disconnected regions.
\end{itemize}

We measure search cost by the mean time-to-threshold, i.e., the time needed for the search to first reach 80\% of the best known Task-region visibility for the given scene and constraint level (the best known value is the highest value achieved by all methods in the same scene under the same constraint setting). Runs that have not reached the threshold within the maximum time (20\,s) are counted as 20\,s. The experimental results are shown in Tables \FGHref{tab:app-forbidden-region-time} and \FGHref{tab:app-feasible-domain-ratio}.

\begin{table}[H]
\color{blue}
\centering
\caption{Mean time-to-threshold (s) under different numbers of keep-out zones.}
\label{tab:app-forbidden-region-time}
\begin{tabularx}{\textwidth}{@{}X*{3}{>{\raggedleft\arraybackslash}p{3.2cm}}@{}}
\toprule
Method & No keep-out zones (s) & 5,000 keep-out zones (s) & Change (s) \\
\midrule
\textbf{InfoField3D (SHGO)} & \textbf{0.00960} & \textbf{0.02290} & \textbf{+0.01330} \\
GP-BO & 0.66680 & 0.70559 & +0.03879 \\
Sobol & 0.67075 & 0.79581 & +0.12506 \\
Random & 0.73543 & 0.83573 & +0.10030 \\
CMA-ES & 3.61458 & 3.29755 & $-0.31703$ \\
CIM & 7.86470 & 7.87240 & +0.00769 \\
GradientNBV & 9.43402 & 9.68389 & +0.24987 \\
VISO-grasp & 10.52613 & 10.80199 & +0.27586 \\
\bottomrule
\end{tabularx}
\end{table}

As shown in \FGHtab{tab:app-forbidden-region-time}, when the number of
keep-out zones increases from 0 to 5,000, the mean time-to-threshold of
InfoField3D (SHGO) increases from 0.00960\,s to 0.02290\,s, an increase
of 0.01330\,s. At 5,000 keep-out zones, its mean time-to-threshold is
approximately 3.2\% of that of the fastest baseline, GP-BO
(0.70559\,s). The changes across the baselines are not uniform: CIM
shows a smaller increase of 0.00769\,s, while the mean time-to-threshold
of CMA-ES decreases by 0.31703\,s. Thus, InfoField3D retains a low
absolute time-to-threshold under 5,000 keep-out zones, although its
time increase is not the smallest among all methods.

\begin{table}[H]
\color{blue}
\centering
\caption{Mean time-to-threshold relative to that of the same method
without additional constraints.}
\label{tab:app-feasible-domain-ratio}
\begin{tabularx}{\textwidth}{@{}X*{2}{>{\raggedleft\arraybackslash}p{4.2cm}}@{}}
\toprule
Method & Volume retained at 10\% & 10 mutually disconnected regions \\
\midrule
\textbf{InfoField3D (SHGO)} & \textbf{1.140$\times$} & \textbf{1.177$\times$} \\
GP-BO & 3.716$\times$ & 2.021$\times$ \\
Sobol & 3.884$\times$ & 1.977$\times$ \\
Random & 3.616$\times$ & 1.714$\times$ \\
CMA-ES & 3.000$\times$ & 2.908$\times$ \\
CIM & 1.054$\times$ & 0.760$\times$ \\
GradientNBV & 1.610$\times$ & 1.469$\times$ \\
VISO-grasp & 1.461$\times$ & 1.394$\times$ \\
\bottomrule
\end{tabularx}
\end{table}

When the feasible domain retains 10\% of its original volume, the mean
time-to-threshold of InfoField3D (SHGO) is $1.140\times$ its result
without additional constraints. The corresponding multiples for the
baselines range from $1.054\times$ to $3.884\times$.

When the feasible domain is split into 10 mutually disconnected regions,
the mean time-to-threshold of InfoField3D (SHGO) is $1.177\times$ its
result without additional constraints. The corresponding multiples for
the baselines range from $0.760\times$ to $2.908\times$. In particular,
CIM's mean time-to-threshold decreases under this setting.

Taken together with \FGHtab{tab:app-forbidden-region-time}, these results
show that InfoField3D maintains a low absolute time-to-threshold with
5,000 keep-out zones and has relative time increases of 14.0\% and
17.7\% under the two feasible-domain settings, respectively. Its
relative increase is not the smallest among all methods in either
setting.

To provide a more detailed view of the results, the following tables
report the mean time-to-threshold at levels of keep-out-zone
count, retained feasible-domain volume, and number of disconnected
regions, respectively.

\begin{table}[H]
\color{blue}
\centering
\caption{Mean time-to-threshold (s) at selected numbers of keep-out
zones.}
\label{tab:app-forbidden-region-levels}
\begin{tabularx}{\textwidth}{@{}X*{4}{>{\raggedleft\arraybackslash}p{2.3cm}}@{}}
\toprule
Method & 0 & 50 & 500 & 5,000 \\
\midrule
\textbf{InfoField3D (SHGO)} & \textbf{0.00960} & \textbf{0.01091} & \textbf{0.01197} & \textbf{0.02290} \\
GP-BO & 0.66680 & 0.19453 & 0.17218 & 0.70559 \\
Sobol & 0.67075 & 0.14118 & 0.13820 & 0.79581 \\
Random & 0.73543 & 0.21656 & 0.15675 & 0.83573 \\
CMA-ES & 3.61458 & 1.32371 & 2.98780 & 3.29755 \\
CIM & 7.86470 & 16.12871 & 9.80278 & 7.87240 \\
GradientNBV & 9.43402 & 10.05327 & 5.07252 & 9.68389 \\
VISO-grasp & 10.52613 & 15.00170 & 10.00181 & 10.80199 \\
\bottomrule
\end{tabularx}
\end{table}

\begin{table}[h]
\color{blue}
\centering
\caption{Mean time-to-threshold (s) at selected retained feasible-domain
volumes.}
\label{tab:app-feasible-volume-levels}
\begin{tabularx}{\textwidth}{@{}X*{4}{>{\raggedleft\arraybackslash}p{2.3cm}}@{}}
\toprule
Method & 100\% & 50\% & 20\% & 10\% \\
\midrule
\textbf{InfoField3D (SHGO)} & \textbf{0.00960} & \textbf{0.01183} & \textbf{0.01219} & \textbf{0.01094} \\
GP-BO & 0.66680 & 0.32895 & 0.16583 & 2.47783 \\
Sobol & 0.67075 & 0.15427 & 0.51839 & 2.60519 \\
Random & 0.73543 & 2.21481 & 0.19269 & 2.65931 \\
CMA-ES & 3.61458 & 3.40364 & 15.22048 & 10.84374 \\
CIM & 7.86470 & 8.61173 & 15.45006 & 8.28939 \\
GradientNBV & 9.43402 & 15.04711 & 20.00000 & 15.18877 \\
VISO-grasp & 10.52613 & 15.00083 & 20.00000 & 15.37868 \\
\bottomrule
\end{tabularx}
\end{table}

\begin{table}[h]
\color{blue}
\centering
\caption{Mean time-to-threshold (s) at selected numbers of disconnected
feasible regions.}
\label{tab:app-disconnected-region-levels}
\begin{tabularx}{\textwidth}{@{}X*{4}{>{\raggedleft\arraybackslash}p{2.3cm}}@{}}
\toprule
Method & No additional constraints & 2 regions & 5 regions & 10 regions \\
\midrule
\textbf{InfoField3D (SHGO)} & \textbf{0.00960} & \textbf{0.01047} & \textbf{0.01067} & \textbf{0.01130} \\
GP-BO & 0.66680 & 0.18137 & 0.35532 & 1.34760 \\
Sobol & 0.67075 & 0.19424 & 0.91920 & 1.32607 \\
Random & 0.73543 & 0.18367 & 0.63162 & 1.26053 \\
CMA-ES & 3.61458 & 0.55612 & 10.44359 & 10.51120 \\
CIM & 7.86470 & 6.12998 & 7.19780 & 5.97717 \\
GradientNBV & 9.43402 & 10.00594 & 20.00000 & 13.85858 \\
VISO-grasp & 10.52613 & 15.00043 & 20.00000 & 14.67343 \\
\bottomrule
\end{tabularx}
\end{table}

\FloatBarrier
\subsection{Self-Occlusion of Objects}
\label{app:self-occlusion}

After \hyperref[sec:linked-3-4]{Section~3.4} partitions a complex object into multiple local units, it must be decided whether the other units of the same object participate in the occlusion computation. InfoField3D excludes only the source unit itself, and the other units of the same object still act as occluders. The rationale is that occlusion within an object does exist in actual observation; for example, the seat cannot be seen from behind the chair's backrest. However, this also raises a concern: occlusion caused by the object's own structure lowers the field value in the same way as external occlusion, which might push the viewpoint away from reasonable observation positions in pursuit of an unnatural occlusion-free viewpoint. We test this with two sets of experiments.

\textbf{Experiments on inducing unnatural viewpoints.} The first set of experiments targets a local part of an object and deliberately constructs the situation that is relatively prone to problems. We take the chair seat as the observation target, place the chair facing the wall and almost touching it, and place a shelf above it. In this case, the front of the seat is blocked by the wall, the top by the shelf, and the back by the chair's own backrest, leaving very little space for reasonable viewpoints, to observe whether the occlusion penalty caused by the chair backrest leads the viewpoint recommendation to an unreasonable position.

\begin{figure}[H]
\centering
\includegraphics[width=0.8\textwidth]{\appFGHfigdir/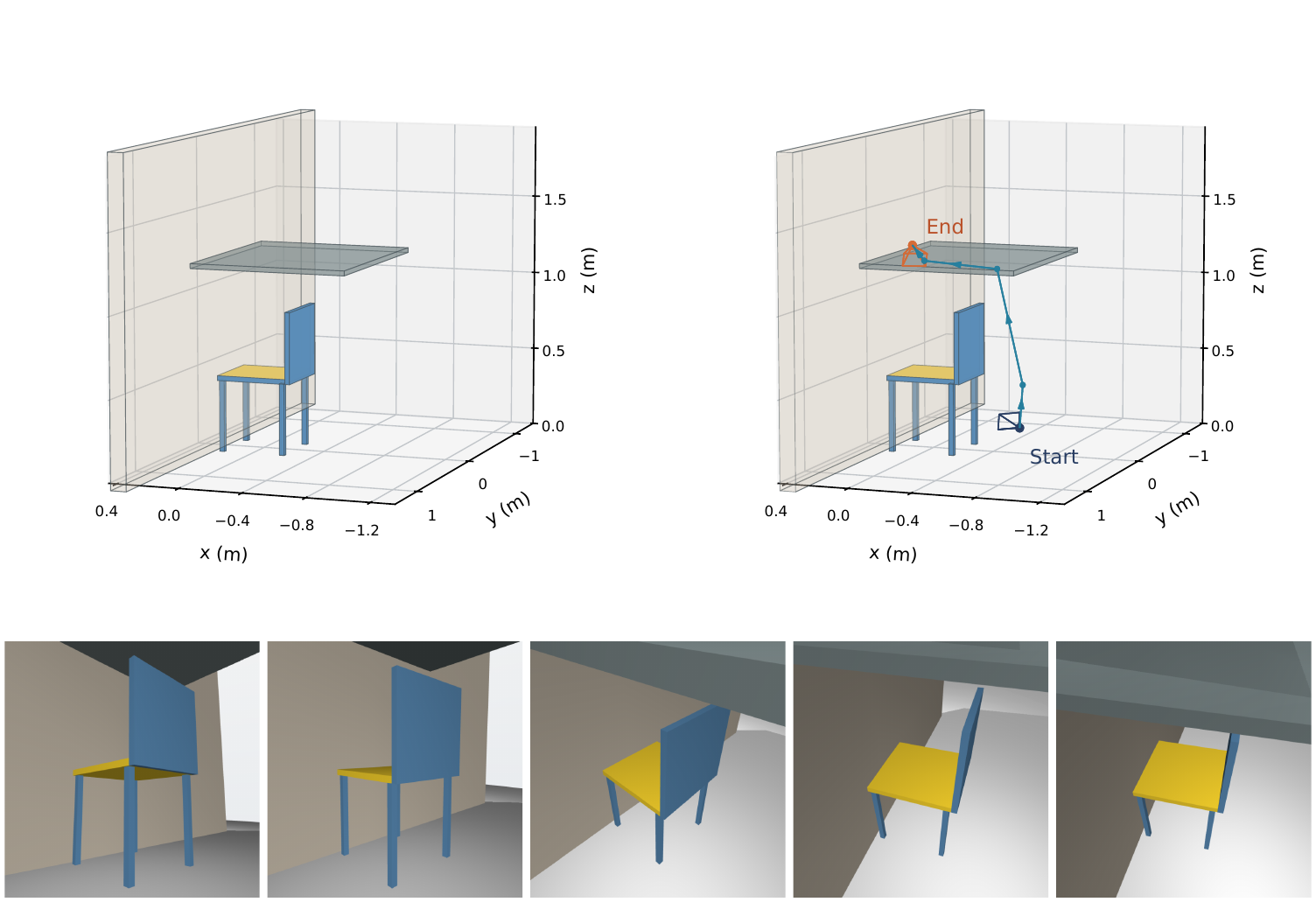}
\caption{Chair-seat-against-wall experiment. Top: scene layout (left) and the search trajectory from the starting viewpoint to the recommended viewpoint (right). Bottom: RGB renderings at successive search iterations, from the starting viewpoint (left) to the recommended viewpoint (right).}\label{fig:app-chair-seat-wall}
\end{figure}

InfoField3D recommends a viewpoint from above and to the side (\FGHfig{fig:app-chair-seat-wall}). Verified by an independent hard-ray check, the visible fraction of the seat at this viewpoint is 100.0\%, and the viewpoint avoids both external occlusion and occlusion by the backrest. This is because the Gaussian occupancy of each unit is concentrated within its own local geometric extent and decays rapidly beyond it. As long as the line of sight does not pass through the region of a unit, the occlusion accumulated from that unit is limited, so a clear decrease in the field value usually corresponds to the task region actually being blocked, rather than merely to the line of sight passing close to the object's own structure.

\textbf{Whole object as the target.} The second set of experiments examines whether counting self-occlusion leads to unnatural viewpoints when the whole object is the target. We select four objects, namely a chair, a handled mug, an open organizer, and a concave bowl, take all surfaces of the object as a uniformly weighted task region, and search for the best viewpoint separately with and without counting the self-occlusion of the same object. The result without self-occlusion serves only as a reference for whether counting self-occlusion changes the recommended viewpoint.

\begin{figure}[h]
\centering
\includegraphics[width=0.45\textwidth]{\appFGHfigdir/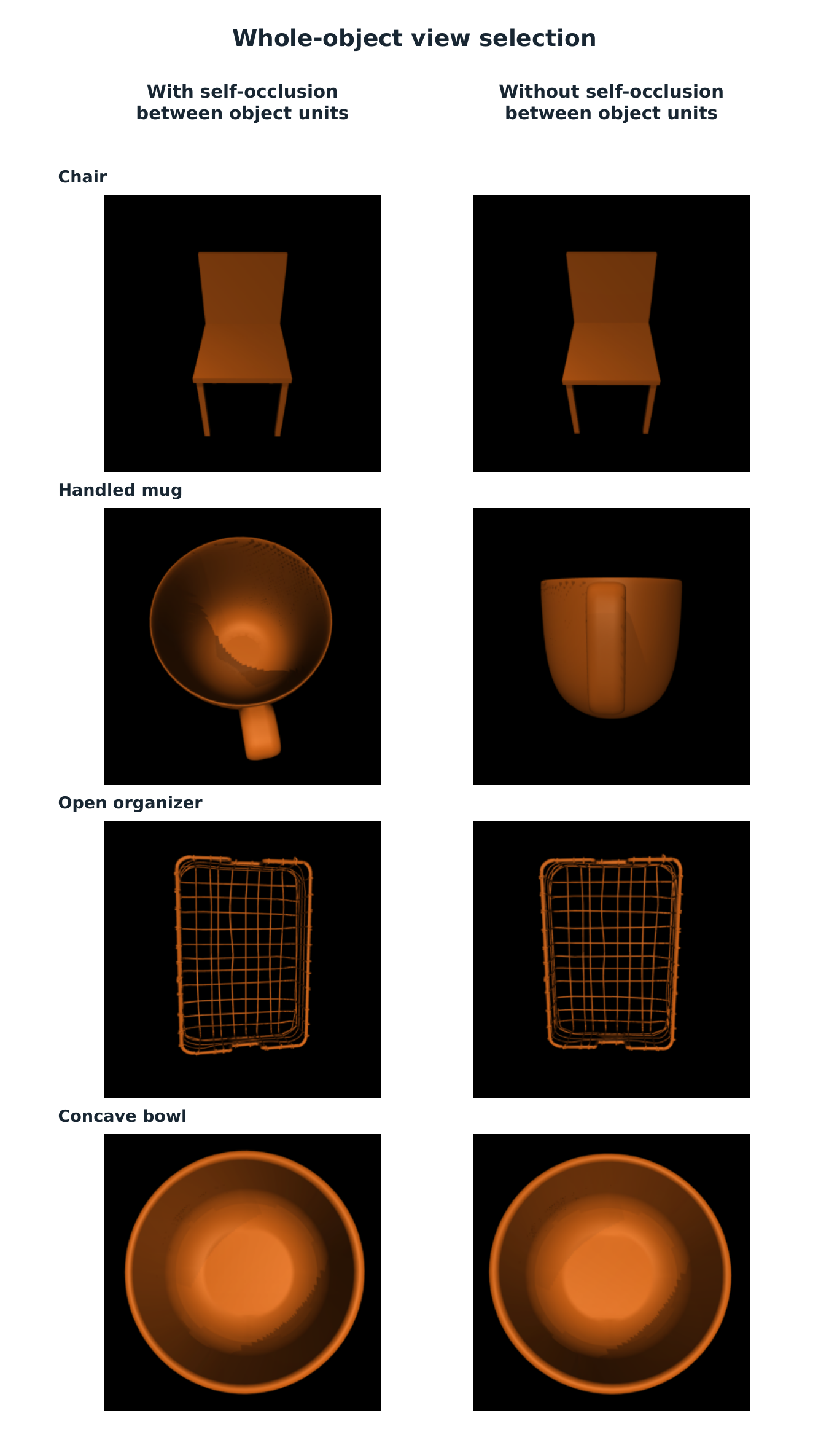}
\caption{Recommended viewpoints for the four objects, with (left column) and without (right column) self-occlusion between units of the same object.}\label{fig:app-whole-object-self-occlusion}
\end{figure}

\begin{table}[h]
\color{blue}
\centering
\caption{Hard-ray visible fraction (\%) of the recommended viewpoints in the two cases. Values are the hard-ray visible area as a fraction of the total surface area of the object.}\label{tab:app-self-occlusion-visibility}
\begin{tabular}{lrr}
\toprule
Object & With self-occlusion & Without self-occlusion \\
\midrule
Chair & 42.63 & 42.59 \\
Handled mug & 44.74 & 17.03 \\
Open organizer & 40.44 & 42.14 \\
Concave bowl & 48.55 & 48.55 \\
\bottomrule
\end{tabular}
\end{table}

With self-occlusion counted, natural viewpoints are recommended for all four objects (\FGHfig{fig:app-whole-object-self-occlusion} and \FGHtab{tab:app-self-occlusion-visibility}). For the chair, the open organizer, and the bowl, the viewpoints recommended in the two cases are largely consistent. Only the handled mug differs. Without self-occlusion, a side view is recommended, because the field also counts the inner wall, which is actually blocked by the outer wall, as visible. With self-occlusion counted, a top-down view that can truly see the inner wall is recommended. Thus, with or without self-occlusion, the recommended viewpoints are interpretable and generally do not lead to strange results resembling computational errors.

The two sets of experiments show that counting the object's self-occlusion does not push the viewpoint toward unnatural positions. On most objects, it does not change the recommendation, and on objects with pronounced internal occlusion, it avoids counting actually blocked surfaces as visible.

\clearpage
\section{Computational Cost and Additional VLA Evaluation}
\label{app:cost-and-vla}

Two conclusions in the main text depend on the specific measurement or experimental setting. The first is search efficiency, which \hyperref[sec:linked-4-2]{Section~4.2} measures by search time, and time is affected by hardware and implementation. The second is the downstream effect: \hyperref[sec:linked-4-4]{Section~4.4} evaluates the benefit of the recommended viewpoints for the VLA under one controlled setting. This appendix examines whether these two conclusions still hold under a different measurement (\FGHref{app:flops-memory}) and a different setting (\FGHref{app:vla-occluder}), respectively.

\subsection{Computational Cost Measured by FLOPs and Memory}
\label{app:flops-memory}

Search time is a direct measure of efficiency, but it is hard to make the implementation and the degree of hardware utilization fully equal across methods; for example, several baselines run only on the CPU in their standard implementations (\FGHtab{tab:app-flops-memory}). To exclude this factor, we add two measures that are less related to hardware and implementation, namely floating-point operations and peak memory.

\FloatBarrier
The computation needed to reach a given viewpoint quality is determined by two factors: the cost of each query and the number of queries required. We first compare the former, and then use cumulative FLOPs as the budget to compare the combined effect of both. FLOPs count only the basic floating-point operations of the queries themselves, excluding optimizer and program-control overhead. Cumulative FLOPs are the sum of the FLOPs of all queries during the search.

\begin{table}[h]
\color{blue}
\centering
\caption{Per-query FLOPs and peak memory. MFLOPs denotes $10^6$ FLOPs. Values are means across scenes. GPU memory is the peak actually allocated, and process memory is the peak RSS. ``CPU only’’ indicates that the standard implementation of the method does not use the GPU.}\label{tab:app-flops-memory}
\begin{tabularx}{\textwidth}{@{}X*{3}{>{\raggedleft\arraybackslash}p{3.2cm}}@{}}
\toprule
Method & Per query (MFLOPs) & Peak GPU memory (MiB) & Peak process memory (GiB) \\
\midrule
InfoField3D & 11.416 & 29.685 & 1.1215 \\
GP-BO & 311.185 & CPU only & 1.4647 \\
CIM & 311.185 & CPU only & 1.4624 \\
CMA-ES & 311.185 & CPU only & 1.4691 \\
GradientNBV & 399.608 & 135.191 & 1.3439 \\
VISO-grasp & 0.002 & CPU only & 0.0317 \\
Sobol & 311.185 & CPU only & 1.4622 \\
Random & 311.185 & CPU only & 1.4617 \\
\bottomrule
\end{tabularx}
\end{table}

The per-query cost of InfoField3D is about 11.4 MFLOPs, which is 1/35 to 1/27 of that of the ray-based scoring baselines. In terms of memory, the peak GPU memory of InfoField3D is about 30\,MiB, 1/4.5 of that of GradientNBV, and its peak process memory is about 1.12\,GiB, lower than the 1.34 to 1.47\,GiB of the ray-based baselines. VISO-grasp is an exception and is not considered in the above discussion: it relies only on bounding boxes and cannot take the pointwise task-relevance map as input; therefore, it is very fast, but its viewpoint quality is low; see \FGHtab{tab:app-flops-task-visibility}.

Tables \FGHref{tab:app-flops-task-visibility} to \FGHref{tab:app-flops-gpt4o} give the four metrics of each method under different cumulative FLOPs budgets, where ``---'' indicates that no scene produced a complete output under that budget, and averages are computed only over the scenes with outputs. Each method takes the recommended viewpoint from its last update within the budget.

\begin{table}[H]
\color{blue}
\centering
\caption{Task-region visibility (\%) under different cumulative FLOPs budgets. }\label{tab:app-flops-task-visibility}
\small
\setlength{\tabcolsep}{5pt}
\begin{tabular}{lrrrrrr}
\toprule
Method / cumulative FLOPs (MFLOPs) & 100 & 300 & 1,000 & 3,000 & 10,000 & 300,000 \\
\midrule
InfoField3D (SHGO) & \textbf{46.29} & \textbf{47.41} & \textbf{46.30} & \textbf{46.34} & \textbf{46.32} & \textbf{46.32} \\
InfoField3D (Analytic Gradient) & 36.40 & 40.14 & 40.30 & 40.30 & 40.30 & 40.30 \\
InfoField3D (Sampling-Based Gradient) & 15.10 & 31.05 & 39.13 & 39.37 & 39.37 & 39.37 \\
GP-BO & --- & 1.37 & 18.68 & 34.69 & 43.07 & 44.12 \\
CIM surrogate gradient & --- & 0.42 & 2.36 & 5.57 & 12.89 & 30.77 \\
CMA-ES & --- & 0.86 & 5.32 & 15.69 & 30.23 & 40.01 \\
GradientNBV & --- & --- & 1.50 & 9.13 & 18.72 & 19.11 \\
VISO-grasp & 20.54 & 20.54 & 20.54 & 20.54 & 20.54 & 20.54 \\
Sobol & --- & 3.23 & 19.96 & 31.14 & 37.39 & 43.37 \\
Random & --- & 3.00 & 19.94 & 31.81 & 39.44 & 43.20 \\
\bottomrule
\end{tabular}
\end{table}

\begin{table}[H]
\color{blue}
\centering
\caption{Directional visibility response (\%) under different cumulative FLOPs budgets.}\label{tab:app-flops-directional-response}
\small
\setlength{\tabcolsep}{5pt}
\begin{tabular}{lrrrrrr}
\toprule
Method / cumulative FLOPs (MFLOPs) & 100 & 300 & 1,000 & 3,000 & 10,000 & 300,000 \\
\midrule
InfoField3D (SHGO) & \textbf{25.22} & \textbf{26.71} & \textbf{27.43} & \textbf{27.59} & \textbf{27.58} & \textbf{27.58} \\
InfoField3D (Analytic Gradient) & 20.22 & 22.89 & 23.06 & 23.06 & 23.06 & 23.06 \\
InfoField3D (Sampling-Based Gradient) & 7.55 & 16.93 & 21.82 & 22.19 & 22.20 & 22.20 \\
GP-BO & --- & 0.74 & 9.73 & 17.59 & 22.12 & 22.46 \\
CIM surrogate gradient & --- & 0.21 & 1.15 & 2.74 & 6.49 & 15.98 \\
CMA-ES & --- & 0.40 & 2.54 & 7.72 & 15.29 & 20.24 \\
GradientNBV & --- & --- & 0.72 & 4.29 & 8.66 & 8.89 \\
VISO-grasp & 8.87 & 8.87 & 8.87 & 8.87 & 8.87 & 8.87 \\
Sobol & --- & 1.62 & 10.43 & 16.17 & 19.19 & 22.21 \\
Random & --- & 1.42 & 9.88 & 15.89 & 20.22 & 22.17 \\
\bottomrule
\end{tabular}
\end{table}

\begin{table}[h]
\color{blue}
\centering
\caption{InfoField3D score under different cumulative FLOPs budgets.}\label{tab:app-flops-infofield-score}
\small
\setlength{\tabcolsep}{5pt}
\begin{tabular}{lrrrrrr}
\toprule
Method / cumulative FLOPs (MFLOPs) & 100 & 300 & 1,000 & 3,000 & 10,000 & 300,000 \\
\midrule
InfoField3D (SHGO) & \textbf{0.1828} & \textbf{0.1979} & \textbf{0.2081} & \textbf{0.2113} & \textbf{0.2113} & \textbf{0.2113} \\
InfoField3D (Analytic Gradient) & 0.1552 & 0.1764 & 0.1785 & 0.1786 & 0.1786 & 0.1786 \\
InfoField3D (Sampling-Based Gradient) & 0.0685 & 0.1319 & 0.1666 & 0.1709 & 0.1710 & 0.1710 \\
GP-BO & --- & 0.0421 & 0.0991 & 0.1311 & 0.1505 & 0.1497 \\
CIM surrogate gradient & --- & 0.0300 & 0.0393 & 0.0555 & 0.0793 & 0.1177 \\
CMA-ES & --- & 0.0318 & 0.0521 & 0.0823 & 0.1216 & 0.1419 \\
GradientNBV & --- & --- & 0.0371 & 0.0616 & 0.0799 & 0.0809 \\
VISO-grasp & 0.0707 & 0.0707 & 0.0707 & 0.0707 & 0.0707 & 0.0707 \\
Sobol & --- & 0.0420 & 0.1043 & 0.1314 & 0.1402 & 0.1470 \\
Random & --- & 0.0393 & 0.0981 & 0.1289 & 0.1422 & 0.1497 \\
\bottomrule
\end{tabular}
\end{table}

\begin{table}[h]
\color{blue}
\centering
\caption{GPT-4o score under different cumulative FLOPs budgets. Scoring uses the fixed first run of each method in each scene.}\label{tab:app-flops-gpt4o}
\small
\setlength{\tabcolsep}{5pt}
\begin{tabular}{lrrrrrr}
\toprule
Method / cumulative FLOPs (MFLOPs) & 100 & 300 & 1,000 & 3,000 & 10,000 & 300,000 \\
\midrule
InfoField3D (SHGO) & \textbf{80.96} & \textbf{83.06} & \textbf{83.63} & \textbf{84.63} & \textbf{84.33} & 84.33 \\
InfoField3D (Analytic Gradient) & 69.38 & 75.08 & 76.50 & 76.50 & 76.50 & 76.50 \\
InfoField3D (Sampling-Based Gradient) & 29.44 & 53.04 & 65.81 & 66.77 & 66.60 & 66.60 \\
GP-BO & --- & 27.50 & 65.83 & 83.63 & 84.30 & 84.10 \\
CIM surrogate gradient & --- & 24.39 & 29.19 & 43.67 & 61.29 & 76.19 \\
CMA-ES & --- & 26.36 & 38.08 & 61.77 & 76.13 & 82.73 \\
GradientNBV & --- & --- & 26.88 & 43.38 & 51.38 & 53.50 \\
VISO-grasp & 50.58 & 50.58 & 50.58 & 50.58 & 50.58 & 50.58 \\
Sobol & --- & 32.05 & 65.75 & 81.29 & 84.31 & \textbf{84.40} \\
Random & --- & 35.15 & 72.56 & 82.90 & 84.09 & 84.04 \\
\bottomrule
\end{tabular}
\end{table}

With cumulative FLOPs as the budget, the conclusions are the same as with search time as the budget. On Task-region visibility, InfoField3D (SHGO) already approaches its final level at 100 MFLOPs, whereas the best baseline needs 10,000 MFLOPs to reach about 93\% of it. On InfoField3D score and Directional visibility response, InfoField3D reaches 87\% and 91\% of its final values at 100 MFLOPs, respectively, and exceeds 98\% on both at 1,000 MFLOPs, while the best baseline at 10,000 MFLOPs reaches only 71\% and 80\% of InfoField3D's final values.

These experiments, which are less dependent on hardware and implementation optimizations, further demonstrate that InfoField3D's efficiency advantage stems from the method itself: each query incurs a lower computational cost, and fewer queries are needed to achieve the same quality.

\subsection{VLA Evaluation under Another Setting}
\label{app:vla-occluder}

The experiment in \hyperref[sec:linked-4-4]{Section~4.4} starts from information-limited viewpoints selected by hard visibility and tests whether the nearby viewpoints recommended by InfoField3D can improve the execution of the frozen VLA. There, the information limitation comes from the position of the starting point itself. A natural question is whether this conclusion still holds when both the starting point and the source of the information limitation are different. This section examines another common situation: the camera is at the default position most familiar to the policy, but the line of sight is blocked by an extraneous object.

\textbf{Experimental design.} The starting point is fixed at the default third-person camera viewpoint. We place an occluding panel between the default camera and the task center to block the task region. The panel center is located at 40\% of the line from the camera to the task center, with its normal along this line. It is a slender thin plate with a long-to-short side ratio of 8:1 and a thickness of 10\,mm. The short-side length is calibrated per scene so that the occlusion ratio of the task region reaches 90\%, ranging from 30 to 206\,mm.

As in \hyperref[sec:linked-4-4]{Section~4.4}, we restrict viewpoint movement to the vicinity of the starting point to reduce the effect of camera-pose distribution shift. InfoField3D recommends a new viewpoint within $20^\circ$ of the starting point. As controls, we keep the angular displacement unchanged and rotate the recommended movement direction by $90^\circ$, $180^\circ$, and $270^\circ$, respectively, to obtain three control viewpoints. This distinguishes whether the improvement comes from the movement itself or from the direction of movement. At the starting point, the recommended viewpoint, and the three control viewpoints, we execute the task with the frozen $\pi0.5$ using five random seeds, with the same scoring criteria as in \FGHtab{tab:app-completion-criteria}. The experiment includes 80 scenes and 2,000 rollouts in total.

\begin{table}[h]
\color{blue}
\centering
\caption{Mean task completion at each viewpoint.}\label{tab:app-occluder-vla}
\begin{tabularx}{\textwidth}{@{}X>{\raggedleft\arraybackslash}p{3.0cm}>{\raggedleft\arraybackslash}p{3.6cm}@{}}
\toprule
Viewpoint & Mean task completion (\%) & Change relative to the starting point (pp) \\
\midrule
Starting point, default camera with panel & 67.63 & --- \\
InfoField3D recommended viewpoint & 79.44 & $+11.81$ \\
Mean of the three controls & 67.58 & $-0.04$ \\
\bottomrule
\end{tabularx}
\end{table}

The recommended viewpoint raises the mean task completion from 67.63\% to 79.44\%, an improvement of 11.81 percentage points. The recommended viewpoint is 11.85 percentage points higher than the mean of the three controls.  This also shows that merely moving the camera does not bring improvement, and the key lies in the direction of movement.

The magnitude of this improvement is consistent with the result in \hyperref[sec:linked-4-4]{Section~4.4}, where the mean improvement brought by the recommended viewpoint is 11.73 percentage points. Under two settings that differ in both the starting point and the source of the information limitation, the viewpoints recommended by InfoField3D improve the task completion of the frozen VLA and outperform other directions with the same movement distance.

\clearpage
\section{Task Requirements, Approximation Limits, and Upstream Errors}
\label{app:task-limits-errors}

InfoField3D sits between semantic scene understanding and downstream task execution. This appendix examines both ends of this interface as well as its own operating limits. \FGHref{app:small-critical-regions} examines whether the objective function can express fine-grained task requirements. \FGHref{app:score-execution-gap} analyzes in which cases a higher score does not translate into better execution. \FGHref{app:approximation-limits} characterizes in which cases the geometric and occlusion approximations begin to fail, and how sensitive the results are to the parameters. \FGHref{app:semantic-errors} examines how the quality of the recommended viewpoints changes when upstream semantic localization contains errors. \FGHref{app:parameter-settings} gives the parameters used in the experiments.

\subsection{Small Critical Regions}
\label{app:small-critical-regions}

The objective function of InfoField3D focuses on the weighted sum of surface responses (\hyperref[sec:linked-3-2]{Section~3.2}). Fine manipulation often depends on only a small critical region; for example, when removing a handle, what needs to be seen clearly is the part that fixes it, not the entire handle. One concern is that a summation-form objective may be dominated by surfaces with larger area, thereby neglecting such small regions. In fact, the contribution of each surface point is weighted by task relevance, and the objective function leans toward wherever the weights are concentrated. This section illustrates this with an example.

We select one drawer handle consisting of a grip bar, two support rods, and two mounting rings, three parts that differ greatly in size. We take each of these parts in turn as the observation target, and then examine the result when the three are combined with given weights. The task region is specified directly by geometry. In the weighted combination, the responses of the three parts are first normalized separately by surface area and then summed with weights.

\begin{figure}[h]
\centering
\includegraphics[width=0.68\textwidth]{\appFGHfigdir/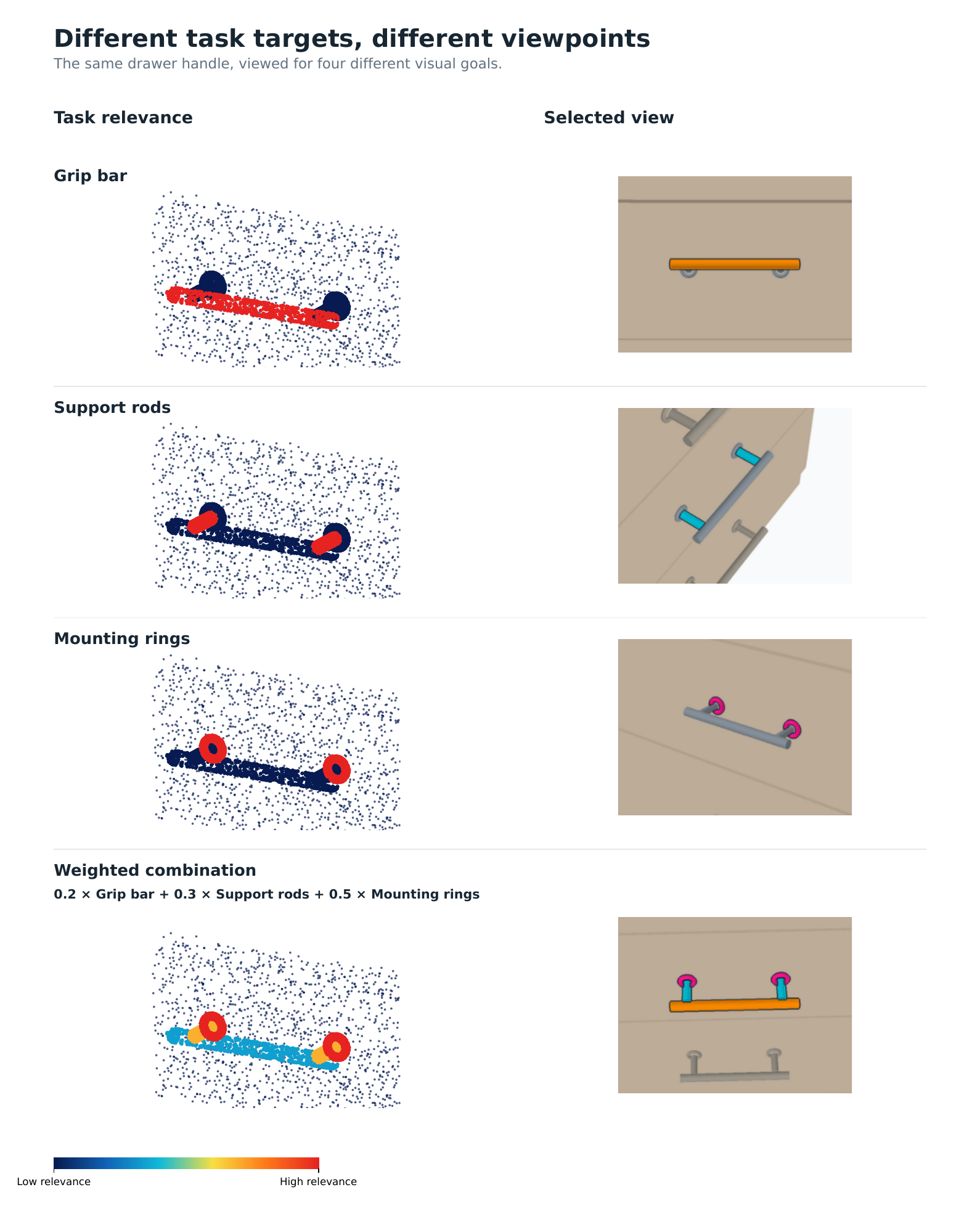}
\caption{Task-relevance maps (left) and recommended viewpoints (right) for the same drawer handle under four observation targets. The four rows are, in order, the grip bar, the support rods, the mounting rings, and the weighted combination of the three.}\label{fig:app-drawer-handle-targets}
\end{figure}

\begin{table}[h]
\color{blue}
\centering
\caption{Hard-ray visible fraction (\%) of the three parts at each recommended viewpoint.}\label{tab:app-drawer-handle-visibility}
\begin{tabular}{lrrr}
\toprule
Observation target & Grip bar & Support rods & Mounting rings \\
\midrule
Grip bar & 51.92 & 26.22 & 23.39 \\
Support rods & 51.92 & 44.55 & 39.30 \\
Mounting rings & 51.28 & 36.09 & 41.86 \\
Weighted combination & 51.12 & 44.36 & 40.63 \\
\bottomrule
\end{tabular}
\end{table}

As shown in \FGHfig{fig:app-drawer-handle-targets} and \FGHtab{tab:app-drawer-handle-visibility}, 
the final recommended viewpoint largely depends on what we prioritize. Regions assigned higher weights will contribute more strongly to the objective function, leading the recommended viewpoint to place greater emphasis on their surface responses. Therefore, if successful task execution depends on observing small joints, we can assign higher weights to these regions. If the task requires considering both small joints and multiple other components, we can include all of them in the objective function and weight them according to their importance. As shown in \FGHapp{app:multiple-targets}, when multiple targets are incorporated into the objective function, their responses can be combined using a product or other weighted formulations to ensure that each target receives sufficient attention, preventing small regions from being overlooked merely because of their limited surface area.

Therefore, in principle, any visual requirement that can ultimately be expressed as a spatial importance distribution can generally be addressed using InfoField3D.

\FloatBarrier
\subsection{Cases Where the Score Improves but Execution Does Not}
\label{app:score-execution-gap}

InfoField3D score measures the observation sufficiency of the task region and is not tailored to any specific downstream policy. \hyperref[sec:linked-4-4]{Section~4.4} and \FGHapp{app:vla-occluder} show that the recommended viewpoints improve the task completion of the frozen VLA overall, but not in every case. Understanding when exceptions occur and why helps delimit the scope of applicability of this score. We observe two main situations.

\textbf{Observation is not the bottleneck of execution.} As shown by the experimental results in \hyperref[fig:trident-and-vla-cases]{Figure~6(A.1)}, when the VLA performs poorly from the initial viewpoint, switching to the viewpoint recommended by InfoField3D can lead to substantial improvements in task execution (an absolute increase of 42.4 percentage points). In contrast, when the VLA already performs well from the initial viewpoint, the additional benefit of InfoField3D is relatively small and may even be negative. One possible explanation is that, when the VLA already achieves strong performance, the observation viewpoint is no longer the main bottleneck, leaving limited room for more informative observations to translate into further performance gains.

\textbf{The recommended viewpoint deviates from the viewpoint distribution familiar to the policy.} A VLA policy may have its own preference for the position of the third-person camera, and the vicinity of the default camera position commonly seen during training may be a region relatively familiar to it. We divide the trials of the \hyperref[sec:linked-4-4]{Section~4.4} experiment into two groups according to whether the angular distance from the recommended viewpoint to the default camera becomes smaller or larger.

\begin{figure}[h]
\centering
\includegraphics[width=0.42\textwidth]{\appFGHfigdir/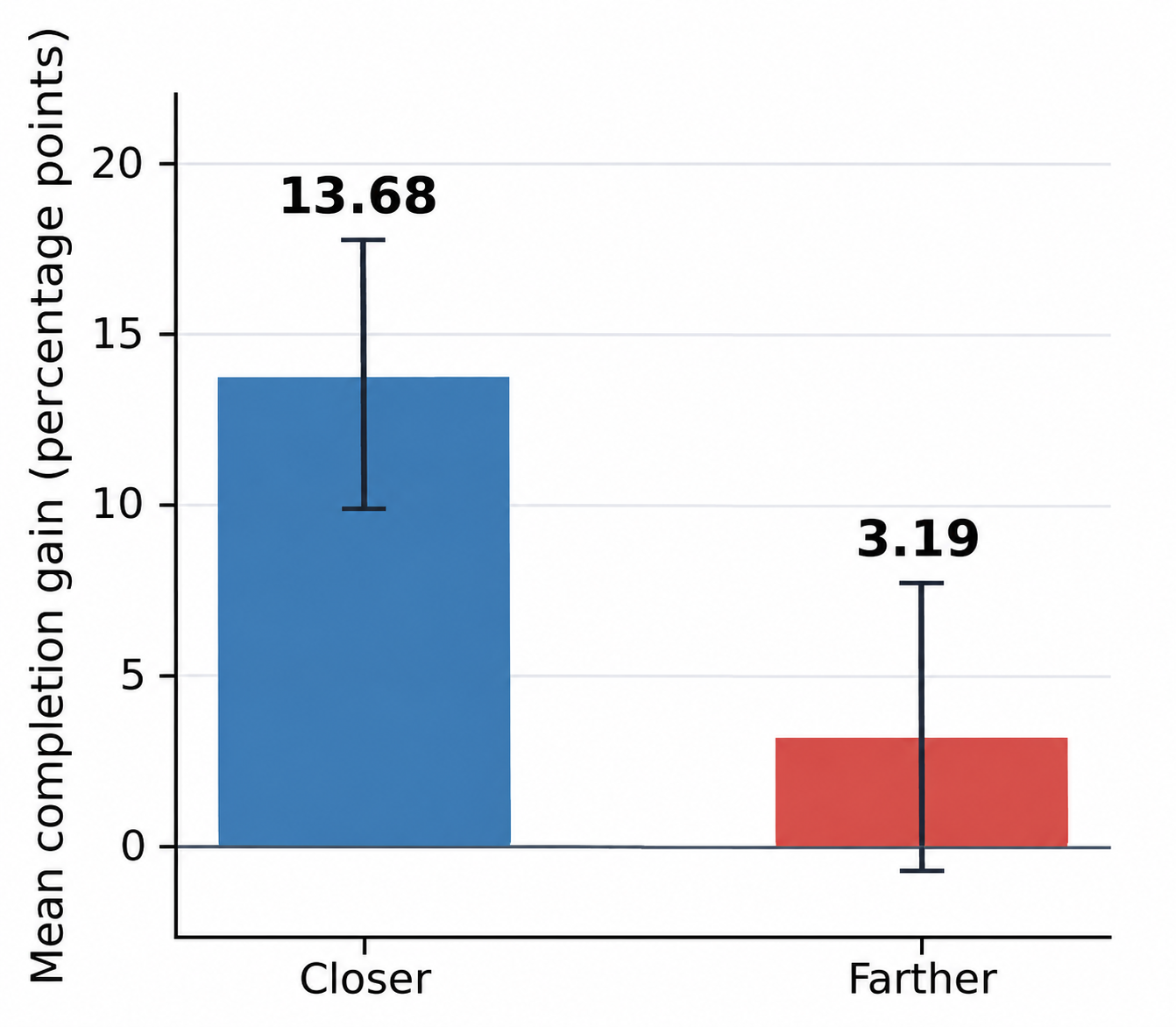}
\caption{Mean task-completion gain, grouped by whether the recommended viewpoint is closer to or farther from the default camera than the initial viewpoint in angular distance. Across all groups in the VLA experiment, 81.65\% of recommended viewpoints move closer to the default camera, while 18.35\% move farther away. Error bars indicate 95\% confidence intervals: [9.92, 17.69] and [-0.72, 7.66], respectively.}\label{fig:app-default-camera-distance}
\end{figure}

When the recommended viewpoint is closer to the default camera, the mean improvement is 13.68 percentage points. When it is farther from the default camera, the mean improvement is only 3.19 percentage points, and in some cases it is worse than the initial viewpoint (\FGHfig{fig:app-default-camera-distance}). In the representative example shown in \FGHfig{fig:app-scene-2087}, both geometric metrics and visual inspection indicate that the recommended viewpoint provides better visibility of the task-critical orange juice. However, the task completion rate decreases from 80\% to 70\%. We reasonably speculate that this may be related to the recommended viewpoint deviating further from the policy's preferred viewpoint.

\textbf{Representative case.} The task in scene 2087 is ``pick up the orange juice and place it in the basket.''

\begin{figure}[h]
\centering
\includegraphics[width=\textwidth]{\appFGHfigdir/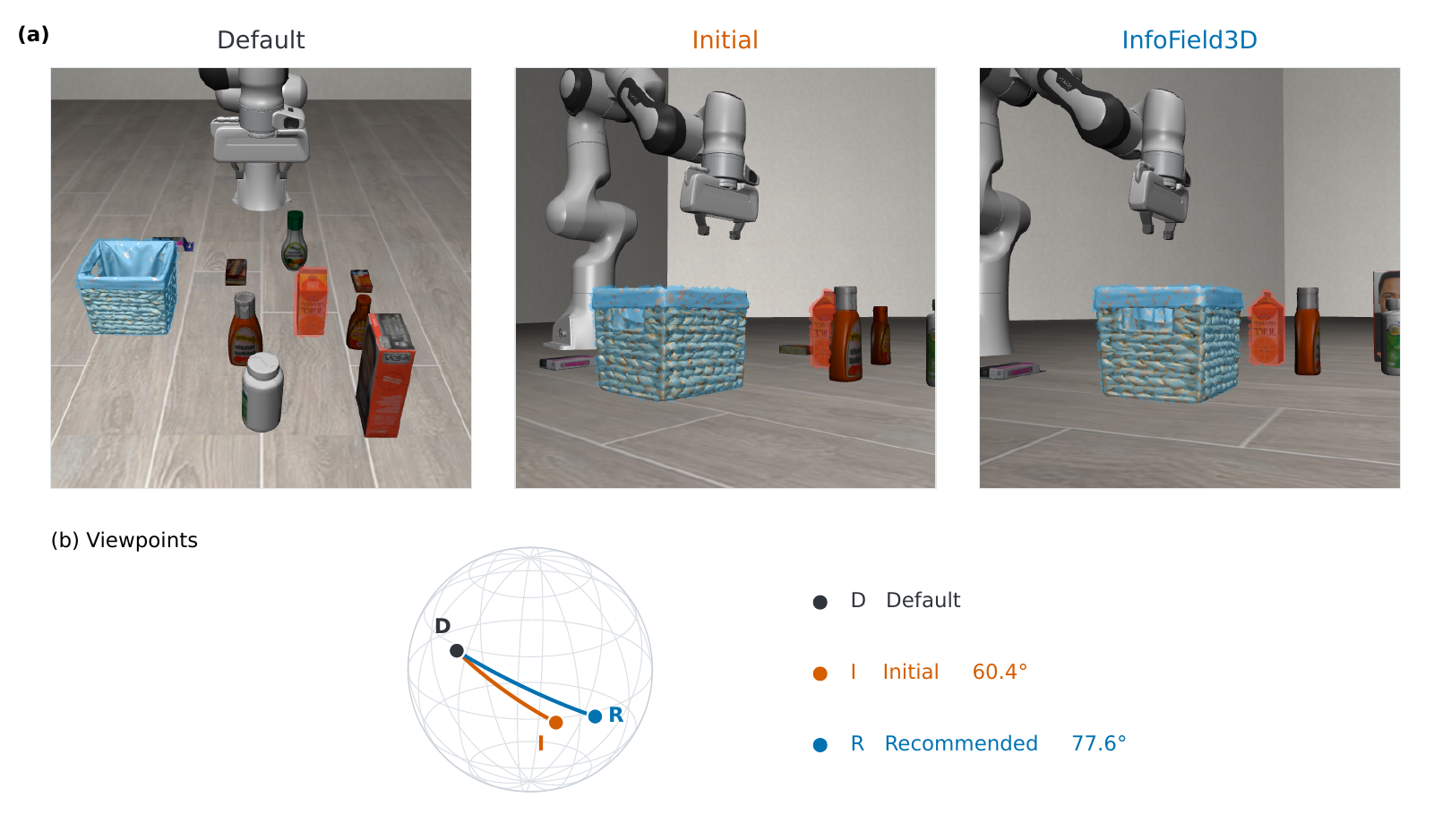}
\caption{Case of scene 2087. (a) Third-person images from the default, initial, and InfoField3D-recommended viewpoints, with task-relevant objects highlighted. (b) The three viewpoints on the observation sphere, together with the angular distances of the initial and recommended viewpoints to the default camera.}\label{fig:app-scene-2087}
\end{figure}

\textbf{Cause analysis.} There are three possible reasons why the score increases but execution does not improve: the geometric approximation distorts the score, the scoring function does not reflect the true observation quality, or the downstream policy fails to make use of the better observation. We examine them one by one.

For the geometric approximation, we re-evaluated the recommended viewpoints of two such cases with hard rays that do not rely on any smoothing approximation. The mean visible fraction over these two cases increases from 21.38\% at the initial viewpoints to 33.57\% at the recommended viewpoints, indicating that the viewpoints have indeed improved and that the score increase is not an artifact caused by approximation error. For the scoring function, the visibility of the task region in these cases has indeed improved, and GPT-4o score gives the same judgment, indicating that the score reflects the quantity it is intended to measure. What it does not model is the specific policy's preference for viewpoints. We therefore believe that the main cause lies on the downstream policy side, and both situations above belong to this category. This is consistent with the discussion in \hyperref[sec:linked-4-4]{Section~4.4}, namely that whether a VLA policy benefits depends not only on observation quality but also on camera-pose distribution shift.

\textbf{Possible extensions.} The above analysis also points to directions for improvement in combination with specific policies. First, if the policy's viewpoint preference can be represented spatially, it can be incorporated into the objective function of InfoField3D, for example, by assigning higher weights to positions closer to the region familiar to the policy, so that the recommended viewpoint strikes a trade-off between observation sufficiency and policy preference. Second, different policies operate differently, and the regions requiring attention also differ; such information can be provided to the multimodal model responsible for task understanding, so that the generated task-relevance map reflects the needs of that policy from the outset.

\subsection{Operating Limits of the Approximations and Parameter Sensitivity}
\label{app:approximation-limits}

Low-order spherical harmonics and smooth Gaussian occupancy enable efficient, differentiable computation. \hyperref[tab:master-1]{Table~1} shows that they maintain high fidelity overall on benchmark scenes. However, overall results cannot reveal on which geometries the approximations fail. This section conducts dedicated tests on the geometries most likely to cause problems (\FGHref{app:challenging-geometries}), and examines the sensitivity of the results to partition granularity and occlusion parameters (\FGHref{app:parameter-sensitivity}).

\subsubsection{Challenging Geometries}
\label{app:challenging-geometries}

We select four types of geometry, each targeting one weakness of the approximations.

\begin{itemize}[leftmargin=1.4em,itemsep=0.25em,topsep=0.2em]
\item \textbf{Narrow gap}: the target is inside a closed box and can only be seen through a vertical slit in the front wall. Smooth occupancy may ``fill in'' the gap. The gap width is progressively narrowed from 80\,mm to 5\,mm.
\item \textbf{Thin shell}: a thin plate stands in front of the target, blocking the line of sight on one side. Smooth occupancy may weaken the occlusion of the thin plate. The plate thickness is progressively reduced from 20\,mm to 1\,mm.
\item \textbf{Concave structure}: the target is the bottom of an open cylindrical cup, which can only be seen from a limited range above the opening. The ratio of cup depth to cup diameter is progressively increased from 0.25 to 3.
\item \textbf{Spatially concentrated target}: the target is a small patch on one face of a cuboid, whose directional response is concentrated within a very narrow range, challenging the representational capacity of low-order spherical harmonics. The ratio of the patch side length to the face side length is progressively reduced from 1 to 0.05.
\end{itemize}

Each type has 5 difficulty levels, and each level contains 5 randomly oriented instances, for a total of 100 configurations. Each configuration contains 50,000 surface points and 4,096 test positions. Task relevance is specified directly by geometry to exclude the influence of semantic annotation.

\begin{figure}[h]
\centering
\includegraphics[width=\textwidth]{\appFGHfigdir/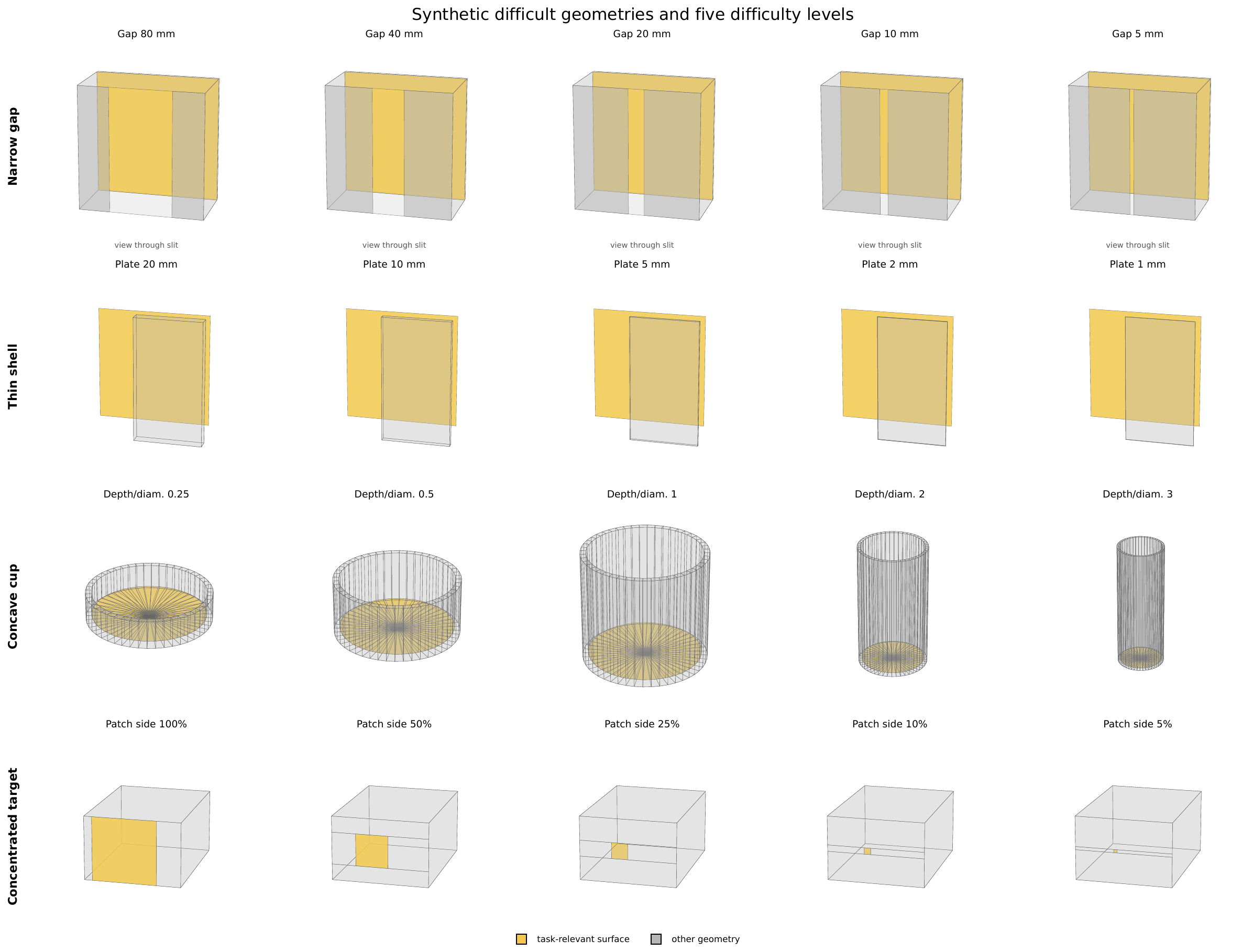}
\caption{The four types of synthetic challenging geometry at all five difficulty levels. Yellow marks the task-relevant surface, and gray marks other geometry.}\label{fig:app-challenging-geometries}
\end{figure}

Each configuration uses pointwise computation with exact occlusion as the reference, and we report two metrics. The first is the optimal-value attainment of \hyperref[sec:linked-4-1]{Section~4.1}, which reflects how good the viewpoint selected by the field is. The second is the ranking correlation, i.e., the Spearman correlation coefficient $\rho$ between the field response and the exact response at the test positions where the exact response is greater than zero, which reflects how accurately the field ranks all positions. For the narrow gap and the thin shell, we also report the scale ratio, i.e., the ratio of the gap width or plate thickness to the Gaussian width of the relevant unit.

\begin{table}[h]
\color{blue}
\centering
\caption{Fidelity on the four types of challenging geometry (\FGHfig{fig:app-challenging-geometries}). Difficulty increases from 1 to 5. Each metric is first computed within each instance and then averaged over the 5 instances of the same level.}\label{tab:app-challenging-geometries}
\small
\setlength{\tabcolsep}{4pt}
\begin{tabularx}{\textwidth}{@{}>{\raggedright\arraybackslash}p{2.5cm}c>{\raggedright\arraybackslash}X r>{\raggedleft\arraybackslash}p{2.1cm}r@{}}
\toprule
Geometry & Difficulty & Geometric size or ratio & Scale ratio & Optimal-value attainment (\%) & Spearman $\rho$ \\
\midrule
Narrow gap & 1 & Gap width 80\,mm & 17.78 & 99.47 & 0.9813 \\
Narrow gap & 2 & Gap width 40\,mm & 9.13 & 98.56 & 0.9550 \\
Narrow gap & 3 & Gap width 20\,mm & 4.66 & 97.84 & 0.9076 \\
Narrow gap & 4 & Gap width 10\,mm & 2.32 & 90.57 & 0.8591 \\
Narrow gap & 5 & Gap width 5\,mm & 1.13 & 86.35 & 0.8332 \\
\midrule
Thin shell & 1 & Plate thickness 20\,mm & 4.40 & 94.53 & 0.7409 \\
Thin shell & 2 & Plate thickness 10\,mm & 2.38 & 93.19 & 0.7342 \\
Thin shell & 3 & Plate thickness 5\,mm & 1.09 & 90.21 & 0.7051 \\
Thin shell & 4 & Plate thickness 2\,mm & 0.46 & 91.29 & 0.6956 \\
Thin shell & 5 & Plate thickness 1\,mm & 0.23 & 91.59 & 0.6682 \\
\midrule
Concave structure & 1 & Depth-to-diameter ratio 0.25 & --- & 100.00 & 0.9970 \\
Concave structure & 2 & Depth-to-diameter ratio 0.5 & --- & 100.00 & 0.9883 \\
Concave structure & 3 & Depth-to-diameter ratio 1 & --- & 99.98 & 0.9892 \\
Concave structure & 4 & Depth-to-diameter ratio 2 & --- & 99.96 & 0.9912 \\
Concave structure & 5 & Depth-to-diameter ratio 3 & --- & 99.96 & 0.9874 \\
\midrule
Spatially concentrated target & 1 & Side-length ratio 1 & --- & 99.97 & 0.9902 \\
Spatially concentrated target & 2 & Side-length ratio 0.5 & --- & 100.00 & 0.9992 \\
Spatially concentrated target & 3 & Side-length ratio 0.25 & --- & 99.92 & 0.9958 \\
Spatially concentrated target & 4 & Side-length ratio 0.1 & --- & 99.72 & 0.9771 \\
Spatially concentrated target & 5 & Side-length ratio 0.05 & --- & 98.74 & 0.9348 \\
\bottomrule
\end{tabularx}
\end{table}

\FloatBarrier
On the concave structure and the spatially concentrated target, the approximations are reliable. At all difficulty levels, the optimal-value attainment is no lower than 98.7\%, and Spearman $\rho$ is no lower than 0.93. Even when the target's side length is only 5\% of that of the face, low-order spherical harmonics are still sufficient to select a near-optimal viewpoint.

The weaknesses of the approximations appear on the narrow gap and the thin shell. On the narrow gap, the optimal-value attainment stays above 97.8\% when the scale ratio is no lower than 4.66. As the gap narrows further, the results gradually degrade: at scale ratios of 2.3 and 1.1, it is 90.57\% and 86.35\%, respectively, and Spearman $\rho$ drops from 0.98 to 0.83. On the thin shell, the optimal-value attainment stays between 90\% and 95\% at all difficulty levels, but the ranking correlation is clearly lower than on the other geometries, with Spearman $\rho$ between 0.67 and 0.74. That is, near thin shells the field can still select a near-optimal viewpoint, but its ranking of all positions is less accurate than on the other geometries. The degradation on both types of geometry is gradual, and no sudden failure occurs.

Reducing the minimum Gaussian width effectively mitigates smoothing errors in thin-shell and narrow-gap settings. As shown in Table~\ref{tab:minimum-width-mitigation}, smaller widths substantially improve the correlation with exact responses, while maintaining or improving optimal-value attainment.

\begin{table}[t]
\color{blue}
\centering
\caption{Effect of reducing the minimum Gaussian width in the
hardest thin-shell and narrow-gap settings.
All other parameters remain fixed.}
\label{tab:minimum-width-mitigation}
\begin{tabular}{@{}ccccc@{}}
\toprule
& \multicolumn{2}{c}{Thin shell}
& \multicolumn{2}{c}{Narrow gap} \\
\cmidrule(lr){2-3}\cmidrule(lr){4-5}
Width / default
& Attainment (\%) & Spearman's $\rho$
& Attainment (\%) & Spearman's $\rho$ \\
\midrule
$1$   & 91.59  & 0.6682 & 86.35 & 0.8332 \\
$1/4$ & 100.00 & 0.8280 & 86.41 & 0.8846 \\
$1/8$ & 99.80  & 0.9111 & 86.48 & 0.8836 \\
\bottomrule
\end{tabular}
\end{table}

To determine the source of the error, we make two replacements separately at the hardest level of the narrow gap and of the thin shell: replacing Gaussian occlusion with exact occlusion, or increasing the spherical harmonic degree from 4 to 8. After switching to exact occlusion, the optimal-value attainment recovers to above 99\%, while increasing the spherical harmonic degree brings no change at all (\FGHtab{tab:app-error-source}). The error comes from the smoothing approximation of occlusion rather than from spherical harmonic truncation.

\begin{table}[h]
\color{blue}
\centering
\caption{Sources of error at the hardest level; values are optimal-value attainment (\%).}\label{tab:app-error-source}
\begin{tabularx}{\textwidth}{@{}l*{3}{>{\centering\arraybackslash}X}@{}}
\toprule
Geometry & Default setting, $L=4$ & With exact occlusion & Spherical harmonic degree increased to 8 \\
\midrule
Narrow gap & 86.35 & 99.24 & 86.35 \\
Thin shell & 91.59 & 99.58 & 91.59 \\
\bottomrule
\end{tabularx}
\end{table}

Therefore, the operating limits of InfoField3D are mainly determined by the relative relationship between the scale of the occluding structures and the Gaussian smoothing scale. When the scene contains gaps or thin plates whose scale is close to the smoothing width, and they play a decisive role in the visibility of the task region, the quality of the recommended viewpoints decreases to some extent; this decrease can be alleviated by further refining the parameters.

\FloatBarrier
\subsubsection{Sensitivity to Partition Granularity and Occlusion Parameters}
\label{app:parameter-sensitivity}

The above analysis shows that the effectiveness of the approximations depends on the smoothing scale. The next question is how sensitive the quality of the recommended viewpoints is to the values of these parameters in benchmark scenes. On the 80 scenes, we change only one of the number of units, the minimum Gaussian width $\sigma_0$, and the attenuation strength $\kappa$ at a time, keeping the others at their defaults.

\begin{table}[h]
\color{blue}
\centering
\caption{Parameter sensitivity. The effect of different parameter settings on the mean Task-region visibility is evaluated on the 80 scenes. Results are expressed as changes relative to the default setting, in percentage points.}\label{tab:app-parameter-sensitivity}
\begin{tabularx}{\textwidth}{@{}>{\raggedright\arraybackslash}X>{\raggedright\arraybackslash}p{3.6cm}>{\raggedleft\arraybackslash}p{3.6cm}@{}}
\toprule
Parameter & Adjustment relative to the default & Change in Task-region visibility (pp) \\
\midrule
Number of units & 0.8--2$\times$ & $-0.30$ to $+0.28$ \\
Minimum Gaussian width $\sigma_0$ & $0.5$--$2\times$ & $+0.31$ to $-3.25$ \\
Attenuation strength $\kappa$ & 0.5--2$\times$ & $-0.38$ to $+0.92$ \\
\bottomrule
\end{tabularx}
\end{table}

When the number of units varies between 0.8 and 2 times the default, the attenuation strength varies between 0.5 and 2 times the default, or the minimum Gaussian width is halved, the change in mean Task-region visibility is less than 1 percentage point (\FGHtab{tab:app-parameter-sensitivity}). Only when the minimum Gaussian width is doubled does the result drop, by 3.25 percentage points. This corresponds to over-smoothing, consistent with the conclusion in \FGHapp{app:challenging-geometries} that the error comes from occlusion smoothing. Overall, the quality of the recommended viewpoints is not highly sensitive to these parameters.

\subsection{Sensitivity to Semantic Localization Errors}
\label{app:semantic-errors}

The objective function of InfoField3D is built on the task-relevance map, which comes from an upstream multimodal model (\FGHapp{app:semantic-annotation}). Upstream semantic localization cannot be perfectly accurate. InfoField3D faithfully optimizes the given objective, so it is necessary to understand how much of the quality of the recommended viewpoints can be retained when the objective itself is wrong.

We simulate three common types of errors in open-vocabulary localization and apply them to the task-relevance map.

\begin{itemize}[leftmargin=1.4em,itemsep=0.25em,topsep=0.2em]
\item \textbf{Noise}: local false-positive patches appear on background surfaces.
\item \textbf{Spill}: the target region expands to surrounding surfaces.
\item \textbf{Erosion}: contiguous parts of the target region are missed.
\end{itemize}

Each type has three strength levels. Strength is measured by the total variation between the task-relevance distributions before and after perturbation, with the three levels being 0.10, 0.25, and 0.50, and the three types of perturbation have the same strength at the same level.

Each method searches for viewpoints using only the perturbed task-relevance map. The recommended viewpoints are then evaluated on the unperturbed map using hard-ray Task-region visibility, so that what is measured is their quality under the true task requirements. At each perturbation level, we report the loss in Task-region visibility for paired runs, defined as the decrease from the unperturbed-map recommendation to the perturbed-map recommendation, with both viewpoints evaluated on the unperturbed map using the same hard-ray metric. Losses are averaged over runs within each scene and then equally over the 80 scenes. The baselines are also tested to determine whether this sensitivity is specific to InfoField3D.

\begin{table}[H]
\color{blue}
\centering
\caption{Mean loss in Task-region visibility
(percentage points; lower is better) under the strongest
perturbation level (total variation 0.50).
Bold values indicate the lowest mean loss in each column.}
\label{tab:app-perturbation-strongest}
\begin{tabular}{lrrr}
\toprule
Method & Noise & Spill & Erosion \\
\midrule
InfoField3D & \textbf{1.55} & 4.00 & 1.27 \\
GP-BO       & 9.94  & \textbf{0.78} & 1.39 \\
CIM         & 12.63 & 2.89 & 4.91 \\
CMA-ES      & 8.38  & 1.92 & 2.58 \\
GradientNBV & 13.30 & 4.87 & 2.77 \\
Sobol       & 6.34  & 1.00 & \textbf{1.26} \\
Random      & 6.88  & 0.82 & 1.38 \\
\bottomrule
\end{tabular}

\smallskip
{\footnotesize\textit{Note:}
VISO-grasp does not read the pointwise task-relevance map,
and its input is unaffected by the perturbations,
so it is not included in the statistics.\par}
\end{table}

\begin{table}[h]
\color{blue}
\centering
\caption{Mean loss in Task-region visibility
(percentage points; lower is better) at the three
perturbation strength levels.}
\label{tab:app-perturbation-all-levels}
\small
\begin{tabular}{lrrr}
\toprule
Method & Noise & Spill & Erosion \\
\midrule
\multicolumn{4}{@{}l}{\textit{Total variation 0.10}} \\
InfoField3D & 0.37 & 0.06 & 0.30 \\
GP-BO       & 0.46 & 0.29 & 0.58 \\
CIM         & 1.30 & 1.48 & 1.11 \\
CMA-ES      & 1.79 & 1.10 & 1.85 \\
GradientNBV & 5.49 & 3.47 & 2.75 \\
Sobol       & 0.38 & 0.11 & 0.48 \\
Random      & \textbf{0.15} & \textbf{0.04} & \textbf{0.25} \\
\midrule
\multicolumn{4}{@{}l}{\textit{Total variation 0.25}} \\
InfoField3D & 1.00 & \textbf{0.15} & \textbf{0.40} \\
GP-BO       & 2.33 & 0.34 & 0.96 \\
CIM         & 4.76 & 2.32 & 2.50 \\
CMA-ES      & 4.19 & 1.62 & 1.71 \\
GradientNBV & 9.00 & 3.48 & 2.80 \\
Sobol       & \textbf{0.87} & 0.37 & 1.00 \\
Random      & \textbf{0.87} & 0.26 & 0.60 \\
\midrule
\multicolumn{4}{@{}l}{\textit{Total variation 0.50}} \\
InfoField3D & \textbf{1.55} & 4.00 & 1.27 \\
GP-BO       & 9.94  & \textbf{0.78} & 1.39 \\
CIM         & 12.63 & 2.89 & 4.91 \\
CMA-ES      & 8.38  & 1.92 & 2.58 \\
GradientNBV & 13.30 & 4.87 & 2.77 \\
Sobol       & 6.34  & 1.00 & \textbf{1.26} \\
Random      & 6.88  & 0.82 & 1.38 \\
\bottomrule
\end{tabular}
\end{table}

Under the two weaker perturbation levels, InfoField3D maintains low losses in Task-region visibility across all three types of error (\FGHtab{tab:app-perturbation-all-levels}). At the strongest level, the three error types exhibit distinct effects (\FGHtab{tab:app-perturbation-strongest}).

For noise, InfoField3D achieves the lowest loss among all methods (1.55 percentage points), compared with 6.34–13.30 points for the baselines other than GP-BO, whose loss is 9.94 points. For erosion, InfoField3D incurs a loss of only 1.27 points, nearly matching the lowest loss of 1.26 points. For spill, however, its loss increases to 4.00 points, exceeding that of most baselines.

We conjecture that these differences arise from the spatial structure of the errors. Spill introduces a spatially coherent region of erroneous relevance near the true target, which can systematically shift the recommended viewpoint toward directions from which the expanded region is visible. In contrast, scattered false positives do not necessarily favor the same observation direction, while local misses typically preserve the target's approximate location and main orientation. Thus, these two errors tend to have less influence on the recommended viewpoint. Although all three perturbations have the same total-variation strength, equal changes in relevance weight do not imply equal spatial effects.

Overall, InfoField3D shows strong stability against background false positives and local misses, while being more sensitive to substantial overestimation of the target region. This highlights a limitation of the method under spatially coherent upstream semantic errors.

\subsection{Parameter Settings}
\label{app:parameter-settings}

All experiments in the main text use the same partitioning and occlusion settings for every scene, without tuning for individual scenes. The only scene-dependent quantity is the number of units $M$, which is chosen from three levels by a fixed rule. \FGHtab{tab:app-field-parameters} lists these settings.

\begin{table}[h]
\color{blue}
\centering
\caption{Partitioning and occlusion settings, shared by all scenes.}\label{tab:app-field-parameters}
\begin{tabularx}{\textwidth}{@{}l>{\raggedright\arraybackslash}Xl@{}}
\toprule
Item & Setting & Reference \\
\midrule
\multicolumn{3}{@{}l}{\textit{Partitioning}} \\
\quad Scheme & Recursive principal-axis bisection & Sec.~3.4 \\
\quad Unit to split & Largest area-weighted spread & See below \\
\quad Split position & Area-weighted median along the principal axis & \\
\quad Number of units $M$ & 224, 448, or 896 (605 on average) & See below \\
\addlinespace[0.4em]
\multicolumn{3}{@{}l}{\textit{Occlusion}} \\
\quad Occupancy model & One anisotropic Gaussian per unit & Eqs.~(9)--(11) \\
\quad Minimum Gaussian width $\sigma_0$ & 2\,mm & Eq.~(10) \\
\quad Attenuation strength $\kappa_k$ & Centerline optical depth of 12 & Eq.~(12) \\
\quad Propagation path & Finite segment, integrated in closed form & Eq.~(18) \\
\quad Occluders & All other units, without spatial truncation & \\
\bottomrule
\end{tabularx}
\end{table}

The settings are described as follows.

\begin{itemize}[leftmargin=1.4em,itemsep=0.25em,topsep=0.2em]
\item \textbf{Partitioning}: the unit to be split is the one with the largest total area-weighted squared deviation $\sum_{i\in k} a_i\lVert p_i-\mu_k\rVert^2$, with $a_i$ and $\mu_k$ as in Eq.~(9), and it is split at the area-weighted median along its principal axis, so that the two subsets have approximately equal area. No semantic part priors are used (\FGHapp{app:geometric-partitioning}). Each resulting unit serves both as an information source and as an occluder for other units.
\item \textbf{Number of units}: $M=224$ by default; $M=448$ when the lateral width of the scene exceeds 0.83\,m; and $M=896$ when the surface area of the task objects exceeds 0.38\,m$^2$. When both conditions are met, the larger value is taken. The thresholds are set empirically and are the same for all scenes. Levels are used because the scale differs greatly across scenes, and the same number of units for all scenes would not be reasonable.
\item \textbf{Occupancy and attenuation}: the Gaussian of each unit uses the area-weighted mean and covariance of its surface samples, with purely geometric area weights that do not include task relevance. The centerline optical depth is the optical depth along the line through the Gaussian center along its narrowest principal axis; setting it to 12 means that a line of sight crossing the thinnest part of a unit retains only $e^{-12}\approx 6\times10^{-6}$ of the response. The same value is used for all units, so $\kappa_k$ is determined by the shape of each unit rather than tuned per scene.
\item \textbf{Occlusion path}: attenuation is integrated along the finite segment from the reference center of the source unit to the observation position, in closed form and without numerical discretization. All units other than the source unit itself are included, including other units of the same object, with no truncation by distance or Gaussian support.
\end{itemize}

\endgroup

\end{document}